\documentclass[sigconf]{acmart}
\usepackage[utf8]{inputenc}
\usepackage{booktabs,multirow,siunitx}
\usepackage{listings}

\lstdefinelanguage{JavaScript}{
  morekeywords={break,case,catch,continue,debugger,default,delete,do,else,finally,for,function,if,in,instanceof,new,return,switch,this,throw,try,typeof,var,let,const,void,while,with,yield,async,await,import,from,export,class,extends,super
  },
  sensitive=true,
  morecomment=[l]{//},
  morecomment=[s]{/*}{*/},
  morestring=[b]",
  morestring=[b]',
}
\lstdefinestyle{jsstyle}{
  language=JavaScript,
  basicstyle=\ttfamily\footnotesize,
  numbers=left,
  numberstyle=\tiny,
  frame=single,
  breaklines=true,
  columns=fullflexible,
  tabsize=2,
  showstringspaces=false,
  keywordstyle=\color{black},
  commentstyle=\color{black},
  stringstyle=\color{black},
  xleftmargin=2em
}
\usepackage{float}
\AtBeginDocument{%
  }

\copyrightyear{2026}
\acmYear{2026}
\setcopyright{cc}
\setcctype{by}
\acmConference[CHI '26]{Proceedings of the 2026 CHI Conference on Human Factors in Computing Systems}{April 13--17, 2026}{Barcelona, Spain}
\acmBooktitle{Proceedings of the 2026 CHI Conference on Human Factors in Computing Systems (CHI '26), April 13--17, 2026, Barcelona, Spain}
\acmPrice{}
\acmDOI{10.1145/3772318.3790668}
\acmISBN{979-8-4007-2278-3/2026/04}

\acmSubmissionID{8686}

\usepackage{acmart-taps} 
\usepackage{cleveref} 

\usepackage{graphicx}

\newcommand{\circlenum}[1]{%
  \raisebox{-0.25em}{%
    \textnormal{%
      \ooalign{%
        \hfil\resizebox{!}{1.0em}{\color{black!50}$\bullet$}\hfil\cr
        \hfil\raisebox{0.28em}{\color{white}\sffamily\bfseries\small #1}\hfil\cr
      }%
    }%
  }%
}

\definecolor{themebg}{gray}{0.90}

\begin{document}

\title{WordCraft: Scaffolding the Keyword Method for L2 Vocabulary Learning with Multimodal LLMs}



\author{Yuheng Shao}
\authornote{Both authors contributed equally to this research.}
\orcid{0009-0008-6991-6427}
\affiliation{%
  \institution{School of Information Science and Technology, ShanghaiTech University}
  \city{Shanghai}
  \country{China}
}
\email{shaoyh2024@shanghaitech.edu.cn}

\author{Junjie Xiong}
\authornotemark[1]
\orcid{0009-0004-6005-7508}
\affiliation{%
  \institution{School of Information Science and Technology, ShanghaiTech University}
  \city{Shanghai}
  \country{China}}
\email{xiongjj2025@shanghaitech}

\author{Chaoran Wu}
\orcid{0009-0009-3547-7400}
\affiliation{%
  \institution{School of Information Science and Technology, ShanghaiTech University}
  \city{Shanghai}
  \country{China}}
\email{wuchr2023@shanghaitech.edu.cn}

\author{Xiyuan Wang}
\orcid{0009-0008-1839-2010}
\affiliation{%
  \institution{School of Information Science and Technology, ShanghaiTech University}
  \city{Shanghai}
  \country{China}}
\email{wangxy7@shanghaitech.edu.cn}

\author{Ziyu Zhou}
\orcid{0009-0002-0866-690X}
\affiliation{%
  \institution{School of Creativity and Art, ShanghaiTech University}
  \city{Shanghai}
  \country{China}}
\email{zhouzy12023@shanghaitech.edu.cn}

\author{Yang Ouyang}
\orcid{0009-0000-5841-7659}
\affiliation{%
  \institution{School of Information Science and Technology, ShanghaiTech University}
  \city{Shanghai}
  \country{China}}
\email{ouyy@shanghaitech.edu.cn}

\author{Qinyi Tao}
\orcid{0009-0003-7979-747X}
\affiliation{%
  \institution{Shanghai Fengxian Dai Wen Middle School}
  \city{Shanghai}
  \country{China}}
\email{kominatoshione25@gmail.com}

\author{Quan Li}
\authornote{Corresponding Author.}
\orcid{0000-0003-2249-0728}
\affiliation{%
  \institution{School of Information Science and Technology, ShanghaiTech University}
  \city{Shanghai}
  \country{China}
}
\email{liquan@shanghaitech.edu.cn}

\renewcommand{\shortauthors}{Yuheng Shao, Juejie Xiong et al.}

\begin{abstract}
Applying the keyword method for vocabulary memorization remains a significant challenge for L1 Chinese–L2 English learners. They frequently struggle to generate phonologically appropriate keywords, construct coherent associations, and create vivid mental imagery to aid long-term retention. Existing approaches, including fully automated keyword generation and outcome-oriented mnemonic aids, either compromise learner engagement or lack adequate process-oriented guidance. To address these limitations, we conducted a formative study with L1 Chinese-L2 English learners and educators (N=18), which revealed key difficulties and requirements in applying the keyword method to vocabulary learning. Building on these insights, we introduce \textit{WordCraft}, a learner-centered interactive tool powered by Multimodal Large Language Models (MLLMs). \textit{WordCraft} scaffolds the keyword method by guiding learners through keyword selection, association construction, and image formation, thereby enhancing the effectiveness of vocabulary memorization. Two user studies demonstrate that \textit{WordCraft} not only preserves the generation effect but also achieves high levels of effectiveness and usability.
\end{abstract}

\ccsdesc[500]{Human-centered computing}
\ccsdesc[500]{Human-centered computing~Human computer interaction (HCI)}
\ccsdesc[300]{Human-centered computing~Interactive systems and tools}
\ccsdesc[300]{Human-centered computing~User interface toolkits}
\keywords{Keyword Method, Vocabulary Learning, Generation Effect, Creative Support Learning}

\maketitle

\section{INTRODUCTION}

\par Effective vocabulary retention in a second language (L2) requires learners to develop a deep, flexible understanding of lexical items~\cite{CHINGSHYANGCHANG2007534,goulden1990large,schmitt2017much}. Yet many commonly adopted instructional approaches prioritize efficient information delivery~\cite{chen2012effects}, positioning learners as largely passive recipients. Such ``one-size-fits-all'' approaches can reduce opportunities for learners to engage with rich, cross-modal information about words and may constrain the deeper cognitive processing required for durable vocabulary learning.

\par In contrast, the keyword method~\cite{pressley1982mnemonic, shapiro2005investigation,atkinson1975application,levin1981mnemonic,beaton1995retention,al2011effectiveness}, as a widely used mnemonic technique, offers distinct advantages in addressing these limitations. It involves two core steps: first, learners identify a keyword from their first language (L1) that is phonetically similar to the target L2 word, and second, they construct a contextualized connection between the keyword and the meaning of the L2 word by forming a vivid mental image~\cite{atkinson1975application}. This process enhances memory by capitalizing on dual coding~\cite{paivio2013imagery,clark1991dual} through the integration of the phonological and visual information of words, while also relying heavily on the learner's active cognitive participation to trigger the generation effect—a robust finding in cognitive psychology indicating that information actively generated by learners tends to be remembered better than information received passively~\cite{slamecka1978generation,bertsch2007generation}. Grounded in these cognitive principles, the keyword method has proven empirically effective for vocabulary retention~\cite{QU2024e25212}. However, applying the keyword method in practice places a high demand on L2 learners and presents significant challenges. Learners frequently encounter difficulties in selecting phonologically appropriate keywords~\cite{park2006phonological,consiglio2018keyword}, evaluating semantic compatibility~\cite{savva2014transphoner,article9,thomas1996learning}, constructing coherent associations~\cite{thomas1996learning}, and generating vivid mental imagery~\cite{bell2017lindamood}. These creatively demanding steps can result in low-quality mnemonics and increased cognitive load, ultimately undermining both the learning experience and memory performance. In this work, we specifically investigate and address these challenges within the context of L1-Chinese learners of L2-English.

\par Researchers have developed various computational enhancements for the keyword method. Many efforts focus on enhancing learner-generated outcomes to strengthen dual-coding, for instance, by visualizing associations as images~\cite{attygalle2025text} or embedding them in Augmented Reality environments~\cite{weerasinghe2022vocabulary}. However, these methods still rely on the learner to complete the creative work independently, offering little to no scaffolding for the cognitively demanding generative process. To more directly address this cognitive load, other efforts turn to end-to-end automation, where systems generate keywords and associations for the learner~\cite{lee2024exploring,savva2014transphoner}. While this strategy can streamline the learning process, it often comes at the cost of diminishing the generation effect. Consequently, a significant gap remains in providing process-oriented scaffolding that successfully navigates the trade-off between reducing cognitive load and preserving generation effect essential for memory retention.

\par For such cognitively demanding tasks, current Human-Computer Interaction (HCI) research has extensively explored the potential of creativity support tools (CSTs) to augment user creative practice~\cite{frich2019mapping}. In the context of language learning, CSTs such as semantic maps~\cite{article3,article10,margosein1982effects,heimlich1986semantic}, flashcards~\cite{mutar2024flashcard,kornell2009optimising,nugroho2012improving}, and language games~\cite{ismayilli2025impact,yip2006online,rojabi2022kahoot,10.1145/3613905.3648107} have been widely employed to facilitate content ideation and sustain learner motivation. However, these tools are predominantly limited to providing initial creative stimulation or exploring inter-word relationships. More recently, the integration of Multimodal Large Language Models (MLLMs) with CSTs has shown promising advances in creative support~\cite{10.1145/3706598.3713935}, mutimodal alignment~\cite{10.1145/3706598.3713818}, and cognitive scaffold~\cite{10.1145/3613904.3642185}. However, existing research remains insufficiently aligned with the unique demands of the keyword method. First, many systems prioritize maximizing support during the process to ensure optimal outcomes, offering limited scaffolding that preserves the level of cognitive effort essential for effective learning. Second, their interaction paradigms tend to be simple and rigid, restricting their ability to provide sustained assistance across the iterative and non-linear cognitive processes inherent to the keyword method. These limitations, both in computational enhancements to the keyword method and in current MLLM-integrated CSTs, highlight the pressing need for research that delivers meaningful, process-level support to guide learners through its cognitively intensive tasks.

\par Drawing on feedback from L1 Chinese learners of English, instructors, and researchers, we mapped learners' cognitive workflows for applying the keyword method, identified key challenges and considerations for MLLM integration, and distilled six design requirements to guide system development. Building on these insights, we developed \textit{WordCraft}, a process-level creativity support tool designed to scaffold the keyword method while managing cognitive load and preserving the generation effect. The current system is constrained to micro-learning sessions of 8–16 vocabulary items to maintain content quality and prevent cognitive overload. To evaluate the system, we conducted two complementary user studies with L1-Chinese L2-English participants. The first was a between-subjects experiment with 48 participants comparing the effectiveness and usability of \textit{WordCraft} against a GPT-4o–based chat interface and a traditional flashcard method. The second was a within-subjects experiment with 20 participants testing whether \textit{WordCraft} preserves the generation effect by comparing recall performance using self-generated versus system-generated cues. In summary, the main contributions of this work are as follows: 
\begin{itemize}
    \item We conduct a formative study to uncover cognitive workflows, key challenges, and considerations for integrating MLLMs into the keyword method.
    \item We present \textit{WordCraft}, a process-level creativity support tool that integrates MLLMs to guide the keyword method while balancing cognitive load and preserving the generation effect.
    \item We evaluate \textit{WordCraft} through two user studies that demonstrate its effectiveness, usability, and ability to maintain the generation effect.
\end{itemize}

\section{BACKGROUND AND RELATED WORK}

\subsection{Background}
\subsubsection{Vocabulary Learning Strategies}
\par Research on L2 vocabulary learning has proposed a wide range of instructional approaches, which can be broadly categorized into methods that emphasize efficient exposure to lexical items~\cite{nation2001learning,AKBARIAN2020102261}, structured practice~\cite{dekeyser2007practice,article123}, and multimedia enrichment~\cite{article1234}. Efficiency-oriented techniques such as word lists~\cite{inbook,article1234567,dang2025applications}, flashcards~\cite{inbook,lei2022learning,boroughani2023mobile}, and spaced-repetition systems~\cite{mizuno2000test,article12345} aim to maximize the number of items learners can encounter and rehearse within limited time. Classroom-based explicit instruction and teacher-led explanation~\cite{ellis1994study,article123456} similarly prioritize streamlined delivery of word meanings and usage patterns. Computer-assisted language learning tools~\cite{arakawa2022vocabencounter,peng2023storyfier,leong2024putting, zhao2024language,lee2024open,ngo2024use} often extend this information-transmission paradigm through the incorporation of audio, images, or example sentences, enriching the input while still presenting largely prepackaged content.

\par Although these approaches have clear benefits, they typically operate within a ``one-size-fits-all'' paradigm of efficient information delivery that casts learners as largely passive recipients. Consequently, they offer limited opportunities for learners to explore and integrate rich, cross-modal information about words, potentially constraining the deeper cognitive processing needed for durable vocabulary retention. In response to these limitations, the keyword method~\cite{pressley1982mnemonic, shapiro2005investigation,atkinson1975application,levin1981mnemonic,beaton1995retention,al2011effectiveness} was introduced to foster more active engagement with new vocabulary and has been shown to substantially enhance recall~\cite{shapiro2005investigation,sagarra2006key,piribabadi2014effect,al2011effectiveness}.

\subsubsection{Keyword Method}

\par The keyword method is a mnemonic strategy that involves two core cognitive processes~\cite{atkinson1975application}. It pairs a familiar native-language keyword with the target word, and then constructing a vivid mental image or scenario sentences linking the keyword to the target word's meaning. 


\par Extensive research has validated the keyword method's efficacy across various learning contexts~\cite{shapiro2005investigation,sagarra2006key,piribabadi2014effect,al2011effectiveness}. For example, Atkinson et al.~\cite{atkinson1975application} found that it significantly outperformed a control group (72\% vs. 46\%) in learning Russian words. Beaton et al.~\cite{beaton1995retention} showed lasting retention even after ten years, with brief reviews boosting recall. Field studies in regions such as Asia~\cite{qu2024facilitative}, the Americas~\cite{campos2010efficacy}, and the Middle East~\cite{al2019effect} have further demonstrated its adaptability, enhancing both immediate and delayed recall. The method's effectiveness is supported by two key cognitive theories. Dual-coding theory suggests that encoding information through both verbal and visual channels leads to richer, more durable memories~\cite{paivio2013imagery}, with the keyword method creating two cognitive pathways by linking phonological associations with vivid imagery~\cite{cancino2021role}. Additionally, the generation effect posits that self-generated information is remembered better than passively received information~\cite{slamecka1978generation}, which the keyword method leverages by having learners generate their own keywords and associations, resulting in stronger memory traces. Studies have also shown that self-generated keywords improve recall compared to those provided by instructors or experimenters~\cite{campos2004importance}. Despite its proven effectiveness, the keyword method presents several practical challenges for learners in enhancing vocabulary retention.  Identifying native language keywords with sufficient phonetic similarity to the target word can be difficult~\cite{bird1999examination,lee2024exploring}. Even with appropriate keywords, learners often struggle to construct vivid and semantically meaningful associations, largely due to a lack of effective cognitive strategies for linking form and meaning~\cite{chen2017phonetic,savva2014transphoner}.
\par Our work is motivated by the benefits of the keyword method and the persistent gap between its empirically validated potential and its challenging practical application.


\subsection{Computational Enhancements in Keyword Method}

\par Researchers have proposed various enhancements to the keyword method to address its practical challenges. Many of these efforts focus on augmenting learner-generated outcomes to strengthen dual-coding, such as incorporating image generation through multimodal large models to convert mental imagery into visual stimuli, thereby enhancing retention~\cite{attygalle2025text,weerasinghe2022vocabulary}. However, these methods still require learners to independently complete the creative tasks, offering minimal scaffolding for the cognitively demanding generative process. 
\par To alleviate this cognitive load, other approaches have turned to end-to-end automation, where systems generate keywords and associations for learners~\cite{savva2014transphoner,lee2024exploring,lee2023smartphone,kang2025phonitale}. For instance, \textit{TransPhoner}~\cite{savva2014transphoner} identifies cross-lingual keywords semantically aligned with the target vocabulary, while Lee et al.~\cite{lee2024exploring} introduced an “overgenerate \& rank” framework that utilizes large language models to produce keyword candidates, which are then ranked using psychometric indicators like imageability. While such strategies streamline the learning process, they often reduce the impact of the generation effect.

\par In response to this gap, our work seeks to enhance the keyword method by providing learners with process guidance and interactive support, thereby maximizing the benefits of the generation effect and improving the efficiency and effectiveness of vocabulary acquisition.

\subsection{Creativity Support Tools}
\par In HCI research, Creativity Support Tools (CSTs) are designed to enhance creative cognition and user engagement by offering innovative ways to support ideation, conceptual organization, and iterative refinement~\cite{10.1145/3290605.3300619,10.1145/3491102.3501933,10.1145/3706598.3713381}. Such tools have been widely adopted in the context of language learning, including applications like semantic maps~\cite{article3,article10,margosein1982effects,heimlich1986semantic}, flashcards~\cite{mutar2024flashcard,kornell2009optimising,nugroho2012improving}, and language games~\cite{ismayilli2025impact,yip2006online,rojabi2022kahoot,10.1145/3613905.3648107}. Several CSTs specifically aim to support language acquisition through creative interaction~\cite{sanosi2018effect,duolingo}. For instance, \textit{Duolingo}~\cite{duolingo} incorporates gamified exercises, such as timed challenges and role-playing scenarios, to promote creative problem-solving and contextual language use. While these tools effectively support vocabulary organization and conceptual understanding by providing structural and visual prompts, they often offer only limited, one-off stimulation. As a result, learners still bear a significant cognitive load in generating high-quality keywords and constructing semantically rich associations on their own.

\par The integration of Multimodal Large Language Models (MLLMs) into CSTs has demonstrated strong potential for enhancing both creative direction and content generation~\cite{10.1145/3706598.3713935,10.1145/3706598.3713818,10.1145/3613904.3642185,10.1145/3706598.3713381}. For example, \textit{ProductMeta}~\cite{10.1145/3706598.3713935} facilitates metaphor-driven product ideation by guiding users through multimodal prompt refinement and visual-semantic alignment. While these tools effectively support high-quality creative output, they often prioritize end results over process, offering limited support for learner agency, exploration, and self-directed engagement. Recent studies have begun to investigate how co-led human–AI collaboration can be structured to better balance system guidance with user autonomy~\cite{mosqueira2023human,wu2022survey,10.1145/3706598.3713649}. For instance, Xu et al.~\cite{10.1145/3706598.3713649} examined varying degrees of AI involvement in design workflows, finding that mixed-initiative models can scaffold reflection without undermining user control. This approach of AI-integrated cognitive guidance has been applied within the field of learning science~\cite{xu2025advancing, do2025paige}. Xu et al.~\cite{xu2025advancing} investigated the use of AI to enhance problem-based learning in medical education by supporting diagnostic reasoning and guiding learners through clinical scenarios. Nevertheless, their interaction paradigms tend to be simple and rigid, restricting their ability to provide sustained assistance across the iterative and non-linear cognitive processes inherent to the keyword method. These limitations underscore the need for more adaptive, co-creative paradigms to better support the demands of techniques like the keyword method.

\par Our work introduces \textit{WordCraft}, an MLLM-powered creativity support tool that provides process-level creative support for co-creation while helping learners maintain self-awareness throughout the keyword-method workflow.

\section{FORMATIVE STUDY}
\par With institutional IRB approval, we conducted a formative study comprising two keyword-construction tasks and a focus group discussion. The study examined learners' cognitive workflows in applying the keyword method for vocabulary memorization, identified associated challenges, and derived design considerations for integrating MLLMs into keyword-method-based learning.

\subsection{Participants and Experimental Preparation}
\label{31}

\par We recruited 14 university students (L1-L14, 8 males, 6 females) via social media for a study on English as a Foreign Language (EFL) learners preparing for the TOEFL exam. Participants were proficient in phonetic spelling rules and had a basic understanding of the keyword method, receiving \$10 for completing the study. We also included two experienced English teachers (T1, T2) and two Human-Computer Interaction researchers (R1, R2). The teachers each had over 5 years of experience in vocabulary instruction, while the researchers specialized in AI for Education. Each was compensated \$30 for their involvement, which included observing participants' interactions and contributing to post-experiment discussions. A summary of all participants is provided in \cref{tab:participants}. 


\par We screened 2,768 TOEFL vocabulary words from the MRC Psycholinguistic Database\footnote{\url{https://websites.psychology.uwa.edu.au/school/mrcdatabase/}} and classified them into four categories based on syllable length and imageability: high-imageability short words, high-imageability long words, low-imageability short words, and low-imageability long words. Participants reviewed the list and selected 4 unfamiliar words from each category, resulting in 16 experimental words per participant. These words were divided into two groups for construction tasks and presented in randomized order.


\par A custom-built system was built for structured data collection and consistent task execution. In each trial, it presented the target word with its definition and pronunciation, aiding participants' understanding. The interactive interface allowed participants to record self-generated keywords and associations while guiding them through the workflow. An illustration of the interface is provided in Appendix \ref{fs_system}.

\begin{table}[h]
\vspace{-1mm}
\centering
\caption{Detailed information of the participants in formative study.}
\vspace{-1mm}
\label{tab:participants}
\begin{tabular}{ccccc}
\toprule
\textbf{ID} & \textbf{Role} & \textbf{Gender} & \textbf{Age} & \textbf{Frequency / Experience} \\
\midrule
L1 & Student & Male & 20 & Frequently \\
L2 & Student & Female & 22 & Occasionally \\
L3 & Student & Male & 19 & Sometimes \\
L4 & Student & Female & 21 & Frequently \\
L5 & Student & Male & 23 & Occasionally \\
L6 & Student & Female & 20 & Occasionally \\
L7 & Student & Male & 24 & Frequently \\
L8 & Student & Female & 22 & Sometimes \\
L9 & Student & Male & 25 & Occasionally \\
L10 & Student & Female & 23 & Rarely \\
L11 & Student & Male & 21 & Frequently \\
L12 & Student & Female & 23 & Occasionally \\
L13 & Student & Male & 24 & Sometimes \\
L14 & Student & Male & 22 & Rarely \\
\midrule
T1 & Teacher & Female & 35 & 5-10 years \\
T2 & Teacher & Female & 42 & >10 years \\
\midrule
R1 & Researcher & Female & 29 & 1-5 years \\ 
R2 & Researcher & Male & 36 & 5-10 years \\
\bottomrule
\end{tabular}
\vspace{-5mm}
\end{table}

\subsection{Procedure}

\begin{figure*}[h]
    \centering
    \includegraphics[width=\textwidth]{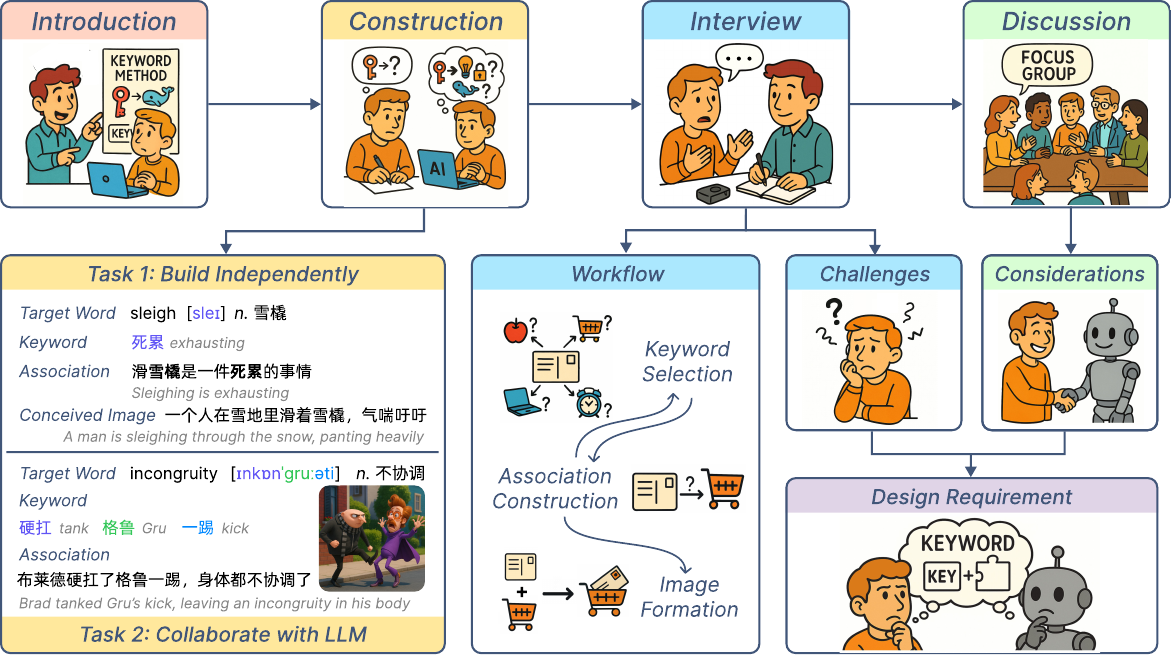}
    \caption{Procedure of the formative study. The study unfolded in four phases: 1) \textbf{Introduction}: Participants were introduced to the keyword method and familiarized with the system's operation. 2) \textbf{Construction}: Participants completed two keyword-construction tasks—first independently and then collaboratively with GPT-4o—while following a think-aloud protocol. 3) \textbf{Interview}: Semi-structured interviews were conducted to probe participants' cognitive processes and challenges. 4) \textbf{Discussion}: A focus group with learners, teachers, and researchers was held to gather insights on integrating MLLMs into the keyword method.}
    \label{fig:FormativeStudy}
\end{figure*}

\par The formative study was conducted into four phases: \textbf{\textit{Introduction}}, \textbf{\textit{Construction}}, \textbf{\textit{Interview}} and \textbf{\textit{Discussion}}. The overall process is illustrated in \cref{fig:FormativeStudy}.


\par During the \textbf{\textit{Introduction}} phase, participants were familiarized with the fundamental principles of the keyword method through examples and received instructions on the experimental system, allowing them to become familiar with the interface and procedural workflow. This phase lasted approximately 15 minutes.


\par In the \textbf{\textit{Construction}} phase, participants applied the keyword method to two target vocabulary sets, documenting their keywords, associations, and the images they conceived or generated. For the first set, participants independently created memory cues using the keyword method. This task was designed to examine learners' cognitive workflows in applying the technique and to identify challenges during unguided use. For the second set, participants interacted with GPT-4o via a \textit{ChatGPT}\footnote{\url{https://chatgpt.com/}} interface to construct memory cues. This task aimed to derive design considerations for effectively integrating MLLMs into keyword-method-based vocabulary learning. Throughout this phase, participants followed a think-aloud protocol~\cite{article4} to verbalize their thought processes, enabling researchers to capture their selection preferences, association strategies, and construction methods. This phase lasted approximately 40 minutes.

\par The \textbf{\textit{Interview}} phase involved semi-structured interviews following each construction task. The interviews aimed to explore participants' cognitive processes and the difficulties with the keyword method. Questions focused on three main areas: (1) explanations of the constructed keywords and associated scenarios, (2) cognitive strategies and decision-making during ideation, and (3) challenges or confusions experienced during the tasks. Each interview lasted approximately 15 minutes.

\par In the \textbf{\textit{Discussion}} phase, six participants (L1-L6) joined the teachers and researchers in a focus group to explore optimal ways of integrating MLLMs into the keyword method. Participants shared their perspectives on user experience and learning needs, while the expert panel provided insights from the perspectives of teaching practice and human-computer collaboration. The discussion lasted approximately one hour.

\subsection{Data Analysis}
\label{sec:data_analysis}
\par We adopted an inductive thematic analysis approach~\cite{Braun01012006} to systematically examine participant interviews and focus group discussions. The process began with automatic transcription using \textit{ClovaNote}\footnote{\url{https://clovanote.naver.com/}}, followed by meticulous manual review to ensure transcription accuracy. Two researchers then independently conducted open initial coding of the textual data. This independent coding phase generated a rich set of 83 initial codes (Researcher A) and 76 initial codes (Researcher B), reflecting diverse observations and interpretations of participants' cognitive processes. Following the independent coding, the researchers discuss to compare their coding outcomes, resolve discrepancies, and merge overlapping concepts. Through this collaborative reconciliation, the initial pool of codes was consolidated into a shared coding scheme comprising 14 categories and 52 codes. Based on this scheme, three overarching themes were developed. Through iterative comparison between the original data and the coded segments, these themes were continuously refined to ensure logical clarity, semantic coherence, and representativeness. The final codebook is provided in Appendix \ref{code_book}.

\subsection{Findings}
\par Based on our analysis, the findings are presented in three sections: \textit{Cognitive Workflow of the Keyword Method}, \textit{Challenges} and \textit{Considerations for Integrating MLLMs}.

\begin{table*}[h]
\caption{Identified challenges (C1–C6) and considerations (M1–M4), along with their corresponding design requirements.}
\label{tab:challenges}
\centering 
\begin{tabular}{llr}
\toprule
\textbf{Topics} & \textbf{Challenges \& Considerations} & \textbf{DRs}\\
\midrule
{\textbf{Generation Effect}} & \textbf{M1}. Over-reliance on MLLM reducing learner's generative thinking. & \textbf{DR1}\\
\midrule
\multirow{3}{*}{\textbf{Keyword Selection}} & \textbf{C1}. Identify phonetically similar and memorable keywords. & \textbf{DR2}\\
& \textbf{C2}. Understand and explore the meaning of the target word. & \textbf{DR3}\\
& \textbf{M2}. Meaningless keyword generation by MLLMs reducing quality and creativity.& \textbf{DR2}\\
\midrule
\multirow{3}{*}{\textbf{Association Construction}} & \textbf{C3}. Cognitive overload in association construction. & \textbf{DR4}\\
& \textbf{C4}. Create high-quality keyword-meaning associations. & \textbf{DR4}\\
& \textbf{M3}. Associative confusion due to information asymmetry.& \textbf{DR4}\\
\midrule
\multirow{3}{*}{\textbf{Image Formation}} & \textbf{C5}. Map keywords to visual elements in mental images. & \textbf{DR5}\\
& \textbf{C6}. Maintain recall path from image to target word. & \textbf{DR6}\\
& \textbf{M4}. Difficulty aligning generated visuals with learner's mental imagery.& \textbf{DR5}\\
\bottomrule
\end{tabular}
\end{table*}

\subsubsection{Cognitive Workflow of Keyword Method}
\par We identified a recurring cognitive workflow that learners typically follow when applying the keyword method, which progresses through three main phases:
\begin{enumerate}

\item \textbf{Searching for keyword candidates}: Learners segment the target word phonologically. They then brainstorm phonetically similar chunks while considering the target meaning to guide their thinking, ultimately selecting a suitable keyword.

\item \textbf{Constructing associations between keywords and meanings}: Learners connect the selected keyword with previously identified ones, collectively anchoring them to the meaning of the target word by forming a coherent idea, scene, or narrative. This process often involves revisiting earlier reflections on the word's semantic content and envisioning how the keywords might interact, overlap, or form an integrated conceptual structure in memory.

\item \textbf{Forming an integrated image}: Learners develop a vivid mental image that unites the keywords with the target meaning in a concrete scene or story, visualizing how the elements appear, interact, or unfold dynamically to make the association more distinctive and memorable.

\end{enumerate}

\par This cognitive workflow is cumulative and iterative, following a non-linear developmental trajectory. Learners frequently cycle between keyword searching and association construction, continuously discovering and incorporating new keyword elements while refining and optimizing the associative relationships between keywords and target word meanings. They often revisit earlier steps, adjusting choices or altering strategies when the resulting associations are less effective than anticipated. As L3 described: ``\textit{I had a few keywords, but when I tried to combine them into a scenario, it didn't quite fit, so I went back and reconsidered, replacing one of the keywords to make the whole association more appropriate.}'' Through this process, they build a comprehensive associative network that leads to the formation of a mental image.

\subsubsection{Challenges}
\par We categorized the six identified challenges (C1-C6) into three topics: \textit{Keyword Selection}, \textit{Association Construction}, and \textit{Image Formation} in \cref{tab:challenges}.

\par \textbf{Keyword Selection.} Learners consistently emphasized that selecting appropriate keywords is crucial but often challenging. A common challenge was identifying keywords that were both phonetically similar to the target word and easy to remember (\textbf{C1}). As L3 explained, ``\textit{When I heard `pique', I tried to find something that sounded similar in my language, but I couldn't come up with any word that felt concrete or usable.}''

\par Another challenge involved understanding and exploring the meaning of the target word (\textbf{C2}), especially for low-imageability items that offered little semantic grounding. Learners struggled to imagine contexts that could inspire keyword ideas or support meaningful associations, sometimes relying on vague interpretations or skipping this step entirely. As L4 noted, ``\textit{The word `resentment' felt kind of abstract—I looked it up, but the definition didn't really help. I couldn't think of where I'd use it or what it would look like, so I just told myself it's something emotional and left it at that.}''

\par \textbf{Association Construction.} Learners faced considerable challenges in constructing coherent associations between keywords and meanings. Many learners found it cognitively demanding to generate associations that were both coherent and memorable (\textbf{C3}). The complex landscape of semantic types and their interrelations introduced cognitive uncertainty, making it difficult for learners to determine a consistent or meaningful link. This challenge became more pronounced when the target word had more syllables, leading to a broader range of potential keyword combinations. As L10 described, ``\textit{For the word `metropolis', I picked the keyword ` \raisebox{-0.15em}{\includegraphics[page=1, height=0.9em]{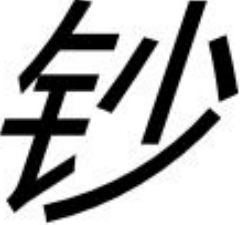}}' (money) from the  \includegraphics[page=1, height=0.7em]{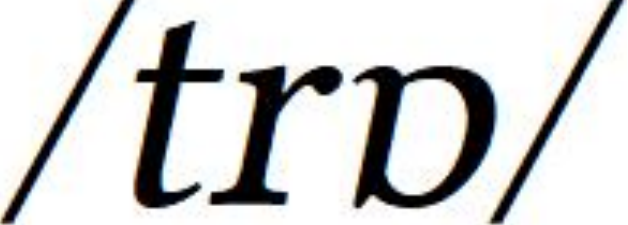} sound and the meaning, but there were so many possible ways it could connect to 'metropolis'. I wasn't sure how to move forward.}''


\par Another challenge involved the tendency to generate low-quality associations between keywords and meanings (\textbf{C4}). The quality of associations was often affected by factors such as the plausibility of the semantic connection, the degree of phonological similarity, the conceptual closeness between the keyword and the target word, and the imageability of both. Learners frequently lacked clear standards for evaluating these aspects, making it difficult to determine whether an association was meaningful or memorable. As L6 reflected, ``\textit{Sometimes I just made up a link between the keyword and the meaning because I wasn't sure what counted as a good one. It sounded okay at the time, but later I realized it didn't really help me remember the word.}''


\par \textbf{Image Formation.} Learners found it challenging to integrate keywords and meanings into coherent visual mental images. A central challenge was accurately mapping each keyword and its semantic connection to visual elements within a scene (\textbf{C5}). This process depended heavily on learners' imaginative capacity, and many found it challenging to construct vivid, well-structured images. When the concepts were abstract or the number of elements increased, the resulting imagery often became cluttered or disjointed, which reduced its effectiveness as a memory aid. As L8 shared, ``\textit{With too many elements to include—the keyword, the meaning, and other parts I'd come up with. I tried to imagine a scene, but it quickly got confusing. I couldn't tell what was supposed to be where, and it didn't help me remember anything}''

\par A further challenge was the disruption of the recall path linking the mental image, the keyword, the target word, and the constructed association (\textbf{C6}). Learners sometimes focused on recalling the image itself, but overlooked the reasoning that connected it to the keyword and the target word, disrupting the associative chain and weakening overall recall. As L12 recalled, \textit{``I got really immersed in the scene I imagined—it was a rapper bouncing on top of a building—but after just a short while, I couldn't remember what word it was for.'' (target word: skyscraper)}

\subsubsection{Consideration for MLLMs Integration}
\par We identified four key considerations (M1-M4) for integrating MLLMs into the keyword method process in \cref{tab:challenges}.


\par The primary concern from the focus group discussion was the risk of diminishing the generation effect (\textbf{M1}), as learners may become less engaged in the overall generative process. Although many participants acknowledged that MLLMs could occasionally produce suggestions with high perceived relevance to the target word, this often led students to bypass generative thinking. As L5 described, \textit{``I did consider letting the model guide me step by step to the answer, because thinking on my own was just too hard. Once I saw it could give a usable suggestion right away, I couldn't help going straight to it.''} Teachers emphasized the importance of maintaining sufficient cognitive effort and time when using the keyword method. They encouraged learners to develop their own understanding of the target word as part of the generative process. As T1 noted, \textit{``I observed that when learners used MLLMs, they spent significantly less time thinking, and the keywords they generated were noticeably less personalized than those they came up with on their own. That really concerned me.''}


\par Additionally, \textbf{M2} notes that MLLMs often generate meaningless keywords during the \textbf{keyword selection} phase, resulting in outcomes that disappoint learners. As L5 noted, \textit{``Sometimes the keywords generated by the MLLM weren't even words I've seen in my native language, which really diminished its ability to spark creativity.''} In addition, the lack of fine-grained control over the generation process hindered learners' ability to obtain desired keywords. As L1 described, \textit{``When I told the model it could sacrifice some phonetic similarity in favor of stronger semantic connection, it gave me keywords that had no phonetic connection to the target word at all.''}


\par Another consideration, \textbf{M3}, is the associative confusion created by MLLM outputs during the \textbf{association construction} phase. Learners often struggle to articulate their thoughts clearly, creating an information gap between them and the model. As L2 mentioned, \textit{``It's hard for me to explain to the MLLMs the logic behind my keyword selection, especially when several elements are involved. I feel like I can't explain it clearly.''} Due to this barrier, MLLMs cannot accurately understand the learner's intended logic when selecting keywords, often leading to overly divergent and incoherent associative content. As described by R2, ``While these associations are imaginative and diverse, they don't align with the learner's train of thought, which instead increases cognitive load and unnecessary complexity.''


\par During the \textbf{image formation} phase, learners often relied on elaborate textual descriptions to convey their mental imagery to MLLMs. They emphasized the difficulty of generating visual content that meets their expectations (\textbf{M4}). As L6 noted, \textit{``When I tried to map `metropolis' into an image, I told the model I wanted the background filled with skyscrapers. But it also added many office workers to the scene, which wasn't what I intended. It took too many iterations to get what I had in mind.''} R1 referred to this as a common issue in the textual-centric interaction paradigm~\cite{shi2025brickify, lin2025sketchflex}. T2 also expressed concern over the phenomenon, noting: \textit{``When the generated image does not match the mental expectation, it means the mapping between the keywords and associations is inaccurate, which could affect memory outcomes.''}

\subsection{Design Requirement}
\par In response to the identified challenges and considerations, we propose six design requirements to support learners' cognitive engagement and enhance the effectiveness of the keyword method with MLLM integration.

\par \textbf{DR1. Preserve Cognitive Effort in the Keyword Method Process (M1).} The system should help learners maintain active cognitive engagement throughout the keyword method process. Rather than replacing their cognitive efforts, MLLMs should function as scaffolds that promote deeper involvement in knowledge construction. The design should offer cognitive assistance while preserving the effortful processing necessary to sustain the generation effect.


\par \textbf{DR2. Enhance Relevance and Usability of Generated Keywords (C1, M2).} The system should offer semantically relevant and memorable keyword candidates to minimize ineffective exploration during the keyword selection phase, ensuring high usability and creativity in MLLM outputs to better support learners' constructive needs.


\par \textbf{DR3. Facilitate Contextual Understanding of Target Words (C2).} The system should support learners in deepening their understanding of the target word's semantics by helping them identify and explore concrete contexts, which can stimulate associations and enhance vocabulary connections. By enhancing semantic comprehension. This supports learners in selecting appropriate keywords, thereby laying a solid foundation for subsequent association construction.


\par \textbf{DR4. Support Systematic and Structured Association Construction (C3, C4, M3).} The system should provide structured, goal-oriented support mechanisms to reduce cognitive load during association construction. It should help develop clear and logically coherent associative links between keywords and target words, thereby mitigating confusion and information asymmetry caused by the divergent or ambiguous outputs of the model.

\par \textbf{DR5. Provide Visual-Centric Generation Aligned with Learners' Mental Imagery (C5, M4).} The system should support learners in generating mental imagery through a visually centered approach, enabling more precise control over image formation. Furthermore, the generation process should allow for the clear mapping of keywords, target words, and associative content onto the image, ensuring that each semantic element is accurately represented in the visual output. This enhances the structural clarity of the imagery and its effectiveness as a memory aid.

\par \textbf{DR6. Maintain Meaningful Recall Pathways (C6).} The system should assist learners in establishing and maintaining a clear recall path among keywords, associative content, and target words, providing explicit structural cues for subsequent image generation. This enables learners to reflect on and evaluate the generated imagery with reference to the intended associative framework.

\section{WORDCRAFT}
\par Guided by our defined design requirements, we developed \textit{WordCraft}, an interactive creativity support tool that provides process-level assistance for L2 learners using the keyword method. By balancing learner agency with reduced cognitive load, \textit{WordCraft} enhances the overall effectiveness of vocabulary memorization. For easier understanding, we attached a usage scenario in \cref{usage_scenario}.

\begin{figure*}[h]
    \centering
    \includegraphics[width=\textwidth]{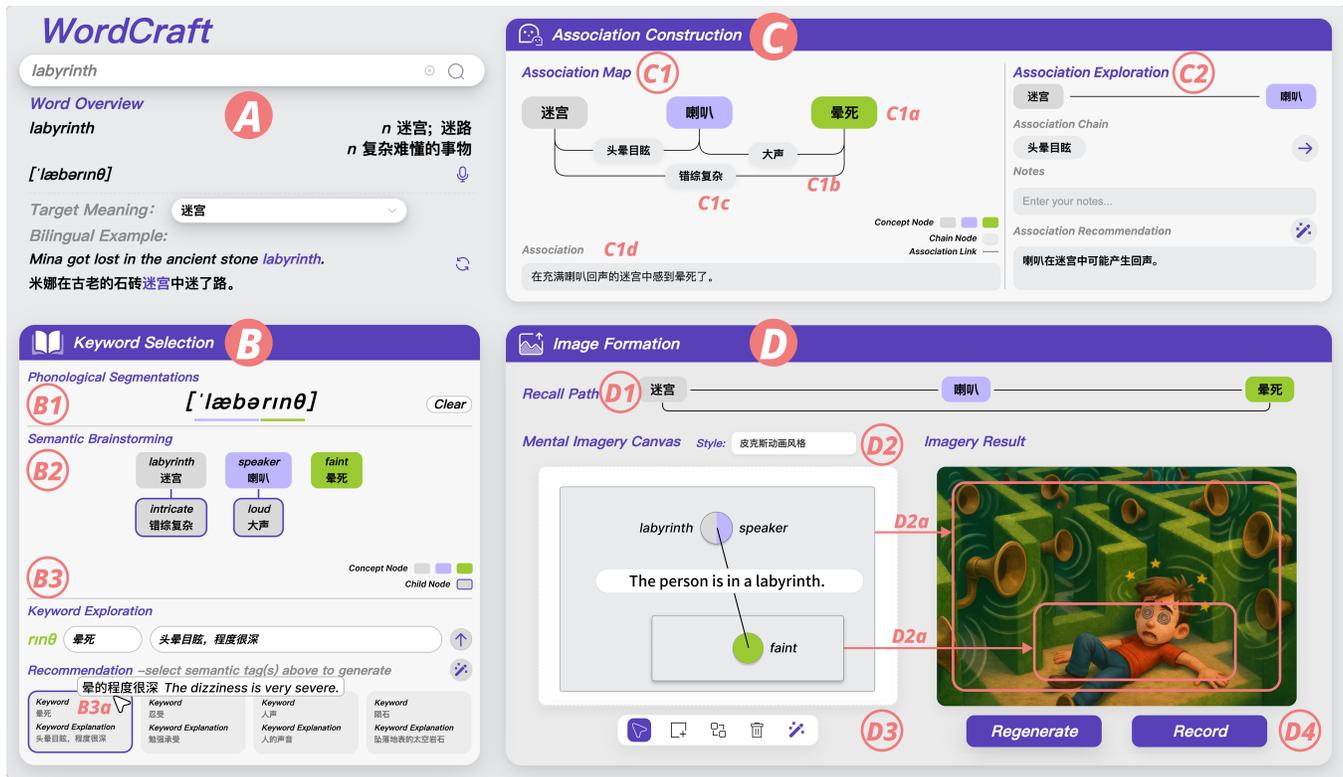}
    \caption{\textit{WordCraft} consists of four primary views: (A) \textbf{Word Overview}, which provides key information about the target word; (B) \textbf{Keyword Selection}, which supports keyword identification and fosters deeper comprehension; (C) \textbf{Association Construction}, which connects the selected keywords to the target meaning; and (D) \textbf{Image Formation}, which translates the keywords and meanings into visual elements to produce a final image.}
    \label{fig:system}
    \vspace{-3mm}
\end{figure*}

\subsection{System Overview}
\par \textit{WordCraft} is designed to assist L2 learners in applying the keyword method for vocabulary memorization. The process begins with users accessing a \textbf{Word Overview} (\cref{fig:system} A), where they can review essential information, such as definitions, phonetic transcription, and example sentences. From this overview, users select the target meaning they wish to memorize, which anchors the subsequent learning workflow. To scaffold this process, \textit{WordCraft} integrates MLLMs with interactive interfaces, featuring three core components: \textbf{Keyword Selection} (\cref{fig:system} B), \textbf{Association Construction} (\cref{fig:system} C), and \textbf{Image Formation} (\cref{fig:system} D).

\par \textbf{Keyword Selection} (\cref{fig:system} B) assists learners in identifying phonologically similar and memorable keyword candidates (\textbf{DR2}), while providing contextual cues to deepen the semantic understanding of the target word (\textbf{DR3}). \textbf{Association Construction} (\cref{fig:system} C) guides learners in linking the selected keywords to the target meaning through structured association-building, thereby reducing cognitive overload and fostering logically coherent connections (\textbf{DR4}). \textbf{Image Formation} (\cref{fig:system} D) provides, a visual-centric interaction mode, enabling learners to map their mental images to concrete visual elements and freely control their placement, descriptions, and interrelationships on the canvas. Simultaneously, it establishes explicit recall pathways that serve as structural cues for future image generation, ensuring the visual output aligns with the intended associative framework (\textbf{DR6}).

\par The three components are designed to be interconnected and adaptive: outputs from one stage directly inform the next, and revisions in later stages can trigger adjustments in earlier ones, ensuring a fluid, learner-centered workflow. Unlike fully automated generation, \textit{WordCraft} does not replace learner thinking; rather, it provides structured guidance and heuristic prompts that support cognitive control and active participation throughout the generative process. By encouraging engagement at every stage of content construction, the system preserves the generation effect (\textbf{DR1}).

\subsection{Keyword Selection}
\subsubsection{Phonological Segmentations}
\par Learners can freely segment International Phonetic Alphabet (IPA) transcriptions by brushing over the phonetic symbols (\cref{fig:system} B1). Segmented phonemes are visually indicated by color-coded lines beneath the IPA representation. Segments can be removed or adjusted via the ``clear'' button, while the system retains the data for deleted segments for future reference.

\subsubsection{Semantic Brainstorming}
\par The target meaning and selected keyword are positioned at the root of a semantic tree (\cref{fig:system} B2). Each node represents a semantic concept, accompanied by contextual cues and their corresponding translations. Learners can create child nodes by interacting with existing ones, either entering their own ideas (\textbf{DR1}) or requesting system suggestions (\textbf{DR3}). This design encourages active engagement while limiting cognitive engagement. To prevent excessive expansion, the semantic tree is restricted to two levels of depth.

\subsubsection{Keyword Exploration}
\par Learners can explore potential keywords by interacting with the color-coded segments (\cref{fig:system} B3). They are encouraged to provide their own keywords and explanations based on personal preferences (\textbf{DR1}). In the Semantic Brainstorming area (\cref{fig:system} B2), learners may select preferred semantic links and receive keyword suggestions presented as keyword cards. Clicking the \raisebox{-1ex}{\includegraphics[height=1.5em]{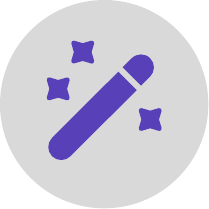}} button refreshes these suggestions, presenting four new cards at a time (\textbf{DR2}). Each keyword card includes: 1) \textbf{Keyword}; 2) \textbf{Keyword Explanation}: a brief phrase providing additional context for the recommended keyword; 3) \textbf{Reasoning}: a natural language explanation accessible on hover, detailing why the keyword is recommended (\cref{fig:system} B3a). When a learner selects a keyword, the information is automatically synchronized across both the Semantic Brainstorming area and the Association Construction components, maintaining continuity throughout the system.

\subsection{Association Construction}
\subsubsection{Association Map}
\par Learners are tasked with constructing a complete association using natural language. \textit{WordCraft} provides structured guidance within the Association Map (\cref{fig:system} C1), supporting learners in progressively refining their ideas (\textbf{DR4}). The map consists of four types of elements:
\begin{itemize}
  \item \textbf{Concept Node}: A large node representing the core elements of the association, including the target meaning or keyword. Its background color corresponds to the source dimension, facilitating rapid visual mapping and cognitive linkage (\cref{fig:system} C1a).
  \item \textbf{Association Link}: Connects two Concept Nodes to denote a relationship between them (\cref{fig:system} C1b). 
  \item \textbf{Chain Node}: A smaller node attached to an Association Link, representing the associative pathway between two Concept Nodes. Its initial content is derived from the learner's preferred semantic chain in the Keyword Selection component (\cref{fig:system} C1c).
  \item \textbf{Note}: Presented in natural language and attached to an Association Link. Learners can add supplementary explanations, reflections, or contextual cues to enrich the association  (\cref{fig:system} C1d).
\end{itemize}

\subsubsection{Association Exploration}
\par Learners can edit Chain Nodes and Notes by selecting any Association Link and interacting with the right-side panel (\cref{fig:system} C2). They can also click the \raisebox{-1ex}{\includegraphics[height=1.5em]{Figure/llm.pdf}} button to receive heuristic suggestions for potential relationships between two Concept Nodes. These suggestions are closely aligned with the learner's existing input but are designed to be indirect and inspirational rather than providing direct answers. This approach encourages learners to expand or reorganize their associations, fostering reflection and deeper cognitive engagement (\textbf{DR1}).

\subsection{Image Formation}
\subsubsection{Recall Path}
\par Learners use a Recall Path (\cref{fig:system} D1) to quickly review their overall cognitive process. Since the Association Map contains dense information, a simplified graph displaying only Concept Nodes and Association Links serves as the Recall Path, providing guidance and reference during mental imagery mapping (\textbf{DR6}).

\begin{figure*}[h]
    \centering
    \includegraphics[width=\textwidth]{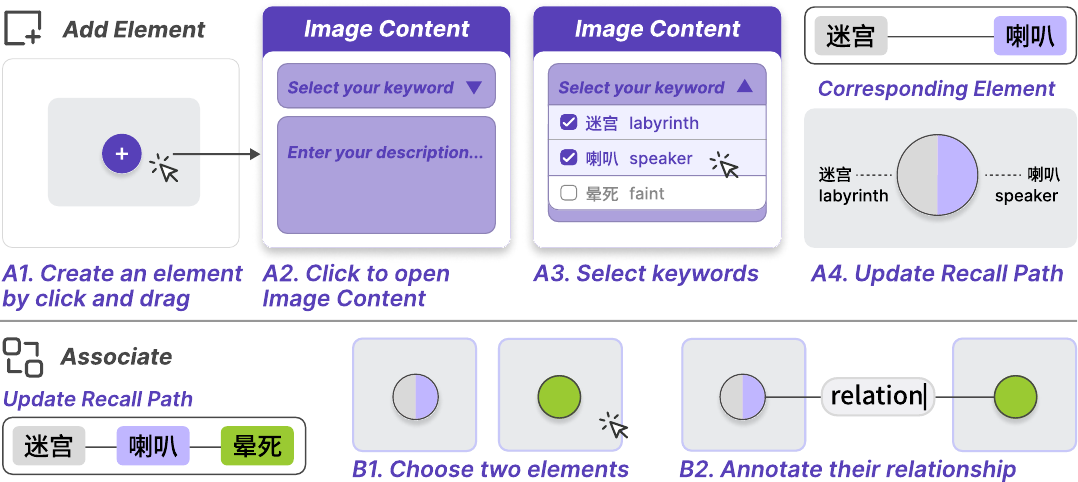}
    \caption{(A) \textbf{Add Element} \raisebox{-1ex}{\includegraphics[height=1.5em]{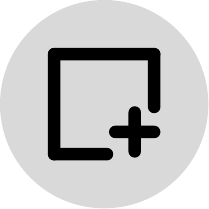}} and (B) \textbf{Associate} \raisebox{-1ex}{\includegraphics[height=1.5em]{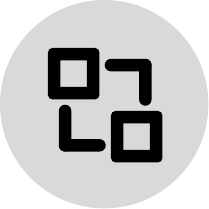}} operations. (A1) Create an element using a click-and-drag action. (A2-A3) Click the plus button to open \textit{Image Content} dialog, where users can select keywords and provide descriptions. (A4) The Recall Path the updates accordingly, and the element is visualized as a pie chart. (B1-B2) Once two elements are added, select them and annotate their relationship.}
    \label{fig:operations}
\end{figure*}

\subsubsection{Mental Imagery Canvas}
\par \textit{WordCraft} provides a visually centered interface that supports learners in progressively constructing mental imagery on the canvas (\cref{fig:system} D2, \textbf{DR5}), ensuring alignment with the final outcome and maintaining cognitive engagement throughout the process (\textbf{DR1}). The system offers five fundamental operations (\cref{fig:system} D3) to facilitate this construction.
\begin{itemize}
    \item \textbf{Select} \raisebox{-1ex}{\includegraphics[height=1.5em]{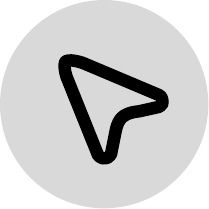}}: Click to select an element, then drag to reposition, resize, or  edit its content.
    \item \textbf{Add Element} \raisebox{-1ex}{\includegraphics[height=1.5em]{Figure/add.pdf}}: Click and drag on the canvas to add an element (\cref{fig:operations} A1). A central plus button opens the \textit{Image Content} dialog (\cref{fig:operations} A2), allowing learners to assign concept tags and provide descriptive details (\cref{fig:operations} A3). Elements are visualized in a pie chart with color encoded according to their concept nodes. Such nodes in the Recall Path update accordingly (\cref{fig:operations} A4).
    \item \textbf{Associate} \raisebox{-1ex}{\includegraphics[height=1.5em]{Figure/associate.pdf}}: Draw connections between any two elements and annotate the relationship (\cref{fig:operations} B1-2). Existing Association Links in the Recall Path must be mapped on the canvas. Descriptions can be omitted if spatial positioning sufficiently conveys the relationship. Updates are synchronized with the Recall Path.
    \item \textbf{Delete} \raisebox{-1ex}{\includegraphics[height=1.5em]{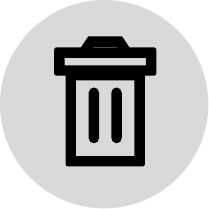}}: Remove any element or link from the canvas.
    \item \textbf{Inspire} \raisebox{-1ex}{\includegraphics[height=1.5em]{Figure/llm.pdf}}: Click the star button to receive heuristic, text-based suggestions for selected concepts or relations. Accepted suggestions update both the element content and the Recall Path.
\end{itemize}

\subsubsection{Imagery Result}
\par Learners can generate and save final mental images constructed on the Mental Imagery Canvas, including the element descriptions, placement, and relationships (\cref{fig:system} D2a). Clicking the \raisebox{-0.7ex}{\includegraphics[height=1.2em]{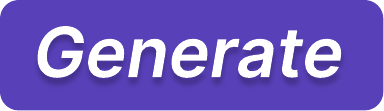}} button creates a visual representation (\cref{fig:system} D4) based on the current canvas configuration, while the \raisebox{-0.7ex}{\includegraphics[height=1.2em]{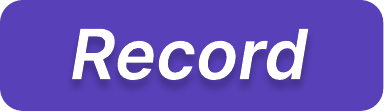}} button saves the learning outcome for the target word. Each record includes the target word, keywords, associations, generated image, and time spent, capturing the learner's cognitive trajectory throughout the vocabulary learning process and providing a foundation for subsequent review (see \cref{fig:examples}-\textit{WordCraft}).

\subsection{Implementation}

\par \textit{WordCraft} is implemented using modern web technologies, featuring a Vue.js\footnote{\url{https://vuejs.org/}} frontend and a Flask\footnote{\url{https://flask.palletsprojects.com/}} backend that provides API services. For keyword selection and association construction, the system leverages OpenAI's GPT-4o model for language-based generation, while image formation is powered by the multimodal GPT-Image-1 model. All interactions with these models are conducted through the OpenAI API\footnote{\url{https://platform.openai.com/}}. Implementation details and complete prompts are provided in Appendix \ref{detail} and \ref{app:prompt}.

\section{EVALUATION}
\par To evaluate \textit{WordCraft}, we conducted two complementary user studies. Study 1 was a between-subjects experiment ($N=48$) that assessed the system's effectiveness and usability relative to two baselines: a GPT-4o-based chat interface, representing the use of the keyword method through unstructured AI support, and a traditional flashcard interface, reflecting the prevailing non-generative learning approach. Study 2 was a within-subjects experiment ($N=20$) designed to further investigate the system's underlying cognitive mechanism, specifically examining whether \textit{WordCraft} preserves the generation effect.

\subsection{Study 1: Effectiveness and Usability of \textit{WordCraft}}

\subsubsection{Baselines}
\par We implemented two baseline conditions to evaluate \textit{WordCraft} from distinct perspectives:

\paragraph{GPT-4o-based Chat Interface}
\par For the first comparison, participants applied the keyword method using a GPT-4o-based chat interface (Appendix \ref{baseline_system}(a)). For each target word, the interface presented its definition and pronunciation, and participants recorded their selected keywords, associations, and mental imagery. Learners could interact with the model in an open-ended manner, requesting support such as keyword suggestions, associative ideas, or imagery descriptions. The system provided responses on demand, without enforcing any structured workflow.
\par This baseline was chosen for two main reasons. First, GPT-4o is a widely accessible and powerful LLM, commonly used by learners for vocabulary practice, making it a realistic comparator to \textit{WordCraft}. Second, although GPT-4o leverages the same underlying model capabilities, it provides only unstructured, on-demand interactions, making it suitable for comparison with \textit{WordCraft}'s structured, in-process guidance.

\paragraph{Traditional Flashcard Interface}
\par To benchmark against a traditional learning method, we developed a custom digital flashcard interface designed to mimic the core functionalities of mainstream vocabulary applications such as \textit{Quizlet}\footnote{https://quizlet.com/} (Appendix \ref{baseline_system}(b)). The interface employed a standard two-sided card metaphor: the front displayed the target word, while the back provided comprehensive lexical information, including the definition, IPA transcription, audio pronunciation, and an illustrative example sentence. Participants could interact with the system by clicking to ``Flip'' the card for self-testing and using navigation buttons to move between words freely.
\par We selected this baseline because the digital flashcard format represents one of the most familiar and widely adopted paradigms for vocabulary learning, ensuring a comfortable and intuitive experience for unassisted self-study. Further, we excluded spaced repetition algorithms~\cite{Nakata_2008,10.1145/2939672.2939850} typically paired with flashcards, as such scheduling strategies apply equally to \textit{WordCraft}. Instead, we encouraged learners to engage in spontaneous review, a design choice that allows us to compare the underlying learning strategies themselves.

\subsubsection{Participants}

\par We recruited 48 university students (24 female, 24 male; aged 19--24) through social media channels (Appendix \ref{participants-study1}). All participants were EFL learners preparing for the TOEFL exam. They reported familiarity with phonetic spelling rules and some prior exposure to the keyword method, though without extensive or systematic practice. 
To ensure balanced conditions, participants were stratified by their self-reported prior familiarity with the keyword method and then assigned to three groups of 16 each~\cite{kernan1999stratified}: the \textit{WordCraft} group, the \textit{GPT-4o} group, and the \textit{Flashcard} group. Each participant received a compensation of \$10 per hour.

\subsubsection{Procedure}
\label{sec:study1_prep}
\par Study 1 include five stages: \textbf{Preparation}, \textbf{Introduction}, \textbf{Learning}, \textbf{Questionnaires \& Interview}, and \textbf{Testing}.

\par \textbf{Preparation (15 min).} Before each session, participants reviewed and signed an informed consent form and received a brief explanation of the study's purpose. They then completed a demographic questionnaire covering age, gender, language learning background, and prior familiarity with the keyword method. A subsequent word-screening task presented each candidate word with its definition and pronunciation. Participants indicated one of three options: ``\textit{I know this word and can use it}'', ``\textit{I recognize this word but am unsure of its exact meaning,}'' or ``\textit{I do not know this word}.'' Only words unanimously rated as unfamiliar were retained, while excluded items were replaced with matched alternatives from the pool. This ensured that all final materials contained appropriately challenging, unfamiliar vocabulary. 8 words were selected in total, balanced for syllable length and imageability across groups. This set size was determined based on two key considerations. First, we followed the experimental design of recent research by Attygalle et al.~\cite{attygalle2025text}, which employed vocabulary sets ranging from 5 to 10 words. 

Second, using \textit{WordCraft} for a large vocabulary set in one session is very time-consuming and cognitively intensive. This could lead to fatigue effects~\cite{mcmorris2018cognitive} that make it difficult to accurately measure how effective the scaffolding mechanisms actually are.

\par \textbf{Introduction (20 min for the \textit{GPT-4o} group, 20 min for the \textit{Flashcard} group, 25 min for the \textit{WordCraft} group).} In the general introduction (10 min), we outlined the study procedure, introduced the research background and objectives, and presented the target vocabulary set. Participants were shown illustrative examples of how the keyword method aids memory through keyword generation, associative links, and image formation. Tool-specific orientation then followed. The \textit{WordCraft} group received a 15-minute guided walkthrough of the interface, including its three main components supporting the keyword method, and completed a short practice task using non-target words to ensure familiarity with the tool and allow questions to be addressed before the main learning phase. The \textit{GPT-4o group} received a 10-minute session focused on practicing interactions within the chat interface, which ensured they could effectively apply the keyword method in an open-ended conversational setting. The \textit{Flashcard} group received a 10-minute introduction to the digital interface and was instructed to utilize the standard active recall strategy~\cite{azabdaftari2012comparing}: viewing the target word, attempting to recall its definition, and verifying the answer by flipping the card, repeating this process as needed.

\par \textbf{Learning (50 min).} Participants were assigned the same set of eight unfamiliar words. The \textit{WordCraft} and \textit{GPT-4o} groups studied them by generating keywords, constructing associations, and forming images, while the \textit{Flashcard} group employed standard active recall. Participants were instructed to conclude the learning session voluntarily upon achieving a subjective sense of mastery. The order of word presentation was counterbalanced across participants using a partial Latin square design to mitigate order effects, and the time spent on each word was automatically recorded.

\par \textbf{Questionnaires \& Interview (25 min).} Participants first completed 7-point Likert questionnaires (10 min) to assess their learning experience. This was followed by a 10-minute semi-structured interview where participants reflected on their assigned learning strategies, provided concrete examples, and shared overall impressions of the system. Finally, they offered suggestions for future improvements. These insights complemented the quantitative data, clarifying the strengths and limitations of \textit{WordCraft} compared to the baselines.

\par \textbf{Testing (20 min).} Participants completed an immediate recall test requiring them to provide the meaning of each studied word. For the \textit{WordCraft} and \textit{GPT-4o} groups, participants were additionally asked to recall the corresponding keywords, associations, and images they had created. A delayed recall test using the same procedure was conducted seven days later.

\subsubsection{Measurements}
\par For the quantitative evaluation, we focused on three core dimensions: \textbf{Perception}, \textbf{Features}, and \textbf{Learning Outcomes}.

\par \textbf{Perception.} This dimension captured participants' overall impressions of system usability, creative support, and cognitive workload. We employed three standardized instruments: the \textit{System Usability Scale (SUS)}~\cite{bangor2008empirical}, the \textit{Creativity Support Index (CSI)}~\cite{cherry2014quantifying}, and the \textit{NASA Task Load Index (NASA-TLX)}~\cite{hart2006nasa}. The SUS assessed general usability, the CSI measured the extent of support for creative engagement, and the NASA-TLX evaluated perceived cognitive effort, which was. Notably, while the SUS and NASA-TLX were completed by all participants, the CSI was restricted to the generative \textit{WordCraft} and \textit{GPT-4o} conditions. Together, these measures provided a comprehensive view of participants' subjective experiences, satisfaction, and task workload.

\par \textbf{Features.} This dimension evaluated specific system capabilities designed to address the identified challenges (Cs) and considerations (Ms) of the keyword method, as summarized in the design requirements (DRs) in \cref{tab:challenges}. It was divided into three aspects: \textit{generation effect}, assessing whether the system maintained active learner engagement (M1; DR1); \textit{functionality}, including support for keyword selection (C1, C2; M2; DR2, DR3), association construction (C3, C4; M3; DR4), and image formation (C5, C6; M4; DR5, DR6); and \textit{iterative workflow}, reflecting participants' perceptions of how smoothly the system enabled revision, integration, and cross-stage use.

\par \textbf{Learning Outcomes.} This dimension assessed the effectiveness of participants' engagement with the learning process. Three measures were used: \textit{completion time}, recording the duration required to finish learning tasks; \textit{result scoring}, evaluating generated cues across Satisfaction, Clarity, Memorability, Novelty, and Relevance \& Coherence; and \textit{recall performance}, including immediate recall counts after learning and delayed recall counts seven days later. To further explore the durability of the mnemonic traces, we also asked participants in the \textit{WordCraft} and \textit{GPT-4o} groups to recall the specific keywords, associations, and images they had created, providing deeper insight into the stability of their learning strategy.


\par In addition, we conducted \textbf{semi-structured interviews} to gather in-depth qualitative feedback. Participants reflected on their use of the tool, discussed strengths and limitations, and offered suggestions for improvement. These qualitative insights complemented the quantitative data, providing richer context to participants' experiences.

\subsection{Study 2: Preserving the Generation Effect}

\subsubsection{Participants}
\par The participant profile was consistent with Study~1. We recruited 20 university students (11 female, 9 male; aged 19--24) through social media channels (Appendix \ref{participants-study2}). Participants were randomly divided into two groups (A and B) and each received a compensation of \$10 per hour.

\subsubsection{Procedure}

\par The early phases in Study 2 largely followed the procedures established in Study 1. Participants completed a \textbf{Preparation (15 min)} phase in which 16 unfamiliar words were selected, followed by an \textbf{Introduction (25 min)} phase that familiarized them with the overall objective and the use of \textit{WordCraft}.

\par In the \textbf{Learning (60 min)} phase, participants in Groups A and B were each assigned a distinct set of eight unfamiliar words. The learning session consisted of two consecutive phases. In the first phase, they used \textit{WordCraft} to generate keywords, associations, and images for their assigned words. In the second phase, groups exchanged word sets, such that Group~A studied words initially assigned to Group~B, and vice versa. For each word, cues were randomly selected from those generated by different peers in the first phase, ensuring participants did not repeatedly receive materials from the same contributor. The session concluded when participants reported completing memorization or when the time limit was reached. Finally, during the \textbf{Testing (20 min)} phase, participants completed an immediate recall test after each of the two learning sessions. They were presented with the studied words and asked to write down their meanings without assistance. If they could not respond, they were provided with the keyword as a cue and asked to attempt recall again. This procedure was conducted separately for both the self-generated and peer-generated conditions. To assess longer-term retention, a delayed recall test was administered seven days later, following the same procedure with both free recall and prompted recall. All tests were administered individually and under consistent time constraints.

\subsubsection{Measurements}

\par To assess the preservation of the generation effect, we measured recall performance both immediately after the learning session and after a seven-day delay. At each time point, we employed two tasks to gauge memory strength. First, in a free recall task, participants were asked to write down the meaning of each target word without any assistance. Second, if a participant failed to recall the meaning, they were provided with the keyword as a cue and asked to attempt recall again in a prompted recall task. The total number of correctly recalled definitions was recorded for each task, allowing us to compare the effectiveness of self-generated versus peer-generated cues on both immediate and long-term retention.

\subsection{Data Analysis}

\par For the quantitative analysis, recall performance in both studies was measured as a binary outcome (0 = incorrect, 1 = correct). In Study 1, recall performance and questionnaire ratings were compared using the Mann–Whitney U test\cite{mcknight2010mann}. All reported $p$-values were adjusted for multiple comparisons using the Bonferroni correction. No inferential comparisons were conducted among the baseline conditions themselves. Feature ratings specific to \textit{WordCraft} were summarized descriptively without cross-group comparison. In Study 2, recall performance was compared between self-generated and peer-generated cue conditions using the Wilcoxon signed-rank test\cite{capanu2006testing}.

\par The qualitative feedback from Study~1 was analyzed using the same inductive thematic analysis workflow described in \autoref{sec:data_analysis}. Briefly, two researchers independently conducted open coding of the qualitative responses, cross-checked their codes, and resolved discrepancies through discussion before organizing the merged code set into higher-level themes. These themes were used to interpret participants' perceptions of the system features and to contextualize the quantitative outcomes.

\begin{table*}[h]
  \centering
  \caption{Statistical comparisons on the System Usability Scale (significance: * $p<.050$, ** $p<.010$, *** $p<.001$).}
  \label{tab:user-feedback-sus}
  \setlength{\tabcolsep}{6pt}
  \renewcommand{\arraystretch}{1.18}

  \begin{tabular}{l
                  c c c
                  S[scientific-notation = true, round-mode=places, round-precision=2, table-format=1.2e-1] c
                  S[scientific-notation = true, round-mode=places, round-precision=2, table-format=1.2e-1] c}
    \toprule
    \multirow{2.5}{*}{\textbf{Dimension}} &
      \multicolumn{3}{c}{\textbf{Mean/S.D.}} &
      \multicolumn{2}{c}{\textbf{WordCraft vs. GPT-4o}} &
      \multicolumn{2}{c}{\textbf{WordCraft vs. Flashcard}} \\
    \cmidrule(lr){2-4} \cmidrule(lr){5-6} \cmidrule(lr){7-8}
    & \textbf{WordCraft} & \textbf{GPT-4o} & \textbf{Flashcard} &
      \textbf{$p$-value} & \textbf{Sig.} &
      \textbf{$p$-value} & \textbf{Sig.} \\
    \midrule
    Frequency      & 6.06/0.77 & 3.50/2.00 & 5.19/0.98 & 0.0012668332  & \textbf{**} & 0.0148 & \textbf{*}    \\
    Ease to use    & 5.88/0.62 & 4.62/1.59 & 6.31/0.70 & 0.0455697620  & \textbf{*}   & 0.1412 &     \\
    Functionality  & 6.25/0.93 & 3.38/0.89 & 3.31/1.30 & 0.0000029230  & \textbf{***} & 0.0000145206 &  \textbf{***}   \\
    Confidence     & 6.12/0.81 & 4.31/1.45 & 3.12/1.50 & 0.0014604012  & \textbf{**} & 0.0000229274 &  \textbf{***}   \\
    Learnability   & 5.62/1.09 & 5.88/0.81 & 6.44/0.63 & 1.1348627510  &             & 0.0561382464 &     \\
    \bottomrule
  \end{tabular}

\end{table*}

\section{RESULTS OF STUDY 1}

\subsection{Perception}
\par We evaluated \textit{WordCraft}'s perception relative to the baselines across three dimensions: \textit{system usability}, \textit{creative support}, and \textit{cognitive effort}.

\subsubsection{System Usability}
\par System usability~\cite{bangor2008empirical} was assessed across five metrics: \textit{frequency}, \textit{ease of use}, \textit{functionality}, \textit{confidence}, and \textit{learnability}. Detailed results are presented in \cref{tab:user-feedback-sus}.

\par Participants indicated a significantly higher willingness to use \textit{WordCraft} frequently (M = 6.06, SD = 0.77) compared to the GPT-4o baseline (M = 3.50, SD = 2.00) and the Flashcard baseline (M = 5.19, SD = 0.98). For \textit{ease of use}, participants rated \textit{WordCraft} at M = 5.88 (SD = 0.62), higher than the GPT-4o baseline (M = 4.62, SD = 1.59), while the Flashcard baseline achieved near-ceiling scores (M = 6.31, SD = 0.70) due to its inherent simplicity. P4 noted, \textit{``\textit{WordCraft}'s interaction design is very thoughtful, especially in the keyword selection stage, where the interface clearly guided me on where to expand thoughts in the Semantic Tree and map them to keywords, requiring just a click for help.''} In terms of \textit{functionality}, \textit{WordCraft} was perceived as offering more comprehensive and better-integrated features (M = 6.25, SD = 0.93), while the baselines were rated lower (GPT-4o: M = 3.38, SD = 0.89; Flashcard: M = 3.31, SD = 1.30). Participants also reported greater \textit{confidence} when applying the keyword method with \textit{WordCraft} (M = 6.12, SD = 0.81) than with the GPT-4o baseline (M = 4.31, SD = 1.45) and the Flashcard baseline (M = 3.12, SD = 1.50). This heightened confidence was attributed by participants to the system's continuous creative support and the consistency maintained across modalities. For \textit{learnability}, however, \textit{WordCraft} received slightly lower ratings (M = 5.62, SD = 1.09) compared to the GPT-4o (M = 5.88, SD = 0.81) and Flashcard (M = 6.44, SD = 0.63) baselines.

\begin{table*}[h]
  \centering
  \caption{Statistical comparisons on the Creativity Support Index (significance: * $p<.050$, ** $p<.010$, *** $p<.001$).}
  \label{tab:csi}
  \setlength{\tabcolsep}{6pt}
  \renewcommand{\arraystretch}{1.18}
  \begin{tabular}{l c c S[scientific-notation=true, round-mode=places, round-precision=2, table-format=1.2e-1] c}
    \toprule
    \textbf{Dimension} &
    \textbf{WordCraft} & \textbf{GPT-4o} &
    {\textbf{$p$-value}} & \textbf{Sig.} \\
    & \textbf{Mean/S.D.} & \textbf{Mean/S.D.} & & \\
    \midrule
    Exploration          & 6.19/0.75 & 4.19/0.98 & 0.0000217034 & \textbf{***} \\
    Expressiveness       & 6.06/0.85 & 3.94/1.29 & 0.0000783225 & \textbf{***} \\
    Enjoyment            & 6.12/0.72 & 3.75/0.86 & 0.0000027365 & \textbf{***} \\
    Immersion            & 5.75/0.93 & 3.56/0.73 & 0.0000061113 & \textbf{***} \\
    Results worth effort & 6.31/1.01 & 4.25/1.24 & 0.0000768479 & \textbf{***} \\
    \bottomrule
  \end{tabular}
\end{table*}

\subsubsection{Creativity Support}
\par Creativity support was assessed using five dimensions from the CSI~\cite{cherry2014quantifying}, comparing \textit{WordCraft} specifically against the GPT-4o baseline: \textit{exploration}, \textit{expressiveness}, \textit{enjoyment}, \textit{immersion}, and \textit{results worth effort} (\cref{tab:csi}). The \textit{collaboration} dimension was excluded, as all tasks were completed individually. 
\par Participants reported greater \textit{exploration} of ideas with \textit{WordCraft} (M = 6.19, SD = 0.75) compared to the baseline (M = 4.19, SD = 0.98). This perception was driven by the system's ability to provide timely inspiration, with participants placing special emphasis on the keyword suggestions. They noted (P1, P3, P13), \textit{``As my thoughts clarified during the process, the keyword anchored my thinking, making the logic behind the exploration distinct.}'' They also rated \textit{expressiveness} higher with \textit{WordCraft} (M = 6.06, SD = 0.85), whereas the baseline was perceived as more restrictive (M = 3.94, SD = 1.29). Most participants linked this enhanced expressiveness to the flexibility of the Mental Imagery Canvas. As P7 remarked, \textit{``Being able to drag and arrange elements on the canvas felt like I was actually building my memory rather than just describing it. It gave me the freedom to express exactly how the keyword and meaning interacted in my head.''} Reports of \textit{enjoyment} were greater with \textit{WordCraft} (M = 6.12, SD = 0.72) than with the baseline (M = 3.75, SD = 0.86), attributed to its ``\textit{gamified nature}'' (P6, P11, P15) and the ``\textit{sustained sense of agency}'' (P2, P9, P10, P16). Participants experienced a deeper sense of \textit{immersion} while using \textit{WordCraft} (M = 5.75, SD = 0.93) relative to the baseline (M = 3.56, SD = 0.73). They credited the Association Map and Recall Path for maintaining cognitive flow. As P14 noted, \textit{``These visual guides kept me constantly thinking about the underlying logic of the connection, rather than worrying about the next step or waiting idly.''} Finally, the perception that \textit{results were worth the effort} was stronger for \textit{WordCraft} (M = 6.31, SD = 1.01) than for the baseline (M = 4.25, SD = 1.24), stemming from ``\textit{the deep personalization of the final outcome}''. Together, these findings underscore \textit{WordCraft}'s superior capacity to support learners' creative engagement in the keyword-method learning tasks.

\subsubsection{Cognitive Effort} 
\par Cognitive effort was assessed using three NASA-TLX~\cite{hart2006nasa} dimensions: \textit{mental demand}, \textit{effort}, and \textit{frustration}. Regarding \textit{mental demand} and \textit{effort}, \textit{WordCraft} elicited moderately higher ratings (Demand: M = 5.12, SD = 0.89; Effort: M = 5.25, SD = 0.93) compared to the GPT-4o baseline (Demand: M = 3.62, SD = 1.50; Effort: M = 3.69, SD = 1.40) and the Flashcard baseline (Demand: M = 3.94, SD = 1.06; Effort: M = 3.88, SD = 1.26). However, regarding \textit{frustration}, \textit{WordCraft} (M = 3.25, SD = 0.86) scored significantly lower than both the unstructured GPT-4o interface (M = 4.69, SD = 0.95) and the Flashcard baseline (M = 4.50, SD = 1.10). Qualitative feedback clarifies these divergent patterns. Participants described the higher load in \textit{WordCraft} as ``\textit{productive engagement}''(P4, P7). Notably, the Association Map was identified as pivotal in mitigating cognitive overload. As P8 explained, \textit{``The elements in the Association Map are well-categorized with clear roles. Thus, the complex logic became manageable rather than overwhelming.''} Conversely, the higher frustration in the GPT-4o group was attributed to the ``\textit{ambiguity of open-ended prompting}'', while the partcipants in the Flashcard group cited the uncertainty of achieving mastery through rote repetition. Overall, \textit{WordCraft} required more cognitive resources while providing a supportive and less frustrating learning experience. Detailed results are presented in \cref{tab:user-feedback-nasatlx}.

\begin{table*}[h]
  \centering
  \caption{Statistical user feedback on the NASA Task Load Index
  (significance: * $p<.050$, ** $p<.010$, *** $p<.001$).}
  \label{tab:user-feedback-nasatlx}
  \setlength{\tabcolsep}{6pt}
  \renewcommand{\arraystretch}{1.18}
  \begin{tabular}{l
                  c c c
                  S[scientific-notation = true, round-mode=places, round-precision=2, table-format=1.2e-1] c
                  S[scientific-notation = true, round-mode=places, round-precision=2, table-format=1.2e-1] c}
    \toprule
    \multirow{2.5}{*}{\textbf{Dimension}} &
      \multicolumn{3}{c}{\textbf{Mean/S.D.}} &
      \multicolumn{2}{c}{\textbf{WordCraft vs. GPT-4o}} &
      \multicolumn{2}{c}{\textbf{WordCraft vs. Flashcard}} \\
    \cmidrule(lr){2-4} \cmidrule(lr){5-6} \cmidrule(lr){7-8}
    & \textbf{WordCraft} & \textbf{GPT-4o} & \textbf{Flashcard} &
      \textbf{$p$-value} & \textbf{Sig.} &
      \textbf{$p$-value} & \textbf{Sig.} \\
    \midrule
    Mental demand  & 5.12/0.89 & 3.62/1.50 & 3.94/1.06 & 0.0111303950  & \textbf{*}  & 0.0082960124 & \textbf{**}    \\
    Effort         & 5.25/0.93 & 3.69/1.40 & 3.88/1.26 & 0.0037135354  & \textbf{**}  & 0.0071959564 & \textbf{**}    \\
    Frustration    & 3.25/0.86 & 4.69/0.95 & 4.50/1.10 & 0.0003386688  & \textbf{***} & 0.0054395862 & \textbf{**}    \\
    \bottomrule
  \end{tabular}
\end{table*}

\subsection{Features of \textit{WordCraft}} 
\par \textit{WordCraft}'s features were evaluated across three dimensions: \textit{generation effect}, \textit{functionality}, and \textit{iterative workflow}. Results for these aspects are summarized descriptively without cross-group comparison, as presented in \cref{tab:wordcraft-features}.

\subsubsection{Generation Effect} 
\par The generation effect in \textit{WordCraft} is embedded throughout the learning process, sustaining active learner engagement. It received high evaluations (M = 6.25, SD = 0.68), consistent with \textbf{DR1}, indicating that the system effectively supported learners in generating their own cues. Most participants (13/16) reported a strong sense of control and agency, describing the process as ``\textit{learning by doing}''. Participants also praised the interaction design for guiding their thinking whenever they sought assistance. As P7 noted when learning the word ``\textit{sear}'', \textit{``I immediately favored the keyword `\thinspace\raisebox{-0.15em}{\includegraphics[page=1, height=0.9em]{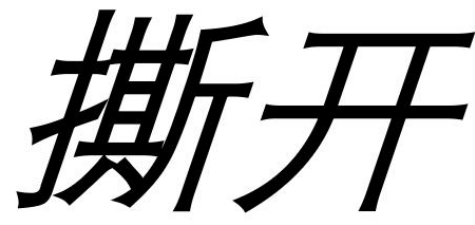}}' (tear) without needing suggestions. But when I got stuck building the association, I clicked `Inspire', and the system prompted me to think about fabric. This heuristic cue bridged the cognitive gap, helping me form my own association that burnt cloth tears more easily.''}

\subsubsection{Functionality: Keyword Selection} 
\par The average score for keyword generation was M = 6.06 (SD = 0.93), indicating that participants generally found the system effective in helping them identify phonologically similar and memorable candidates. This aligns with \textbf{DR2}: most participants (14/16) reported that the generated keywords stimulated their thinking, partly due to their phonological similarity. Participants also valued that the suggestions were not random but meaningfully connected to their semantic brainstorming choices. As P1 explained, \textit{``For \thinspace\raisebox{-0.2em}{\includegraphics[page=1, height=0.9em]{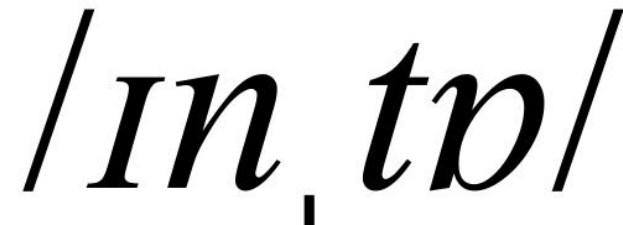}} of `intoxication', I selected the semantic node `delicious' and clicked `inspire'. Suggestions like `\thinspace\raisebox{-0.08em}{\includegraphics[page=1, height=0.9em]{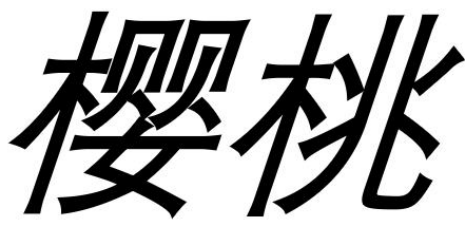}}' (cherry) and `\thinspace\raisebox{-0.08em}{\includegraphics[page=1, height=0.9em]{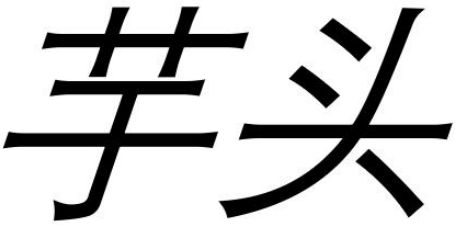}}' (taro) were not only phonetically similar but also aligned with my intended meaning. When I hovered over the card and saw the reasoning `cherry is juicy and sweat,' the link was easy to interpret.''} Some participants, however, noted that the model sometimes struggled to balance practicality and creativity. P2 and P8 commented, ``\textit{Some suggestions used unfamiliar word combinations to create more imaginative or semantically close keywords. Although keyword explanations were provided, this approach did not match my learning habits.''} These perspectives highlight opportunities for refining keyword generation strategies and exploring more personalized support in future iterations.

\par Regarding semantic exploration, participants rated the feature positively (M = 5.75, SD = 1.18), indicating that \textit{WordCraft} helped them develop a deeper understanding of target words and discover contextual cues, consistent with \textbf{DR3}. Participants highlighted that this semantic support was especially valuable for keyword selection. As P14 noted, ``\textit{This support was particularly helpful with low-imageability words. For example, with 'meticulous', the semantic node led me to the word `needle', which immediately gave me a vivid image of precise sewing and helped me generate more usable keyword ideas.''} Furthermore, over half of the participants (10/16) reported that semantic exploration also facilitated association construction. P3 and P16 described the integrated workflow: ``\textit{I could explore related concepts in the Semantic Tree and then accept an AI-generated keyword linked to one of those concepts. The system would then automatically visualize this logical chain in the Association Map on the right, serving as a perfect summary of my semantic bridge.''}

\begin{table*}
\caption{User ratings on \textit{WordCraft} features.}
\label{tab:wordcraft-features}
\centering
\begin{tabular}{llllrr}
\toprule
\textbf{Category} & \textbf{Subcategory} & \textbf{DR} & \textbf{Feature} & \textbf{Mean} & \textbf{SD} \\
\midrule
Generation Effect & --- & DR1 & Active engagement & 6.25 & 0.68 \\
\midrule
\multirow{5}{*}{Functionality} 
 & \multirow{2}{*}{Keyword Selection} & DR2 & Keyword generation & 6.06 & 0.93 \\
 &  & DR3 & Semantic exploration & 5.75 & 1.18 \\
\cmidrule(l){2-6}
 & Association Construction & DR4 & Association mapping & 5.87 & 1.02 \\
\cmidrule(l){2-6}
 & \multirow{2}{*}{Image Formation} & DR5 & Visual imagery & 6.31 & 0.87 \\
 &  & DR6 & Recall path & 5.88 & 0.96 \\
\midrule
Iterative Workflow & --- & --- & Cross-module interaction & 6.19 & 0.75 \\
\bottomrule
\end{tabular}
\end{table*}

\subsubsection{Functionality: Association Construction}
\par The association mapping feature received a mean rating of M = 5.87 (SD = 1.02), indicating \textit{WordCraft} helped participants organize associations systematically, consistent with \textbf{DR4}. Most participants (12/16) reported the association map clarified the relationships between keywords and target meanings, making them easier to manage. \textit{``Seeing the keywords and the target meaning as distinct `Concept Nodes' helped me focus. I could refine one `Association Link' at a time, adjust the pre-filled `Chain Node,' and add my own `Note' to solidify its logic before moving on. This gave me a clear overview instead of a jumble of thoughts.''} (P7, P11) Moreover, participants praised the subtlety and relevance of AI-powered suggestions when exploring associations. As P10 explained, ``\textit{When I clicked for inspiration on a link, it was clear the system had read my notes. Instead of a generic reply, it built directly on my ideas.''} From another perspective, some participants viewed the Association Map as a mediating step whose necessity varied with task complexity. For simple, single-keyword associations, the structure sometimes felt optional. As P15 remarked, ``\textit{When the connection was obvious, I just selected the keyword, drew one link to the target meaning, and typed my association before moving on.}''

\subsubsection{Functionality: Image Formation}
\par Visual imagery received the highest rating among all features (M = 6.31, SD = 0.87), indicating that learners could form mental images effectively through the visual-centric design (\textbf{DR5}). Participants praised how the Mental Imagery Canvas helped them externalize and refine their internal vision. As P7 noted, ``\textit{I imagined a scene, then used the `Add Element' tool to draw a box for the keyword `\thinspace\raisebox{-0.06em}{\includegraphics[page=1, height=0.8em]{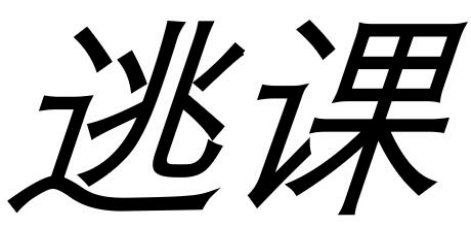}}'(skipping class) and another for the target meaning `\thinspace\raisebox{-0.06em}{\includegraphics[page=1, height=0.8em]{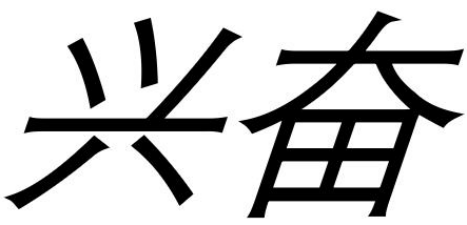}}'(exhilaration). Arranging them on the canvas step-by-step made my mental image clearer, and the final generated picture, following my layout, perfectly confirmed and strengthened that image.}'' The system’s imaginative mapping suggestions were also appreciated (P3, P5, P16). However, some participants noted limitations: P9 and P13 highlighted concerns regarding the limited preset styles for imagery and the relatively long time required for image generation.

\par The recall path received a strong rating (M = 5.88, SD = 0.96), showing that it helped learners maintain clear links between keywords and meanings, consistent with \textbf{DR6}. Many participants emphasized that this feature supported reviewing their cognitive process and keeping mental images coherent. As P13 explained, ``\textit{I would look at the unlit parts of the Recall Path to decide which mapping to create next on the canvas.}'' The interactive feedback also reinforced the task goal, as P2 remarked, ``\textit{I thought I had finished mapping and clicked 'generate', but the system reminded me that a connection in my Recall Path was still missing. This forced me to complete the logic before creating the final image.}'' Some limitations were highlighted. In certain contexts, the recall path was considered overly simplistic, and its necessity was questioned. P9 reflected, ``\textit{How much I could actually get from the recall path depended a lot on the cognitive effort I had already invested earlier, so sometimes it felt less helpful.''} He suggested that giving learners more control over the density of information displayed and integrating the recall path more closely with the association map could enhance its effectiveness.

\subsubsection{Iterative Workflow} 
\par The iterative workflow was rated positively (M = 6.19, SD = 0.75), showing that learners valued the system’s ability to link different modules and support back-and-forth adjustments across stages. This appreciation emerged early, as participants highlighted the flexibility and value of the Phonological Segmentation module. As P10 noted, ``\textit{I could brush and re-brush different phonetic chunks, experimenting with multiple keyword ideas before settling on one.}'' This iterative quality extended throughout the system. \textit{``I went back to the Keyword Selection stage to change a keyword, and when I returned to the map, it was already updated. I didn’t have to redo everything from scratch.}'' (P6) This automated synchronization across modules supported a non-linear, exploratory thought process, which P11 described as feeling \textit{``natural, and more like real thinking.''}

\subsection{Learning Outcomes}
\par
\begin{figure*}[h]
    \centering
    \includegraphics[width=\textwidth]{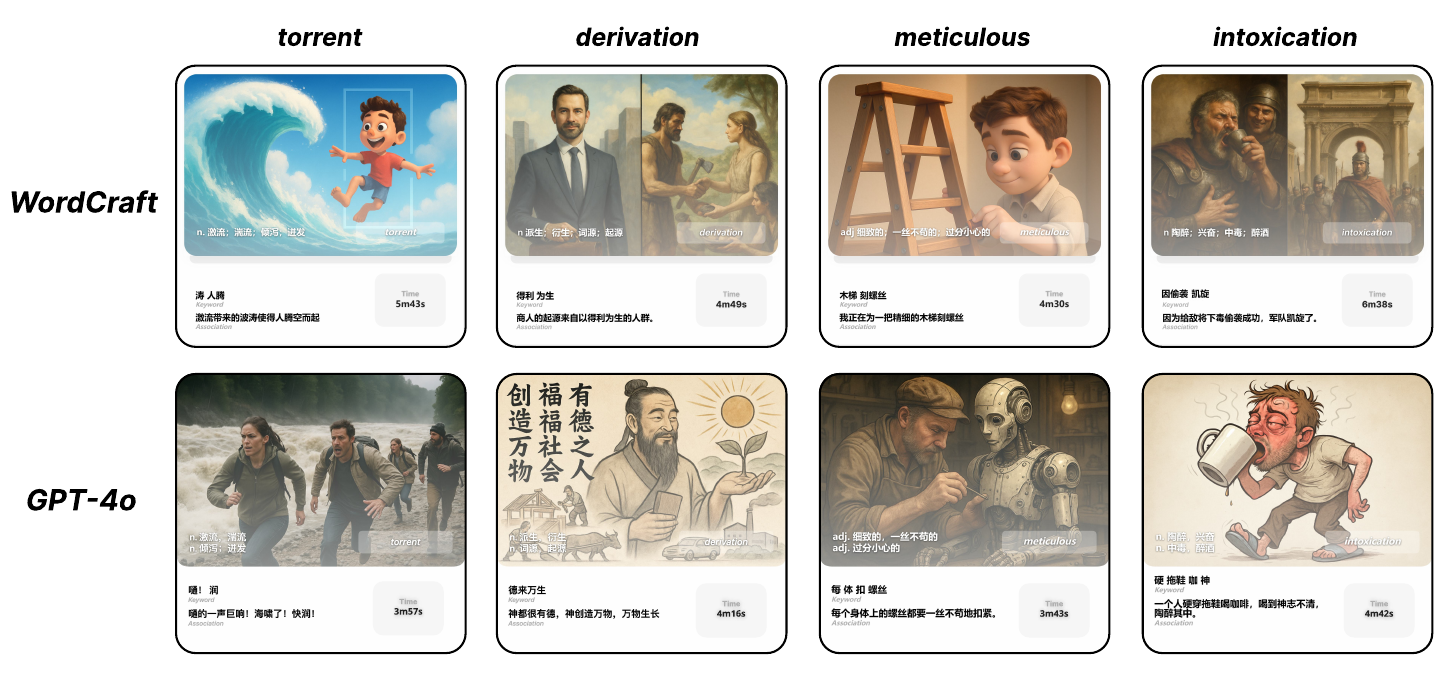}
    \caption{Examples of the generated outputs using \textit{WordCraft} and \textit{GPT-4o}.}
    \label{fig:examples}
\end{figure*}

 Learning outcomes were evaluated using \textit{completion time}, \textit{result scoring}, and \textit{recall performance}, providing a comprehensive assessment of participant learning and enabling direct comparison between \textit{WordCraft} and the baseline conditions. Representative examples of the generated outputs are presented in \cref{fig:examples}.

\begin{figure*}[h]
    \centering
    \includegraphics[width=\textwidth]{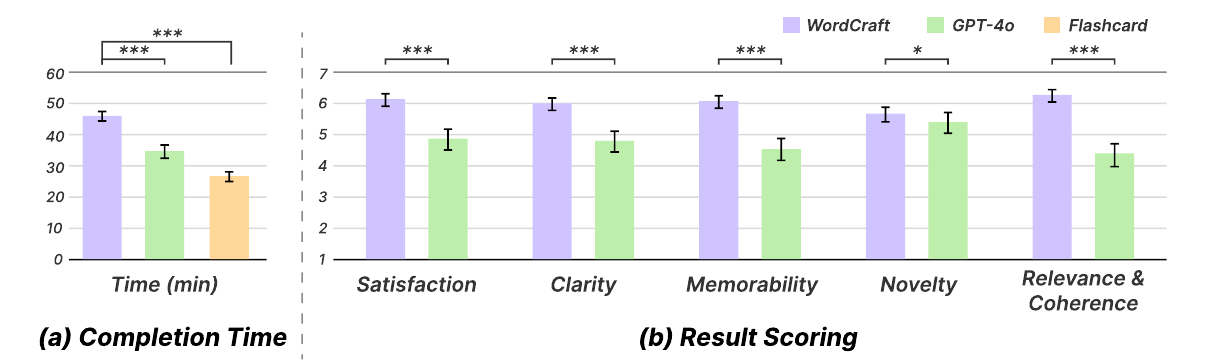}
    \caption{Generated Clues Evaluation. (a) Completion Time with \textit{WordCraft} is longer than with the \textit{GPT-4o} baseline and \textit{Flashcard} baseline, indicating deeper cognitive engagement. (b) Result Scoring shows that cues generated with \textit{WordCraft} receive higher ratings across multiple dimensions compared to \textit{GPT-4o}. Overall, all metrics demonstrate improvement over with the baseline.}
    \label{fig:state_fig2}
    \vspace{-1mm}
\end{figure*}

\subsubsection{Completion Time} \par As shown in \cref{fig:state_fig2} (a), participants using \textit{WordCraft} spent more time on average (M = 46.49, SD = 4.81) compared to those using the GPT-4o baseline (M = 34.88, SD = 7.54) and the Flashcard baseline (M = 26.94, SD = 5.16). This increased duration reflects the deep cognitive processing fostered by \textit{WordCraft}'s structured workflow, contrasting with the rapid, surface-level review of the Flashcard condition and the unstructured prompting of the GPT-4o baseline. Participants noted that this time investment was acceptable given the perceived benefits. As P2 explained, ``\textit{It did take longer, but I didn't feel the time passing because I was engaged step-by-step. In the end, the extra time felt worthwhile for a stronger memory.}''

\subsubsection{Result Scoring} Result scoring assessed participants’ evaluations of the generated cues across five dimensions: \textit{Satisfaction}, \textit{Clarity}, \textit{Memorability}, \textit{Novelty}, and \textit{Relevance \& Coherence}. Overall, cues produced with \textit{WordCraft} received higher average ratings than baseline, indicating that the structured workflow generated outputs perceived as more useful and reliable for vocabulary learning (\cref{fig:state_fig2} (b)). Differences were greatest in \textit{Relevance \& Coherence}, suggesting that \textit{WordCraft} helped learners form stronger and more consistent connections between cues and target meanings. \textit{Satisfaction} also rose notably, reflecting learners’ positive perception of outcomes and their deeper engagement in the creative process.

\subsubsection{Recall Performance} \par
\begin{figure*}[h]
    \centering
    \includegraphics[width=\textwidth]{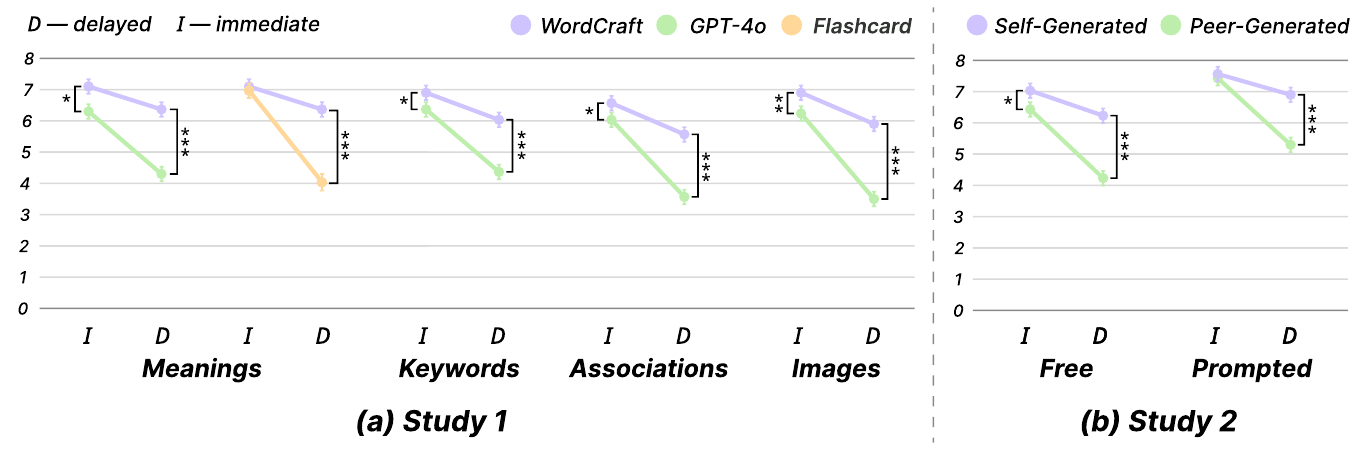}
    \caption{Recall Performance. (a) In Study 1, the WordCraft group consistently outperformed both the GPT-4o and Flashcard baselines across all recall metrics. This advantage was particularly evident in the delayed test, conducted one week later, where the difference became substantially larger. (b) In Study 2, for free recall, self-generated cues conferred a modest immediate advantage that grew significantly in the delayed test. For prompted recall, performance in both conditions approached ceiling levels initially, yet self-generated cues maintained a clear advantage in the delayed test.}
    \label{fig:state_fig1}
\end{figure*}

 We compared recall performance across the three conditions for word meanings, and additionally compared the \textit{WordCraft} and \textit{GPT-4o} groups on their recall of the created keywords, associations, and images. As shown in \cref{fig:state_fig1} (a), the \textit{WordCraft} group demonstrated the highest recall rates across all measures. 
Regarding word meaning recall, both the \textit{WordCraft} and \textit{Flashcard} groups performed well in the immediate test, with \textit{WordCraft} showing a slight advantage, though not significant. However, in the delayed test conducted one week later, the \textit{WordCraft} group significantly outperformed the \textit{Flashcard} group, suggesting stronger support for long-term retention. Similarly, for recalling created cues, the \textit{WordCraft} group outperformed the GPT-4o group in both the immediate and delayed tests, with the advantage becoming more pronounced after one week. This indicates that the structured process creates more durable and interconnected learning traces. 


\subsubsection{Summary} \par Overall, compared with the baselines, \textit{WordCraft} received higher ratings for generated cue quality and supported stronger recall performance, particularly in the delayed test, though it required more time to complete tasks. The GPT-4o condition showed high variability and user frustration, with outcomes heavily dependent on individual prompting skill. The Flashcard condition, though simple, proved ineffective for robust initial encoding. The next section presents qualitative findings to explain these divergent outcomes. 

\subsection{Impact on the Learning Process} Beyond quantitative recall performance, our qualitative analysis shows that \textit{WordCraft} fundamentally reshapes the learning process through its multimodal creation workflow under structured scaffold and its emphasis on deep, reflective engagement with target words.

\subsubsection{Multimodal Creation under Structured Scaffolding Enriches the Learning Experience and Fosters Retention} \par Compared with traditional flashcards, the multimodal creation process in both generative conditions significantly enriched the learning experience. Participants (27/32) consistently highlighted the visual output as a critical driver of motivation. As P10 described, \textit{``seeing abstract ideas turn into concrete images felt like a reward that kept me focused.''} This sharply contrasted with the Flashcard condition, which participants described as \textit{``mechanical drudgery''} (P36, P41, P43). Moreover, participants praised the effectiveness of externalizing mental imagery into concrete visual artifacts, which provided \textit{``an additional retrieval channel that strengthened memory traces beyond text alone.''} (P16)

\par This structured, step-by-step approach to multimodal creation also distinguished \textit{WordCraft} from the unstructured GPT-4o baseline. Participants explained that the interaction workflow in \textit{WordCraft} required them to move from the textual logic of the Association Map to the spatial arrangement on the Mental Imagery Canvas, compelling them to build coherent and interconnected mnemonics. This resulted in memories described as \textit{``logical chains,''} where recalling one element often triggered accurate recall of the entire narrative. In contrast, the GPT-4o baseline’s text-centric interaction encouraged trial-and-error prompting, producing less integrated \textit{``aha''} moments from model-generated images that lacked a self-authored foundation. Additionally, several participants (P1, P3, P15) emphasized that coherence between the generated image and their textual intent increased confidence. As P15 explained, \textit{``when the system displayed an image that perfectly matched what I intended, I knew I had created a mnemonic I would remember for a long time.''}

\subsubsection{Deep Processing could Promote More Reflective Learning} \par The deep cognitive processing required by \textit{WordCraft} created highly durable, episodic memories. Participants reported that this prompted more efficient review. As P9 noted, \textit{``I didn't even need to open the word card afterward; I just remembered my thought process at the time, and the memory would often replay in my mind.''} Further, participants envisioned that \textit{WordCraft} would function as a powerful tool for dedicating time to memorizing difficult words, distinguishing it from traditional rote memorization. This deliberate investment of effort, they noted, would foster a more reflective learning process. As P8 explained, it prompted him to self-assess \textit{``whether I truly mastered a word,''} ultimately improving overall memory efficacy.
However, contrasting views were also raised. Due to its time-consuming nature, \textit{WordCraft} was perceived as ``\textit{ill-suited for learning new words in fragmented moments}'' (P6, P14).




\subsection{Limitations and Contextual Boundaries of \textit{WordCraft}}
\par Despite \textit{WordCraft}'s overall effectiveness, our qualitative analysis also revealed the key boundaries of its application, focusing on its constraints in flexibility and misalignments with user intent.

\subsubsection{\textit{WordCraft} May Constrain Flexible Application of Memory Strategies}\par Some participants noted that for relatively simple words such as `sear', they could remember the meaning directly without going through the keyword method. In such cases, the system's scaffolding felt redundant and added extra burden (P12, P15). Other participants reported that the fixed workflow of the keyword method constrained the diverse strategies they normally use. For example, some learners preferred to flexibly combine morphological decomposition, semantic memory, or partial keyword associations based on the characteristics of the word, especially when dealing with longer words (P4, P11). As P11 stated: ``\textit{When I was learning `intoxication', I preferred to break out the prefixes and suffixes `in-' and `-ication', and then use the keyword method to memorize the root `tox'. But the system does not support combining these two memory methods.}'' These participants felt that the unified, keyword-centered workflow reduced their flexibility and creativity.

\subsubsection{MLLM-Generated Suggestions Sometimes Misalign with User Intent and Experience}\par The suggestions generated by the \textit{MLLM} also showed certain limitations. First, learners were accustomed to generating highly personalized associations rooted in their unique cognitive and experiential networks, which made it difficult for the model to provide targeted suggestions (P4, P6, P14). As P6 described: ``\textit{When memorizing `vibrant,' I first extracted \thinspace\raisebox{-0.08em}{\includegraphics[page=1, height=0.8em]{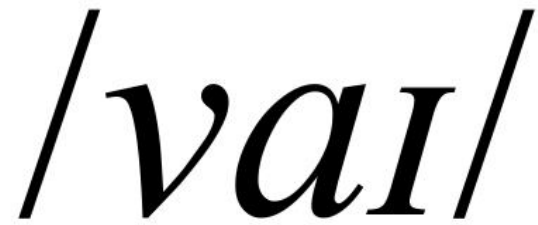}} and chose `\thinspace\raisebox{-0.06em}{\includegraphics[page=1, height=0.8em]{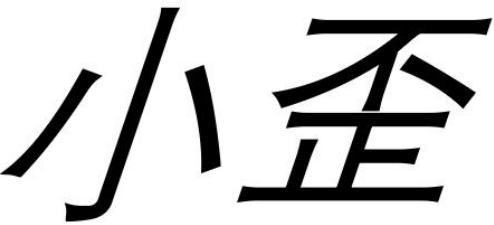}}'(name of her dog) as the keyword. It vividly reflecting the meaning of the word. This association is natural and easy to remember for me. Although the LLM's suggested keywords had similar sounds, they could never be more fitting than what I came up with myself.}'' In contrast, the model sometimes generated unfamiliar words or associations for learners, which instead increased cognitive load (P3, P6, P9). As P9 explained: ``\textit{When I was learning `derivation', I extracted \thinspace\raisebox{-0.1em}{\includegraphics[page=1, height=0.9em]{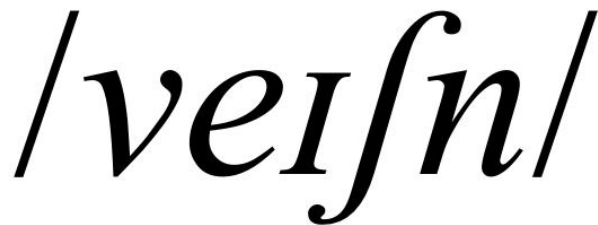}}. One of the model's keyword recommendations was `\thinspace\raisebox{-0.06em}{\includegraphics[page=1, height=0.8em]{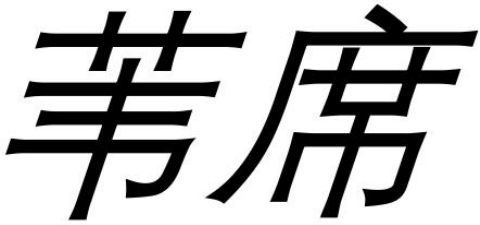}}'(reed mat), which does sound similar, but I had no idea what it was. This confused me.}'' Some participants also noted that their most memorable cues often relied on exaggerated and intense imagery, sometimes involving death or violence. However, due to safety constraints, such imagery could not be generated or expanded by the model, causing the resulting memory cues to lack vividness and impact, and failing to match their intentions(P2, P14). As P14 described: ``\textit{When learning `slough', I thought of the keyword `\thinspace\raisebox{-0.06em}{\includegraphics[page=1, height=0.8em]{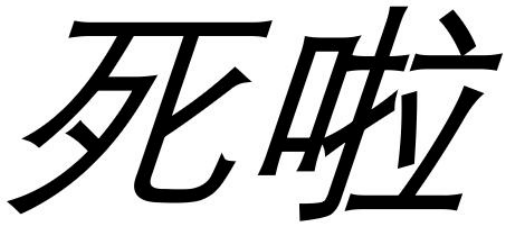}}'(dead), and planned to draw a corpse with peeling skin. However, LLM directly rejected my idea because it involved death, which made me very frustrated.}`''


\section{RESULTS OF STUDY 2}

\par To evaluate whether \textit{WordCraft} preserved the generation effect, we compared recall performance between self-generated and peer-generated cue conditions. The results are shown in \cref{fig:state_fig1} (b). 

\par For \textbf{immediate free recall}, participants correctly recalled slightly more words with self-generated cues (M = 7.10, SD = 0.91) than with peer-generated cues (M = 6.45, SD = 1.00), indicating an early advantage of self-generation in short-term memory. This advantage became more pronounced in the delayed free recall test conducted one week later, with self-generated cues yielding substantially higher recall counts (M = 6.25, SD = 0.72) compared to peer-generated cues (M = 4.20, SD = 0.77). For \textbf{prompted recall}, the number of correctly recalled items increased in both conditions. In the immediate prompted recall test, performance in both groups approached ceiling levels (self-generated: M = 7.60, SD = 0.75; peer-generated: M = 7.45, SD = 0.83), leaving little room for differentiation. In contrast, during the delayed prompted recall, self-generated cues maintained a clear advantage (M = 6.95, SD = 0.76) over peer-generated cues (M = 5.25, SD = 0.79).

\par Overall, these results indicate that \textit{WordCraft} effectively preserved the generation effect, with self-generated cues consistently outperforming peer-generated cues across both free and prompted recall. The advantage was modest in the short term but became increasingly pronounced in delayed recall, highlighting the deeper and more durable learning benefits of active cue creation.

\section{DISCUSSION}
\subsection{Design Implications}
\subsubsection{Designing Support to Sustain the Generation Effect}

\par Compared with cases where learners struggle to perform all stages of the keyword method alone or where ready-made materials offer excessive aid, \textit{WordCraft} seeks to balance learner effort and sustain the generation effect. Challenges such as reduced learner agency~\cite{10.1145/3613905.3651042}, increased hallucination~\cite{guerreiro2023hallucinations,10.1145/3613904.3642428}, and inconsistent output quality~\cite{10.1145/3613905.3651042}, common when directly introducing MLLMs, also arise here. Through \textit{WordCraft}’s structured workflow and moderated human–AI collaboration, learners’ sense of self-agency was supported by both qualitative feedback and quantitative data. Two key design considerations emerged: when to intervene and how to assist. For the former, \textit{WordCraft} lets users input their own answers and provides structured guidance at each stage. Future versions could adopt a more proactive AI approach~\cite{10.1145/3706598.3715579,10.1145/3613905.3650912} that captures user behavior, builds learner profiles, and uses real-time data for adaptive, context-aware interventions. Incorporating techniques like the Socratic style~\cite{paul2007critical,10.1145/3544548.3580672} and staged answer revelation~\cite{10.1145/3706598.3713748,10.1145/3341525.3387411,10.1145/3613905.3650937} may further sustain the generation effect by fostering deeper reasoning and gradual self-construction, offering concrete directions for design refinement.


\subsubsection{Designing Image Generation Around User Roles}

\par Visual-centric interaction approaches~\cite{shi2025brickify} help users externalize their thoughts and creativity by turning abstract ideas into visual elements and structures. \textit{WordCraft} adopts this approach, allowing users to construct mental imagery on a canvas using entity locations and descriptive inputs. Early in the design, this was compared with layout-guided text-to-image generation~\cite{lin2025sketchflex,li2023gligen}, as both emphasize layout and entity information. However, given the target users, \textit{WordCraft} does not aim for precise positional control; instead, it uses layout input to support user expression and cognitive externalization. This open design also gives MLLMs more freedom to interpret user intent, enhancing flexibility and personalization.


\par Additional insights from the semi-structured interviews highlighted two distinct participant roles. Most participants acted as learners, focusing on constructing and refining mental imagery and valuing outputs that supported inner visualization. P11's explanation confirmed this, noting the generated image felt supportive because it matched her process of building a mental picture. In contrast, two participants (P5, P14) were more inclined to act as creators, seeking higher visual fidelity and control for social sharing. As P5 said, \textit{``If I could have stronger control over the visual characteristics, positions, and relationships among elements in the image, it would improve my experience when sharing it!''} These insights suggest that learners prioritize the cognitive process of image construction, while creators emphasize control and visual quality, indicating that future designs should support both process-oriented learning and creator-oriented customization.


\subsubsection{Balancing User Creativity and Structured Cognitive Workflow}
\begin{figure*}[h]
    \centering
    \includegraphics[width=\textwidth]{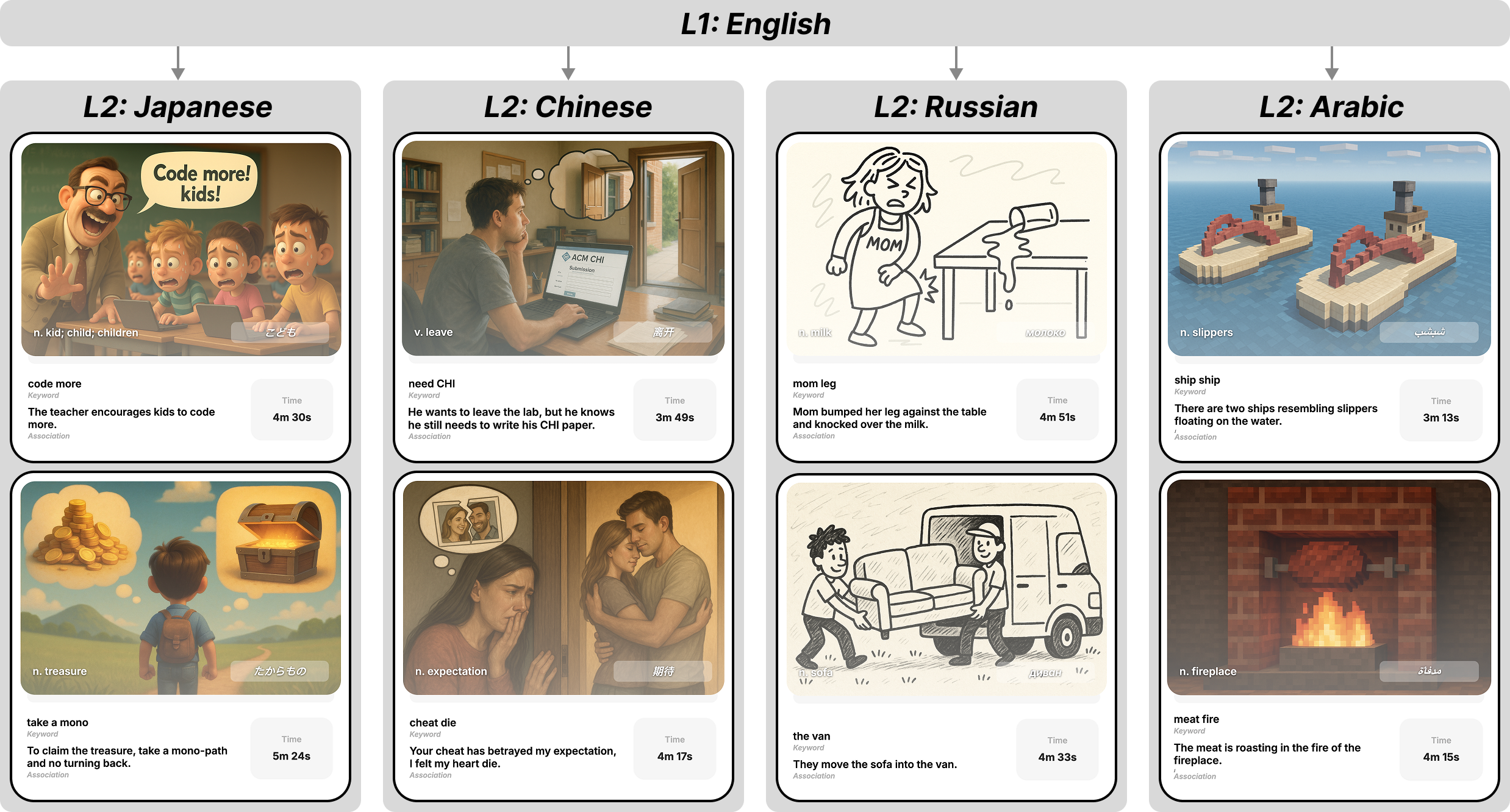}
    \caption{Examples of \textit{WordCraft} applied to various language pairs.}
    \label{fig:generalizability}
\end{figure*}

\par \textit{WordCraft} was designed with a clear, stage-based workflow that systematically supports the keyword method while encouraging learner creativity. Its structured process integrates semantic brainstorming, association mapping, and visual imagery to promote consistent keyword generation, reduce cognitive load, and align images with learners’ mental representations.

\par However, this same structured approach introduces a trade-off between logical coherence and creative freedom, as two participants felt that the predefined sequence limited the kind of open-ended, spontaneous exploration they could pursue in the GPT-4o interface, where learners were free to input ideas in a more exploratory, ambiguous manner. This trade-off highlights the challenge of designing tools that maintain systematic guidance without constraining flexible, creative thinking. Looking ahead, future versions of \textit{WordCraft} could offer optional ``free-flow'' modes or adjustable stage controls, enabling learners to shift between structured guidance and open exploration as needed while still benefiting from the system's core support for the keyword method.




\subsection{Implications for L2 Vocabulary Learning and Human–AI Collaborative Education}

\par Both user studies revealed clear memory benefits: participants using \textit{WordCraft} achieved higher immediate and delayed recall scores, showing stronger long-term retention of target words. Additionally, testing observations also revealed that a single keyword in the \textit{WordCraft} group often triggered the full associative path, indicating deeper engagement and more coherent associations. Beyond these quantitative results, the findings align with constructivist learning theory, which emphasizes active knowledge construction over passive reception~\cite{hein1991constructivist, bada2015constructivism, narayan2013constructivism}. \textit{WordCraft} acts as cognitive augmentation, providing dynamic scaffolding that supports learners in building understanding while preserving the agency vital for deep learning~\cite{leong2024putting, chen2024retassist}. In the formative study, teachers (T1, T2) noted ongoing challenges in vocabulary instruction, including large class sizes and limited opportunities for individualized, creativity-driven learning. Our results suggest that \textit{WordCraft} has the potential to address these issues by providing targeted cognitive support for foundational vocabulary learning, enabling educators to focus more on designing rich learning experiences, guiding critical inquiry, and fostering holistic student development~\cite{atchley2024human, molenaar2024human, dang2025human}. This perspective points toward a human–AI collaborative paradigm in which AI offers precise cognitive assistance while teachers act as designers and facilitators of students' intellectual growth.

\subsection{Fostering a Collaborative Keyword-Method Community}

\par Community-based vocabulary learning~\cite{zhu2024surveying,10.1145/3605390.3605395} is a common phenomenon in which learners share learning resources, exchange memory strategies, and motivate one another, thereby forming a collaboration- and feedback-driven learning ecosystem. Most participants (15/16) expressed a strong desire to share their results and see how others used associative techniques to remember the same word, especially for complex or abstract vocabulary. As P3 noted, \textit{``I really want to see how others imagine this word. Maybe their stories could inspire me.''} This underscores \textit{WordCraft}'s potential to evolve into a community of keyword-method practitioners, where users share their keywords, associative processes, and imagery concepts while exploring others’ ideas—transforming an individual memory task into an open, mutually inspiring learning experience.


\par However, this interest in community features reveals a potential tension with the generation effect. P1, P4, and P12 expressed similar concerns during interviews, noting that prematurely viewing others’ complete associations or images might diminish motivation to think independently. To balance sharing with the generation effect, future designs should move beyond displaying final images. The system could invite creators to upload their complete cognitive workflows, enabling others to trace how ideas developed. The community might also maintain a curated repository of keywords and associative links, where users contribute partial inspirations. Techniques like retrieval-augmented generation (RAG)~\cite{lewis2021retrievalaugmentedgenerationknowledgeintensivenlp} could further integrate shared fragments into large language models to produce more diverse and practical prompts throughout \textit{WordCraft}'s workflow.

\subsection{Generalizability}

\par Although \textit{WordCraft} was initially designed for L1 Chinese–L2 English learners, its underlying design principles are not linguistically exclusive. \textit{WordCraft}'s scaffolding process, including keyword generation, association construction, and image formation, focuses on providing process-level cognitive support rather than language-specific content, making it adaptable to other language pairs. \Cref{fig:generalizability} illustrates several conceptual examples of how \textit{WordCraft} could be instantiated for various linguistic contexts. These examples serve to demonstrate design flexibility and potential for adaption, rather than to provide empirical evidence of effectiveness in those contexts. However, it is crucial to note that \textit{WordCraft} primarily provides cognitive scaffolding for the process of applying the keyword method and cannot overcome the inherent limitations of the method itself, such as its reduced effectiveness for language pairs with low phonological similarity~\cite{pressley1982mnemonic,kang2025phonitale}. Beyond the keyword method, \textit{WordCraft}'s framework may potentially support the integration of other mnemonic strategies. For example, the pegword method~\cite{BOWER1972478,sahadevan2021imagery,bower1970imagery,bugelski1968images} pairs rhyming pegwords (e.g., "one–sun," "two–shoe") with target words through semantic associations and imagery, closely aligning with \textit{WordCraft}'s workflow with minimal adjustments. Similarly, narrative mnemonics~\cite{bower1969narrative,bellezza1981mnemonic,pressley1982mnemonic} and the link method~\cite{higbee1990using,sahadevan2021imagery,roediger1980effectiveness} also rely on semantic association and imagery encoding, making them broadly compatible with the system. Crucially, \textit{WordCraft} emphasizes guided generation, engaging learners actively and preserving the benefits of the generation effect, while adhering to cognitive intervention principles in HCI, making it potentially transferable to other high-cognitive-load learning tools. Finally, \textit{WordCraft} supports modular integration, functioning as an on-demand tool within vocabulary systems to provide mnemonic support for challenging words.

\subsection{Limitations and Future Work}
\par This work has several limitations. First, our participant pool was linguistically and culturally homogeneous, consisting exclusively of L1 Chinese learners of English. While we discussed cases generated by \textit{WordCraft} for other languages, the absence of multilingual participants in the experiments constrains the cross-linguistic generalizability of our findings. Future work should involve a more diverse learner group to confirm these results across different language combinations. Second, the scale of the learning task was limited, particularly regarding the number of target words. While we clarified the methodological rationale for this set size in Section \ref{sec:study1_prep}, we acknowledge that a larger vocabulary set would better reflect classroom-scale efficacy. Future work will involve a longitudinal tracking experiment designed to simulate more realistic learning scenarios, evaluating the system's sustained efficacy over extended periods. Finally, the system operates within the constraints of the keyword method, limiting support for polysemous vocabulary by focusing on a single word meaning at a time and prioritizing word meaning over spelling. Subsequent studies should explore ways to enhance its support for more comprehensive vocabulary learning.

\section{CONCLUSION}
\par L2 learners often experience high cognitive load when applying the keyword method independently, while direct reliance on MLLMs can lead to a weakened generation effect and unstable output quality. To address these challenges, this study introduced \textit{WordCraft}, a process-level creativity support tool that helps L2 learners apply the keyword method while balancing learner agency and cognitive load. Two user studies respectively demonstrated that \textit{WordCraft} effectively supports keyword-method learning and preserves the generation effect for improved long-term vocabulary retention. The discussion also highlights \textit{WordCraft}'s potential to foster creativity through adaptive guidance, flexible user control, and community-based sharing while sustaining the generation effect.

\begin{acks}
We gratefully acknowledge the anonymous reviewers for their insightful feedback. This research was supported by the National Natural Science Foundation of China (No. 62372298), the Shanghai Engineering Research Center of Intelligent Vision and Imaging, the Shanghai Frontiers Science Center of Human-centered Artificial Intelligence (ShangHAI), and the MoE Key Laboratory of Intelligent Perception and Human-Machine Collaboration (KLIP-HuMaCo).
\end{acks}

\bibliographystyle{ACM-Reference-Format}
\bibliography{sample-base}

@article{atkinson1975application,
  title={An application of the mnemonic keyword method to the acquisition of a Russian vocabulary.},
  author={Atkinson, Richard C and Raugh, Michael R},
  journal={Journal of experimental psychology: Human learning and memory},
  volume={1},
  number={2},
  pages={126},
  year={1975},
  publisher={American Psychological Association}
}

@article{beaton1995retention,
  title={Retention of foreign vocabulary learned using the keyword method: A ten-year follow-up},
  author={Beaton, Alan and Gruneberg, Michael and Ellis, Nick},
  journal={Second Language Research},
  volume={11},
  number={2},
  pages={112--120},
  year={1995},
  publisher={Sage Publications Sage CA: Thousand Oaks, CA}
}

@article{al2019effect,
  title={The effect of mnemonic keyword strategy instruction on vocabulary retention of students with learning disabilities},
  author={Al-Khawaldeh, Mohammad Abedrabbu and Al-Khasawneh, Fadi Maher},
  journal={International Journal of English Linguistics},
  volume={9},
  number={4},
  pages={138--144},
  year={2019},
  publisher={Canadian Center of Science and Education}
}

@article{campos2010efficacy,
  title={Efficacy of the keyword mnemonic method in adults},
  author={Campos, Alfredo and P{\'e}rez-Fabello, Mar{\'\i}a Jos{\'e} and Camino, Estefan{\'\i}a},
  journal={Psicothema},
  volume={22},
  number={4},
  pages={752--757},
  year={2010}
}

@article{qu2024facilitative,
  title={The facilitative effect of the keyword mnemonic on L2 vocabulary retrieval practice},
  author={Qu, Kejia and Liu, Tianzhi and Qiao, Yihuan and Wang, Pengcheng},
  journal={Heliyon},
  volume={10},
  number={3},
  year={2024},
  publisher={Elsevier}
}

@book{paivio2013imagery,
  title={Imagery and verbal processes},
  author={Paivio, Allan},
  year={2013},
  publisher={Psychology Press}
}

@article{cancino2021role,
  title={The Role of Visual Cues in the Keyword Method: Assessing Variations of the Mnemonic Approach in L2 Vocabulary Learning.},
  author={Cancino, Marco and Silva, Javier and Gatica, Francisca},
  journal={Mextesol Journal},
  volume={45},
  number={1},
  pages={n1},
  year={2021},
  publisher={ERIC}
}

@article{slamecka1978generation,
  title={The generation effect: Delineation of a phenomenon.},
  author={Slamecka, Norman J and Graf, Peter},
  journal={Journal of experimental Psychology: Human learning and Memory},
  volume={4},
  number={6},
  pages={592},
  year={1978},
  publisher={American Psychological Association}
}

@article{campos2004importance,
  title={The importance of the keyword-generation method in keyword mnemonics},
  author={Campos, Alfredo and Amor, Angeles and Gonz{\'a}lez, Mar{\'\i}a Angeles},
  journal={Experimental psychology},
  volume={51},
  number={2},
  pages={125--131},
  year={2004},
  publisher={Hogrefe \& Huber Publishers}
}

@article{bird1999examination,
  title={An examination of the keyword method: How effective is it for native speakers of Chinese learning English},
  author={Bird, Stephen A and Jacobs, George M},
  journal={Asian Journal of English Language Teaching},
  volume={9},
  pages={75--97},
  year={1999},
  publisher={English Language Teaching Unit and the Department of English at CUHK \& 中文~…}
}

@inproceedings{savva2014transphoner,
  title={Transphoner: Automated mnemonic keyword generation},
  author={Savva, Manolis and Chang, Angel X and Manning, Christopher D and Hanrahan, Pat},
  booktitle={Proceedings of the SIGCHI Conference on Human Factors in Computing Systems},
  pages={3725--3734},
  year={2014}
}

@article{lee2024exploring,
  title={Exploring Automated Keyword Mnemonics Generation with Large Language Models via Overgenerate-and-Rank},
  author={Lee, Jaewook and McNichols, Hunter and Lan, Andrew},
  journal={arXiv preprint arXiv:2409.13952},
  year={2024}
}

@inproceedings{lee2023smartphone,
  title={Smartphone: Exploring keyword mnemonic with auto-generated verbal and visual cues},
  author={Lee, Jaewook and Lan, Andrew},
  booktitle={International Conference on Artificial Intelligence in Education},
  pages={16--27},
  year={2023},
  organization={Springer}
}

@article{weerasinghe2022vocabulary,
  title={Vocabulary: Learning vocabulary in ar supported by keyword visualisations},
  author={Weerasinghe, Maheshya and Biener, Verena and Grubert, Jens and Quigley, Aaron and Toniolo, Alice and Pucihar, Klen {\v{C}}opi{\v{c}} and Kljun, Matja{\v{z}}},
  journal={IEEE Transactions on Visualization and Computer Graphics},
  volume={28},
  number={11},
  pages={3748--3758},
  year={2022},
  publisher={IEEE}
}

@article{shapiro2005investigation,
  title={An investigation of the cognitive processes underlying the keyword method of foreign vocabulary learning},
  author={Shapiro, Amy M and Waters, Dusty L},
  journal={Language teaching research},
  volume={9},
  number={2},
  pages={129--146},
  year={2005},
  publisher={Sage Publications Sage CA: Thousand Oaks, CA}
}

@article{sagarra2006key,
  title={The key is in the keyword: L2 vocabulary learning methods with beginning learners of Spanish},
  author={Sagarra, Nuria and Alba, Matthew},
  journal={The modern language journal},
  volume={90},
  number={2},
  pages={228--243},
  year={2006},
  publisher={Wiley Online Library}
}

@article{piribabadi2014effect,
  title={The effect of the keyword method and word-list method instruction on ESP vocabulary learning},
  author={Piribabadi, Ana and Rahmany, Ramin},
  journal={Journal of Language Teaching and Research},
  volume={5},
  number={5},
  pages={1110--1115},
  year={2014},
  publisher={Academy Publication Co., Ltd.}
}

@article{al2011effectiveness,
  title={The Effectiveness of Keyword-Based Instruction in Enhancing English Vocabulary Achievement and Retention of Intermediate Stage Pupils with Different Working Memory Capacities.},
  author={Al-Zahrani, Mona Abdullah Bakheet},
  journal={Online Submission},
  year={2011},
  publisher={ERIC}
}

@article{chen2017phonetic,
  title={Phonetic Matching, Semanticized Phonetic Matching and Phono-Semantic Matching as Techniques in Keyword Selection},
  author={Chen, Yan},
  journal={English Language and Literature Studies},
  volume={7},
  number={1},
  year={2017}
}

@article{CHINGSHYANGCHANG2007534,
title = {The impact of vocabulary preparation on L2 listening comprehension, confidence and strategy use},
journal = {System},
volume = {35},
number = {4},
pages = {534-550},
year = {2007},
issn = {0346-251X},
doi = {https://doi.org/10.1016/j.system.2007.06.003},
url = {https://www.sciencedirect.com/science/article/pii/S0346251X07000632},
author = {Anna {Ching-Shyang Chang}}
}

@article{goulden1990large,
  title={How large can a receptive vocabulary be?},
  author={Goulden, Robin and Nation, Paul and Read, John},
  journal={Applied linguistics},
  volume={11},
  number={4},
  pages={341--363},
  year={1990},
  publisher={Oxford University Press}
}

@article{schmitt2017much,
  title={How much vocabulary is needed to use English? Replication of van Zeeland \& Schmitt (2012), Nation (2006) and Cobb (2007)},
  author={Schmitt, Norbert and Cobb, Tom and Horst, Marlise and Schmitt, Diane},
  journal={Language Teaching},
  volume={50},
  number={2},
  pages={212--226},
  year={2017},
  publisher={Cambridge University Press}
}

@article{QU2024e25212,
title = {The facilitative effect of the keyword mnemonic on L2 vocabulary retrieval practice},
journal = {Heliyon},
volume = {10},
number = {3},
pages = {e25212},
year = {2024},
issn = {2405-8440},
doi = {https://doi.org/10.1016/j.heliyon.2024.e25212},
url = {https://www.sciencedirect.com/science/article/pii/S240584402401243X},
author = {Kejia Qu and Tianzhi Liu and Yihuan Qiao and Pengcheng Wang}
}

@article{levin1981mnemonic,
  title={The mnemonic ‘80s: Keywords in the classroom},
  author={Levin, Joel R},
  journal={Educational Psychologist},
  volume={16},
  number={2},
  pages={65--82},
  year={1981},
  publisher={Taylor \& Francis}
}

@inproceedings{frich2019mapping,
  title={Mapping the landscape of creativity support tools in HCI},
  author={Frich, Jonas and MacDonald Vermeulen, Lindsay and Remy, Christian and Biskjaer, Michael Mose and Dalsgaard, Peter},
  booktitle={Proceedings of the 2019 CHI conference on human factors in computing systems},
  pages={1--18},
  year={2019}
}

@article{article3,
author = {Muthyala, Udaya},
year = {2022},
month = {01},
pages = {},
title = {USING SEMANTIC MAPS AS A TEACHING STRATEGY FOR VOCABULARY DEVELOPMENT},
volume = {6},
journal = {European Journal of English Language Teaching},
doi = {10.46827/ejel.v6i5.4095}
}

@article{mutar2024flashcard,
  title={Flashcard strategy role in teaching english vocabulary: a systematic review},
  author={Mutar, QM},
  journal={International Journal of Social Science Research and Review},
  volume={7},
  number={4},
  pages={37--53},
  year={2024}
}

@article{ismayilli2025impact,
  title={The Impact of Educational Games on Speaking Skills in the Foreign Language Teaching Process.},
  author={Ismayilli, Turkan Mehraj and Mammadova, Khatira Mahammad and Asadova, Aysel Alakbar},
  journal={Novitas-ROYAL (Research on Youth and Language)},
  volume={19},
  number={1},
  pages={229--240},
  year={2025},
  publisher={ERIC}
}

@inproceedings{10.1145/3706598.3713935,
author = {Zhou, Qinyi and Deng, Jie and Liu, Yu and Wang, Yun and Xia, Yan and Ou, Yang and Lu, Zhicong and Ma, Sai and Li, Scarlett and Xu, Yingqing},
title = {ProductMeta: An Interactive System for Metaphorical Product Design Ideation with Multimodal Large Language Models},
year = {2025},
isbn = {9798400713941},
publisher = {Association for Computing Machinery},
address = {New York, NY, USA},
url = {https://doi.org/10.1145/3706598.3713935},
doi = {10.1145/3706598.3713935},
booktitle = {Proceedings of the 2025 CHI Conference on Human Factors in Computing Systems},
articleno = {428},
numpages = {24},
location = {
},
series = {CHI '25}
}

@inproceedings{10.1145/3706598.3713818,
author = {Kim, Yewon and Lee, Sung-Ju and Donahue, Chris},
title = {Amuse: Human-AI Collaborative Songwriting with Multimodal Inspirations},
year = {2025},
isbn = {9798400713941},
publisher = {Association for Computing Machinery},
address = {New York, NY, USA},
url = {https://doi.org/10.1145/3706598.3713818},
doi = {10.1145/3706598.3713818},
booktitle = {Proceedings of the 2025 CHI Conference on Human Factors in Computing Systems},
articleno = {187},
numpages = {28},
location = {
},
series = {CHI '25}
}

@inproceedings{10.1145/3290605.3300619,
author = {Frich, Jonas and MacDonald Vermeulen, Lindsay and Remy, Christian and Biskjaer, Michael Mose and Dalsgaard, Peter},
title = {Mapping the Landscape of Creativity Support Tools in HCI},
year = {2019},
isbn = {9781450359702},
publisher = {Association for Computing Machinery},
address = {New York, NY, USA},
url = {https://doi.org/10.1145/3290605.3300619},
doi = {10.1145/3290605.3300619},
booktitle = {Proceedings of the 2019 CHI Conference on Human Factors in Computing Systems},
pages = {1–18},
numpages = {18},
location = {Glasgow, Scotland Uk},
series = {CHI '19}
}

@inproceedings{10.1145/3491102.3501933,
author = {Palani, Srishti and Ledo, David and Fitzmaurice, George and Anderson, Fraser},
title = {”I don’t want to feel like I’m working in a 1960s factory”: The Practitioner Perspective on Creativity Support Tool Adoption},
year = {2022},
isbn = {9781450391573},
publisher = {Association for Computing Machinery},
address = {New York, NY, USA},
url = {https://doi.org/10.1145/3491102.3501933},
doi = {10.1145/3491102.3501933},
booktitle = {Proceedings of the 2022 CHI Conference on Human Factors in Computing Systems},
articleno = {379},
numpages = {18},
location = {New Orleans, LA, USA},
series = {CHI '22}
}

@article{sanosi2018effect,
  title={The effect of Quizlet on vocabulary acquisition},
  author={Sanosi, Abdulaziz B},
  journal={Asian Journal of Education and e-learning},
  volume={6},
  number={4},
  year={2018}
}

@online{duolingo,
  author   = {{Duolingo}},
  title    = {Duolingo},
  year     = {2025},
  url      = {https://www.duolingo.com/},
  note     = {Language learning platform}
}

@inproceedings{10.1145/3613904.3642185,
author = {Xiao, Shishi and Wang, Liangwei and Ma, Xiaojuan and Zeng, Wei},
title = {TypeDance: Creating Semantic Typographic Logos from Image through Personalized Generation},
year = {2024},
isbn = {9798400703300},
publisher = {Association for Computing Machinery},
address = {New York, NY, USA},
url = {https://doi.org/10.1145/3613904.3642185},
doi = {10.1145/3613904.3642185},
booktitle = {Proceedings of the 2024 CHI Conference on Human Factors in Computing Systems},
articleno = {175},
numpages = {18},
location = {Honolulu, HI, USA},
series = {CHI '24}
}

@inproceedings{10.1145/3706598.3713649,
author = {Xu, Xiaotong (Tone) and Konnova, Arina and Gao, Bianca and Peng, Cindy and Vo, Dave and Dow, Steven P.},
title = {Productive vs. Reflective: How Different Ways of Integrating AI into Design Workflows Affect Cognition and Motivation},
year = {2025},
isbn = {9798400713941},
publisher = {Association for Computing Machinery},
address = {New York, NY, USA},
url = {https://doi.org/10.1145/3706598.3713649},
doi = {10.1145/3706598.3713649},
booktitle = {Proceedings of the 2025 CHI Conference on Human Factors in Computing Systems},
articleno = {24},
numpages = {15},
location = {
},
series = {CHI '25}
}

@article{article4,
author = {Ericsson, Karl and Simon, Herbert},
year = {1980},
month = {05},
pages = {215-251},
title = {Verbal Reports as Data},
volume = {87},
journal = {Psychological Review},
doi = {10.1037/0033-295X.87.3.215}
}

@article{Braun01012006,
author = {Virginia Braun and Victoria Clarke},
title = {Using thematic analysis in psychology},
journal = {Qualitative Research in Psychology},
volume = {3},
number = {2},
pages = {77--101},
year = {2006},
publisher = {Routledge},
doi = {10.1191/1478088706qp063oa},
URL = {https://doi.org/10.1191/1478088706qp063oa},
eprint = {https://doi.org/10.1191/1478088706qp063oa}
}

@article{cherry2014quantifying,
  title={Quantifying the creativity support of digital tools through the creativity support index},
  author={Cherry, Erin and Latulipe, Celine},
  journal={ACM Transactions on Computer-Human Interaction (TOCHI)},
  volume={21},
  number={4},
  pages={1--25},
  year={2014},
  publisher={ACM New York, NY, USA}
}

@article{bangor2008empirical,
  title={An empirical evaluation of the system usability scale},
  author={Bangor, Aaron and Kortum, Philip T and Miller, James T},
  journal={Intl. Journal of Human--Computer Interaction},
  volume={24},
  number={6},
  pages={574--594},
  year={2008},
  publisher={Taylor \& Francis}
}

@inproceedings{hart2006nasa,
  title={NASA-task load index (NASA-TLX); 20 years later},
  author={Hart, Sandra G},
  booktitle={Proceedings of the human factors and ergonomics society annual meeting},
  volume={50},
  number={9},
  pages={904--908},
  year={2006},
  organization={Sage publications Sage CA: Los Angeles, CA}
}

@article{mcknight2010mann,
  title={Mann-whitney U test},
  author={McKnight, Patrick E and Najab, Julius},
  journal={The Corsini encyclopedia of psychology},
  pages={1--1},
  year={2010},
  publisher={Wiley Online Library}
}

@article{capanu2006testing,
  title={Testing for preference using a sum of Wilcoxon signed rank statistics},
  author={Capanu, Marinela and Jones, Gregory A and Randles, Ronald H},
  journal={Computational statistics \& data analysis},
  volume={51},
  number={2},
  pages={793--796},
  year={2006},
  publisher={Elsevier}
}

@inproceedings{10.1145/3613905.3651042,
author = {Guo, Jiajing and Mohanty, Vikram and Piazentin Ono, Jorge H and Hao, Hongtao and Gou, Liang and Ren, Liu},
title = {Investigating Interaction Modes and User Agency in Human-LLM Collaboration for Domain-Specific Data Analysis},
year = {2024},
isbn = {9798400703317},
publisher = {Association for Computing Machinery},
address = {New York, NY, USA},
url = {https://doi.org/10.1145/3613905.3651042},
doi = {10.1145/3613905.3651042},
booktitle = {Extended Abstracts of the CHI Conference on Human Factors in Computing Systems},
articleno = {203},
numpages = {9},
location = {Honolulu, HI, USA},
series = {CHI EA '24}
}

@article{guerreiro2023hallucinations,
  title={Hallucinations in large multilingual translation models},
  author={Guerreiro, Nuno M and Alves, Duarte M and Waldendorf, Jonas and Haddow, Barry and Birch, Alexandra and Colombo, Pierre and Martins, Andr{\'e} FT},
  journal={Transactions of the Association for Computational Linguistics},
  volume={11},
  pages={1500--1517},
  year={2023},
  publisher={MIT Press One Broadway, 12th Floor, Cambridge, Massachusetts 02142, USA~…}
}

@inproceedings{10.1145/3613904.3642428,
author = {Leiser, Florian and Eckhardt, Sven and Leuthe, Valentin and Knaeble, Merlin and M\"{a}dche, Alexander and Schwabe, Gerhard and Sunyaev, Ali},
title = {HILL: A Hallucination Identifier for Large Language Models},
year = {2024},
isbn = {9798400703300},
publisher = {Association for Computing Machinery},
address = {New York, NY, USA},
url = {https://doi.org/10.1145/3613904.3642428},
doi = {10.1145/3613904.3642428},
booktitle = {Proceedings of the 2024 CHI Conference on Human Factors in Computing Systems},
articleno = {482},
numpages = {13},
location = {Honolulu, HI, USA},
series = {CHI '24}
}

@inproceedings{10.1145/3706598.3715579,
author = {Prasongpongchai, Thanawit and Pataranutaporn, Pat and Lertsutthiwong, Monchai and Maes, Pattie},
title = {Talk to the Hand: an LLM-powered Chatbot with Visual Pointer as Proactive Companion for On-Screen Tasks},
year = {2025},
isbn = {9798400713941},
publisher = {Association for Computing Machinery},
address = {New York, NY, USA},
url = {https://doi.org/10.1145/3706598.3715579},
doi = {10.1145/3706598.3715579},
booktitle = {Proceedings of the 2025 CHI Conference on Human Factors in Computing Systems},
articleno = {637},
numpages = {16},
location = {
},
series = {CHI '25}
}

@inproceedings{10.1145/3613905.3650912,
author = {Jones, Brennan and Xu, Yan and Li, Qisheng and Scherer, Stefan},
title = {Designing a Proactive Context-Aware AI Chatbot for People's Long-Term Goals},
year = {2024},
isbn = {9798400703317},
publisher = {Association for Computing Machinery},
address = {New York, NY, USA},
url = {https://doi.org/10.1145/3613905.3650912},
doi = {10.1145/3613905.3650912},
booktitle = {Extended Abstracts of the CHI Conference on Human Factors in Computing Systems},
articleno = {104},
numpages = {7},
location = {Honolulu, HI, USA},
series = {CHI EA '24}
}

@article{paul2007critical,
  title={Critical thinking: The art of Socratic questioning},
  author={Paul, Richard and Elder, Linda},
  journal={Journal of developmental education},
  volume={31},
  number={1},
  pages={36},
  year={2007},
  publisher={Appalachian State University d/b/a}
}

@inproceedings{10.1145/3544548.3580672,
author = {Danry, Valdemar and Pataranutaporn, Pat and Mao, Yaoli and Maes, Pattie},
title = {Don’t Just Tell Me, Ask Me: AI Systems that Intelligently Frame Explanations as Questions Improve Human Logical Discernment Accuracy over Causal AI explanations},
year = {2023},
isbn = {9781450394215},
publisher = {Association for Computing Machinery},
address = {New York, NY, USA},
url = {https://doi.org/10.1145/3544548.3580672},
doi = {10.1145/3544548.3580672},
booktitle = {Proceedings of the 2023 CHI Conference on Human Factors in Computing Systems},
articleno = {352},
numpages = {13},
location = {Hamburg, Germany},
series = {CHI '23}
}

@inproceedings{10.1145/3706598.3713748,
author = {Ma, Shuai and Wang, Junling and Zhang, Yuanhao and Ma, Xiaojuan and Wang, April Yi},
title = {DBox: Scaffolding Algorithmic Programming Learning through Learner-LLM Co-Decomposition},
year = {2025},
isbn = {9798400713941},
publisher = {Association for Computing Machinery},
address = {New York, NY, USA},
url = {https://doi.org/10.1145/3706598.3713748},
doi = {10.1145/3706598.3713748},
booktitle = {Proceedings of the 2025 CHI Conference on Human Factors in Computing Systems},
articleno = {585},
numpages = {20},
location = {
},
series = {CHI '25}
}

@inproceedings{10.1145/3341525.3387411,
author = {Wang, Wengran and Rao, Yudong and Zhi, Rui and Marwan, Samiha and Gao, Ge and Price, Thomas W.},
title = {Step Tutor: Supporting Students through Step-by-Step Example-Based Feedback},
year = {2020},
isbn = {9781450368742},
publisher = {Association for Computing Machinery},
address = {New York, NY, USA},
url = {https://doi.org/10.1145/3341525.3387411},
doi = {10.1145/3341525.3387411},
booktitle = {Proceedings of the 2020 ACM Conference on Innovation and Technology in Computer Science Education},
pages = {391–397},
numpages = {7},
location = {Trondheim, Norway},
series = {ITiCSE '20}
}

@inproceedings{10.1145/3613905.3650937,
author = {Xiao, Ruiwei and Hou, Xinying and Stamper, John},
title = {Exploring How Multiple Levels of GPT-Generated Programming Hints Support or Disappoint Novices},
year = {2024},
isbn = {9798400703317},
publisher = {Association for Computing Machinery},
address = {New York, NY, USA},
url = {https://doi.org/10.1145/3613905.3650937},
doi = {10.1145/3613905.3650937},
booktitle = {Extended Abstracts of the CHI Conference on Human Factors in Computing Systems},
articleno = {142},
numpages = {10},
location = {Honolulu, HI, USA},
series = {CHI EA '24}
}

@inproceedings{lin2025sketchflex,
  title={SketchFlex: Facilitating Spatial-Semantic Coherence in Text-to-Image Generation with Region-Based Sketches},
  author={Lin, Haichuan and Ye, Yilin and Xia, Jiazhi and Zeng, Wei},
  booktitle={Proceedings of the 2025 CHI Conference on Human Factors in Computing Systems},
  pages={1--19},
  year={2025}
}

@inproceedings{li2023gligen,
  title={Gligen: Open-set grounded text-to-image generation},
  author={Li, Yuheng and Liu, Haotian and Wu, Qingyang and Mu, Fangzhou and Yang, Jianwei and Gao, Jianfeng and Li, Chunyuan and Lee, Yong Jae},
  booktitle={Proceedings of the IEEE/CVF conference on computer vision and pattern recognition},
  pages={22511--22521},
  year={2023}
}

@misc{lewis2021retrievalaugmentedgenerationknowledgeintensivenlp,
      title={Retrieval-Augmented Generation for Knowledge-Intensive NLP Tasks}, 
      author={Patrick Lewis and Ethan Perez and Aleksandra Piktus and Fabio Petroni and Vladimir Karpukhin and Naman Goyal and Heinrich Küttler and Mike Lewis and Wen-tau Yih and Tim Rocktäschel and Sebastian Riedel and Douwe Kiela},
      year={2021},
      eprint={2005.11401},
      archivePrefix={arXiv},
      primaryClass={cs.CL},
      url={https://arxiv.org/abs/2005.11401}, 
}

@article{zhu2024surveying,
  title={Surveying on English Mobile Learning Among University Students: Current State and Influencing Factors},
  author={ZHU, Haihua and CHEN, Yifan and REN, Yanyan and ZHI, Yuying},
  journal={Sino-US English Teaching},
  volume={21},
  number={3},
  pages={108--119},
  year={2024}
}

@article{bower1969narrative,
  title={Narrative stories as mediators for serial learning},
  author={Bower, Gordon H and Clark, Michal C},
  journal={Psychonomic science},
  volume={14},
  number={4},
  pages={181--182},
  year={1969},
  publisher={Springer}
}

@article{BOWER1972478,
title = {Mnemonic elaboration in multilist learning},
journal = {Journal of Verbal Learning and Verbal Behavior},
volume = {11},
number = {4},
pages = {478-485},
year = {1972},
issn = {0022-5371},
doi = {https://doi.org/10.1016/S0022-5371(72)80030-6},
url = {https://www.sciencedirect.com/science/article/pii/S0022537172800306},
author = {Gordon H. Bower and Judith S. Reitman}
}

@article{higbee1990using,
  title={Using the link mnemonic to remember errands},
  author={Higbee, Kenneth L and Clawson, Charvel and Delano, Lance and Campbell, Sue},
  journal={The Psychological Record},
  volume={40},
  number={3},
  pages={429--436},
  year={1990},
  publisher={Springer}
}

@article{clark1991dual,
  title={Dual coding theory and education},
  author={Clark, James M and Paivio, Allan},
  journal={Educational psychology review},
  volume={3},
  number={3},
  pages={149--210},
  year={1991},
  publisher={Springer}
}

@article{bertsch2007generation,
  title={The generation effect: A meta-analytic review},
  author={Bertsch, Sharon and Pesta, Bryan J and Wiscott, Richard and McDaniel, Michael A},
  journal={Memory \& cognition},
  volume={35},
  number={2},
  pages={201--210},
  year={2007},
  publisher={Springer}
}

@mastersthesis{park2006phonological,
  author  = {Park, Jessica Lynn},
  title   = {The Role of the Phonological Loop When Learning Foreign Vocabulary Using the Keyword Method},
  school  = {University of Dayton},
  year    = {2006},
  type    = {M.A. thesis, Department of Psychology},
  note    = {Graduate Theses and Dissertations, 4833},
  url     = {https://ecommons.udayton.edu/graduate_theses/4833/},
  urldate = {2025-09-11}
}

@mastersthesis{consiglio2018keyword,
  author    = {Consiglio, Jessica Anna},
  title     = {The effectiveness of the keyword method on foreign language vocabulary for students with learning disabilities},
  school    = {Rowan University},
  year      = {2018},
  type      = {M.A. thesis (Special Education)},
  note      = {Theses and Dissertations, No. 2581},
  url       = {https://rdw.rowan.edu/etd/2581/},
  urldate   = {2025-09-11}
}

@article{article9,
author = {Rodríguez, M. and Sadoski, Mark},
year = {2000},
month = {06},
pages = {385-412},
title = {Effects of rote, context, keyword, and context/keyword methods on retention of vocabulary in EFL classrooms},
volume = {50},
journal = {Language Learning}
}

@article{kernan1999stratified,
  title={Stratified randomization for clinical trials},
  author={Kernan, Walter N and Viscoli, Catherine M and Makuch, Robert W and Brass, Lawrence M and Horwitz, Ralph I},
  journal={Journal of clinical epidemiology},
  volume={52},
  number={1},
  pages={19--26},
  year={1999},
  publisher={Elsevier}
}

@article{thomas1996learning,
  title={Learning by the keyword mnemonic: Looking for long-term benefits.},
  author={Thomas, Margaret H and Wang, Alvin Y},
  journal={Journal of Experimental Psychology: Applied},
  volume={2},
  number={4},
  pages={330},
  year={1996},
  publisher={American Psychological Association}
}

@online{bell2017lindamood,
  author  = {Bell, E.},
  title   = {Lindamood-Bell Learning Process},
  year    = {2017},
  url     = {https://www.readingrockets.org/topics/vocabulary/articles/imagery-language-connection-vocabulary-skills},
  urldate = {2025-09-11},
  note    = {Reprints notice on Reading Rockets}
}

@article{article10,
author = {Harahap, Melisa and Daulay, Sholihatul and Dalimunte, Muhammad},
year = {2025},
month = {04},
pages = {705},
title = {Exploring the Use of Semantic Mapping in English Language Teaching Classroom},
volume = {9},
journal = {Scope : Journal of English Language Teaching},
doi = {10.30998/scope.v9i2.22643}
}

@article{margosein1982effects,
  title={The effects of instruction using semantic mapping on vocabulary and comprehension},
  author={Margosein, Carol M and Pascarella, Ernest T and Pflaum, Susanna W},
  journal={The Journal of early adolescence},
  volume={2},
  number={2},
  pages={185--194},
  year={1982},
  publisher={Sage Publications Sage CA: Thousand Oaks, CA}
}

@book{heimlich1986semantic,
  title={Semantic Mapping: Classroom Applications. Reading Aids Series, IRA Service Bulletin.},
  author={Heimlich, Joan E and Pittelman, Susan D},
  year={1986},
  publisher={ERIC}
}

@article{kornell2009optimising,
  title={Optimising learning using flashcards: Spacing is more effective than cramming},
  author={Kornell, Nate},
  journal={Applied Cognitive Psychology: The Official Journal of the Society for Applied Research in Memory and Cognition},
  volume={23},
  number={9},
  pages={1297--1317},
  year={2009},
  publisher={Wiley Online Library}
}

@article{nugroho2012improving,
  title={IMPROVING STUDENTS’VOCABULARY MASTERY USING FLASHCARDS},
  author={Nugroho, Yosephus Setyo and Nurkamto, Joko and Sulistyowati, Hefy},
  journal={English Education},
  volume={1},
  number={1},
  year={2012}
}

@article{yip2006online,
  title={Online vocabulary games as a tool for teaching and learning English vocabulary},
  author={Yip, Florence WM and Kwan, Alvin CM},
  journal={Educational media international},
  volume={43},
  number={3},
  pages={233--249},
  year={2006},
  publisher={Taylor \& Francis}
}

@inproceedings{rojabi2022kahoot,
  title={Kahoot, is it fun or unfun? Gamifying vocabulary learning to boost exam scores, engagement, and motivation},
  author={Rojabi, Ahmad Ridho and Setiawan, Slamet and Munir, Ahmad and Purwati, Oikurema and Safriyani, Rizka and Hayuningtyas, Nina and Khodijah, Siti and Amumpuni, Rengganis Siwi},
  booktitle={Frontiers in Education},
  volume={7},
  pages={939884},
  year={2022},
  organization={Frontiers Media SA}
}

@inproceedings{10.1145/3605390.3605395,
author = {Cantone, Andrea Antonio and Francese, Rita and Sais, Raffaele and Santosuosso, Otino Pio and Sepe, Aurelio and Spera, Simone and Tortora, Genoveffa and Vitiello, Giuliana},
title = {Contextualized Experiential Language Learning in the Metaverse},
year = {2023},
isbn = {9798400708060},
publisher = {Association for Computing Machinery},
address = {New York, NY, USA},
url = {https://doi.org/10.1145/3605390.3605395},
doi = {10.1145/3605390.3605395},
booktitle = {Proceedings of the 15th Biannual Conference of the Italian SIGCHI Chapter},
articleno = {20},
numpages = {7},
location = {Torino, Italy},
series = {CHItaly '23}
}

@inproceedings{10.1145/3613905.3648107,
author = {Karaosmanoglu, Sukran and Fittschen, Elisabeth L and Eyicalis, Hande and Kraus, David and Nickelmann, Henrik and Tomko, Anna and Steinicke, Frank},
title = {Language of Zelda: Facilitating Language Learning Practices Using ChatGPT},
year = {2024},
isbn = {9798400703317},
publisher = {Association for Computing Machinery},
address = {New York, NY, USA},
url = {https://doi.org/10.1145/3613905.3648107},
doi = {10.1145/3613905.3648107},
booktitle = {Extended Abstracts of the CHI Conference on Human Factors in Computing Systems},
articleno = {635},
numpages = {5},
location = {Honolulu, HI, USA},
series = {CHI EA '24}
}

@article{sahadevan2021imagery,
  title={Imagery-based strategies for memory for associations},
  author={Sahadevan, Shrida S and Chen, Yvonne Y and Caplan, Jeremy B},
  journal={Memory},
  volume={29},
  number={10},
  pages={1275--1295},
  year={2021},
  publisher={Taylor \& Francis}
}

@article{roediger1980effectiveness,
  title={The effectiveness of four mnemonics in ordering recall.},
  author={Roediger, Henry L},
  journal={Journal of Experimental Psychology: Human Learning and Memory},
  volume={6},
  number={5},
  pages={558},
  year={1980},
  publisher={American Psychological Association}
}

@article{bower1970imagery,
  title={Imagery as a relational organizer in associative learning},
  author={Bower, Gordon H},
  journal={Journal of verbal learning and verbal behavior},
  volume={9},
  number={5},
  pages={529--533},
  year={1970},
  publisher={Elsevier}
}

@article{bugelski1968images,
  title={Images as mediators in one-trial paired-associate learning: II. Self-timing in successive lists.},
  author={Bugelski, Bergen Richard},
  journal={Journal of Experimental Psychology},
  volume={77},
  number={2},
  pages={328},
  year={1968},
  publisher={American Psychological Association}
}

@article{bellezza1981mnemonic,
  title={Mnemonic devices: Classification, characteristics, and criteria},
  author={Bellezza, Francis S},
  journal={Review of Educational Research},
  volume={51},
  number={2},
  pages={247--275},
  year={1981},
  publisher={Sage Publications Sage CA: Thousand Oaks, CA}
}

@inproceedings{shi2025brickify,
  title={Brickify: Enabling Expressive Design Intent Specification through Direct Manipulation on Design Tokens},
  author={Shi, Xinyu and Wang, Yinghou and Rossi, Ryan and Zhao, Jian},
  booktitle={Proceedings of the 2025 CHI Conference on Human Factors in Computing Systems},
  pages={1--20},
  year={2025}
}

@inproceedings{10.1145/3706598.3713381,
author = {Wang, Xiyuan and Cao, Yi-Fan and Xiong, Junjie and Chen, Sizhe and Li, Wenxuan and Zhang, Junjie and Li, Quan},
title = {ClueCart: Supporting Game Story Interpretation and Narrative Inference from Fragmented Clues},
year = {2025},
isbn = {9798400713941},
publisher = {Association for Computing Machinery},
address = {New York, NY, USA},
url = {https://doi.org/10.1145/3706598.3713381},
doi = {10.1145/3706598.3713381},
booktitle = {Proceedings of the 2025 CHI Conference on Human Factors in Computing Systems},
articleno = {410},
numpages = {26},
location = {
},
series = {CHI '25}
}

@article{mosqueira2023human,
  title={Human-in-the-loop machine learning: a state of the art},
  author={Mosqueira-Rey, Eduardo and Hern{\'a}ndez-Pereira, Elena and Alonso-R{\'\i}os, David and Bobes-Bascar{\'a}n, Jos{\'e} and Fern{\'a}ndez-Leal, {\'A}ngel},
  journal={Artificial Intelligence Review},
  volume={56},
  number={4},
  pages={3005--3054},
  year={2023},
  publisher={Springer}
}

@article{wu2022survey,
  title={A survey of human-in-the-loop for machine learning},
  author={Wu, Xingjiao and Xiao, Luwei and Sun, Yixuan and Zhang, Junhang and Ma, Tianlong and He, Liang},
  journal={Future Generation Computer Systems},
  volume={135},
  pages={364--381},
  year={2022},
  publisher={Elsevier}
}

@article{hein1991constructivist,
  title={Constructivist learning theory},
  author={Hein, George E},
  journal={Institute for Inquiry},
  volume={14},
  year={1991}
}

@article{bada2015constructivism,
  title={Constructivism learning theory: A paradigm for teaching and learning},
  author={Bada, Steve Olusegun and Olusegun, Steve},
  journal={Journal of Research \& Method in Education},
  volume={5},
  number={6},
  pages={66--70},
  year={2015}
}

@article{narayan2013constructivism,
  title={Constructivism—Constructivist learning theory.},
  author={Narayan, Ratna and Rodriguez, Cynthia and Araujo, Juan and Shaqlaih, Ali and Moss, Glenda},
  year={2013},
  publisher={IAP Information Age Publishing}
}

@inproceedings{leong2024putting,
  title={Putting things into context: Generative AI-enabled context personalization for vocabulary learning improves learning motivation},
  author={Leong, Joanne and Pataranutaporn, Pat and Danry, Valdemar and Perteneder, Florian and Mao, Yaoli and Maes, Pattie},
  booktitle={Proceedings of the 2024 CHI Conference on Human Factors in Computing Systems},
  pages={1--15},
  year={2024}
}

@inproceedings{chen2024retassist,
  title={RetAssist: Facilitating vocabulary learners with generative images in story retelling practices},
  author={Chen, Qiaoyi and Liu, Siyu and Huang, Kaihui and Wang, Xingbo and Ma, Xiaojuan and Zhu, Junkai and Peng, Zhenhui},
  booktitle={Proceedings of the 2024 ACM Designing Interactive Systems Conference},
  pages={2019--2036},
  year={2024}
}

@article{atchley2024human,
  title={Human and AI collaboration in the higher education environment: opportunities and concerns},
  author={Atchley, Paul and Pannell, Hannah and Wofford, Kaelyn and Hopkins, Michael and Atchley, Ruth Ann},
  journal={Cognitive research: principles and implications},
  volume={9},
  number={1},
  pages={20},
  year={2024},
  publisher={Springer}
}

@article{molenaar2024human,
  title={Human-AI collaboration in education: The hybrid future},
  author={Molenaar, Inge},
  year={2024},
  publisher={Nijmegen: Radboud University}
}

@article{dang2025human,
  title={Human--AI collaborative learning in mixed reality: Examining the cognitive and socio-emotional interactions},
  author={Dang, Belle and Huynh, Luna and Gul, Faaiz and Ros{\'e}, Carolyn and J{\"a}rvel{\"a}, Sanna and Nguyen, Andy},
  journal={British Journal of Educational Technology},
  year={2025},
  publisher={Wiley Online Library}
}

@inproceedings{attygalle2025text,
  title={Text-to-image generation for vocabulary learning using the keyword method},
  author={Attygalle, Nuwan T and Kljun, Matja{\v{z}} and Quigley, Aaron and {\v{C}}opi{\v{c}} Pucihar, Klen and Grubert, Jens and Biener, Verena and Leiva, Luis A and Yoneyama, Juri and Toniolo, Alice and Miguel, Angela and others},
  booktitle={Proceedings of the 30th International Conference on Intelligent User Interfaces},
  pages={1381--1397},
  year={2025}
}

@article{pressley1982mnemonic,
  title={The mnemonic keyword method},
  author={Pressley, Michael and Levin, Joel R and Delaney, Harold D},
  journal={Review of Educational Research},
  volume={52},
  number={1},
  pages={61--91},
  year={1982},
  publisher={Sage Publications Sage CA: Thousand Oaks, CA}
}

@inproceedings{kang2025phonitale,
  title={Phonitale: Phonologically grounded mnemonic generation for typologically distant language pairs},
  author={Kang, Sana and Gwon, Myeongseok and Kwon, Su Young and Lee, Jaewook and Lan, Andrew and Raj, Bhiksha and Singh, Rita},
  booktitle={Proceedings of the 2025 Conference on Empirical Methods in Natural Language Processing},
  pages={25572--25604},
  year={2025}
}

@book{nation2001learning,
  title={Learning vocabulary in another language},
  author={Nation, Ian SP and Nation, ISP},
  volume={10},
  year={2001},
  publisher={Cambridge university press Cambridge}
}

@book{dekeyser2007practice,
  title={Practice in a second language: Perspectives from applied linguistics and cognitive psychology},
  author={DeKeyser, Robert},
  year={2007},
  publisher={Cambridge University Press}
}

@article{article123,
author = {Barcroft, Joe},
year = {2002},
month = {12},
pages = {323 - 363},
title = {Semantic and Structural Elaboration in L2 Lexical Acquisition},
volume = {52},
journal = {Language Learning},
doi = {10.1111/0023-8333.00186}
}

@article{article1234,
author = {Al-Seghayer, Khalid},
year = {2001},
month = {01},
pages = {},
title = {The effect of multimedia annotation modes on L2 vocabulary acquisition: A comparative study},
volume = {5},
journal = {Language Learning and Technology},
doi = {10.64152/10125/25117}
}

@article{mizuno2000test,
  title={A test of the effectiveness of the Low-First Spaced Learning Method applied to CAI},
  author={MIZUNO, Rika},
  journal={Educational technology research},
  volume={23},
  number={1-2},
  pages={35--44},
  year={2000},
  publisher={Japan Society for Educational Technology}
}

@article{lei2022learning,
  title={Learning English vocabulary from word cards: A research synthesis},
  author={Lei, Yuanying and Reynolds, Barry Lee},
  journal={Frontiers in Psychology},
  volume={13},
  pages={984211},
  year={2022},
  publisher={Frontiers Media SA}
}

@article{boroughani2023mobile,
  title={Mobile-assisted academic vocabulary learning with digital flashcards: Exploring the impacts on university students’ self-regulatory capacity},
  author={Boroughani, Tahereh and Behshad, Nastaran and Xodabande, Ismail},
  journal={Frontiers in psychology},
  volume={14},
  pages={1112429},
  year={2023},
  publisher={Frontiers Media SA}
}

@article{article12345,
author = {Chukharev-Hudilainen, Evgeny and Klepikova, Tatiana},
year = {2016},
month = {08},
pages = {334-354},
title = {The effectiveness of computer-based spaced repetition in foreign language vocabulary instruction: A double-blind study},
volume = {33},
journal = {CALICO Journal},
doi = {10.1558/cj.v33i3.26055}
}

@book{ellis1994study,
  title={The study of second language acquisition},
  author={Ellis, Rod},
  year={1994},
  publisher={Oxford University}
}

@article{article123456,
author = {Norris, John and Ortega, Lourdes},
year = {2002},
month = {12},
pages = {417 - 528},
title = {Effectiveness of L2 Instruction: A Research Synthesis and Quantitative Meta-analysis},
volume = {50},
journal = {Language Learning},
doi = {10.1111/0023-8333.00136}
}

@article{article1234567,
author = {Lessard-Clouston, Michael},
year = {2013},
month = {05},
pages = {287-304},
title = {Word Lists for Vocabulary Learning and Teaching},
volume = {24},
journal = {The CATESOL Journal},
doi = {10.5070/B5.36167}
}

@article{dang2025applications,
  title={Applications of word lists in second language learning and teaching},
  author={Dang, Thi Ngoc Yen and Webb, Stuart},
  journal={Language Teaching},
  pages={1--21},
  year={2025},
  publisher={Cambridge University Press}
}

@article{AKBARIAN2020102261,
title = {EFL learners’ lexical availability: Exploring frequency, exposure, and vocabulary level},
journal = {System},
volume = {91},
pages = {102261},
year = {2020},
issn = {0346-251X},
doi = {https://doi.org/10.1016/j.system.2020.102261},
url = {https://www.sciencedirect.com/science/article/pii/S0346251X19303781},
author = {Is’haaq Akbarian and Fatemeh Farajollahi and Rosa María {Jiménez Catalán}}
}

@article{Nakata_2008, 
title={English vocabulary learning with word lists, word cards and computers: implications from cognitive psychology research for optimal spaced learning}, 
volume={20}, 
DOI={10.1017/S0958344008000219}, 
number={1}, journal={ReCALL}, 
author={Nakata, Tatsuya}, 
year={2008}, 
pages={3–20}
}

@inbook{inbook,
author = {Elgort, Irina and Nation, Paul},
year = {2010},
month = {01},
pages = {89-104},
title = {Vocabulary Learning in a Second Language: Familiar Answers to New Questions},
isbn = {978-1-349-31287-0},
journal = {Conceptualising 'Learning' in Applied Linguistics},
doi = {10.1057/9780230289772_6}
}

@inproceedings{10.1145/2939672.2939850,
author = {Reddy, Siddharth and Labutov, Igor and Banerjee, Siddhartha and Joachims, Thorsten},
title = {Unbounded Human Learning: Optimal Scheduling for Spaced Repetition},
year = {2016},
isbn = {9781450342322},
publisher = {Association for Computing Machinery},
address = {New York, NY, USA},
url = {https://doi.org/10.1145/2939672.2939850},
doi = {10.1145/2939672.2939850},
booktitle = {Proceedings of the 22nd ACM SIGKDD International Conference on Knowledge Discovery and Data Mining},
pages = {1815–1824},
numpages = {10},
location = {San Francisco, California, USA},
series = {KDD '16}
}

@inproceedings{do2025paige,
  title={PAIGE: Examining learning outcomes and experiences with personalized AI-generated educational podcasts},
  author={Do, Tiffany D and Shafqat, Usama Bin and Ling, Elsie and Sarda, Nikhil},
  booktitle={Proceedings of the 2025 CHI Conference on Human Factors in Computing Systems},
  pages={1--12},
  year={2025}
}

@inproceedings{xu2025advancing,
  title={Advancing problem-based learning with clinical reasoning for improved differential diagnosis in medical education},
  author={Xu, Yuansong and Shao, Yuheng and Dong, Jiahe and Shi, Shaohan and Jiang, Chang and Li, Quan},
  booktitle={Proceedings of the 2025 CHI Conference on Human Factors in Computing Systems},
  pages={1--32},
  year={2025}
}

@article{mcmorris2018cognitive,
  title={Cognitive fatigue effects on physical performance: A systematic review and meta-analysis},
  author={McMorris, Terry and Barwood, Martin and Hale, Beverley J and Dicks, Matt and Corbett, Jo},
  journal={Physiology \& behavior},
  volume={188},
  pages={103--107},
  year={2018},
  publisher={Elsevier}
}

@article{azabdaftari2012comparing,
  title={Comparing vocabulary learning of EFL learners by using two different strategies: Mobile learning vs. flashcards.},
  author={Azabdaftari, Behrooz and Mozaheb, Mohammad Amin},
  journal={The Eurocall Review},
  volume={20},
  number={2},
  pages={47--59},
  year={2012},
  publisher={ERIC}
}

@article{chen2012effects,
  title={Effects of presentation mode on mobile language learning: A performance efficiency perspective},
  author={Chen, I-Jung and Chang, Chi-Cheng and Yen, Jung-Chuan},
  journal={Australasian Journal of Educational Technology},
  volume={28},
  number={1},
  year={2012}
}

@inproceedings{arakawa2022vocabencounter,
  title={VocabEncounter: NMT-powered vocabulary learning by presenting computer-generated usages of foreign words into users’ daily lives},
  author={Arakawa, Riku and Yakura, Hiromu and Kobayashi, Sosuke},
  booktitle={Proceedings of the 2022 CHI conference on human factors in computing systems},
  pages={1--21},
  year={2022}
}

@inproceedings{peng2023storyfier,
  title={Storyfier: Exploring vocabulary learning support with text generation models},
  author={Peng, Zhenhui and Wang, Xingbo and Han, Qiushi and Zhu, Junkai and Ma, Xiaojuan and Qu, Huamin},
  booktitle={Proceedings of the 36th Annual ACM Symposium on User Interface Software and Technology},
  pages={1--16},
  year={2023}
}

@inproceedings{lee2024open,
  title={Open sesame? Open salami! Personalizing vocabulary assessment-intervention for children via pervasive profiling and bespoke storybook generation},
  author={Lee, Jungeun and Yoon, Suwon and Lee, Kyoosik and Jeong, Eunae and Cho, Jae-Eun and Park, Wonjeong and Yim, Dongsun and Hwang, Inseok},
  booktitle={Proceedings of the 2024 CHI Conference on Human Factors in Computing Systems},
  pages={1--32},
  year={2024}
}

@inproceedings{zhao2024language,
  title={Language urban odyssey: A serious game for enhancing second language acquisition through large language models},
  author={Zhao, Yijun and Pan, Jiangyu and Dong, Yan and Dong, Tianshu and Wang, Guanyun and Ying, Fangtian and Shen, Qihang and Cao, Jiacheng},
  booktitle={Extended Abstracts of the CHI Conference on Human Factors in Computing Systems},
  pages={1--7},
  year={2024}
}

@article{ngo2024use,
  title={The use of ChatGPT for vocabulary acquisition: A literature review},
  author={Ngo, Trinh},
  journal={Available at SSRN 5059052},
  year={2024}
}
\onecolumn
\appendix
\clearpage
\section{Formative Study System}
\label{fs_system}
\begin{figure*}[h]
    \centering
    \includegraphics[width=\linewidth]{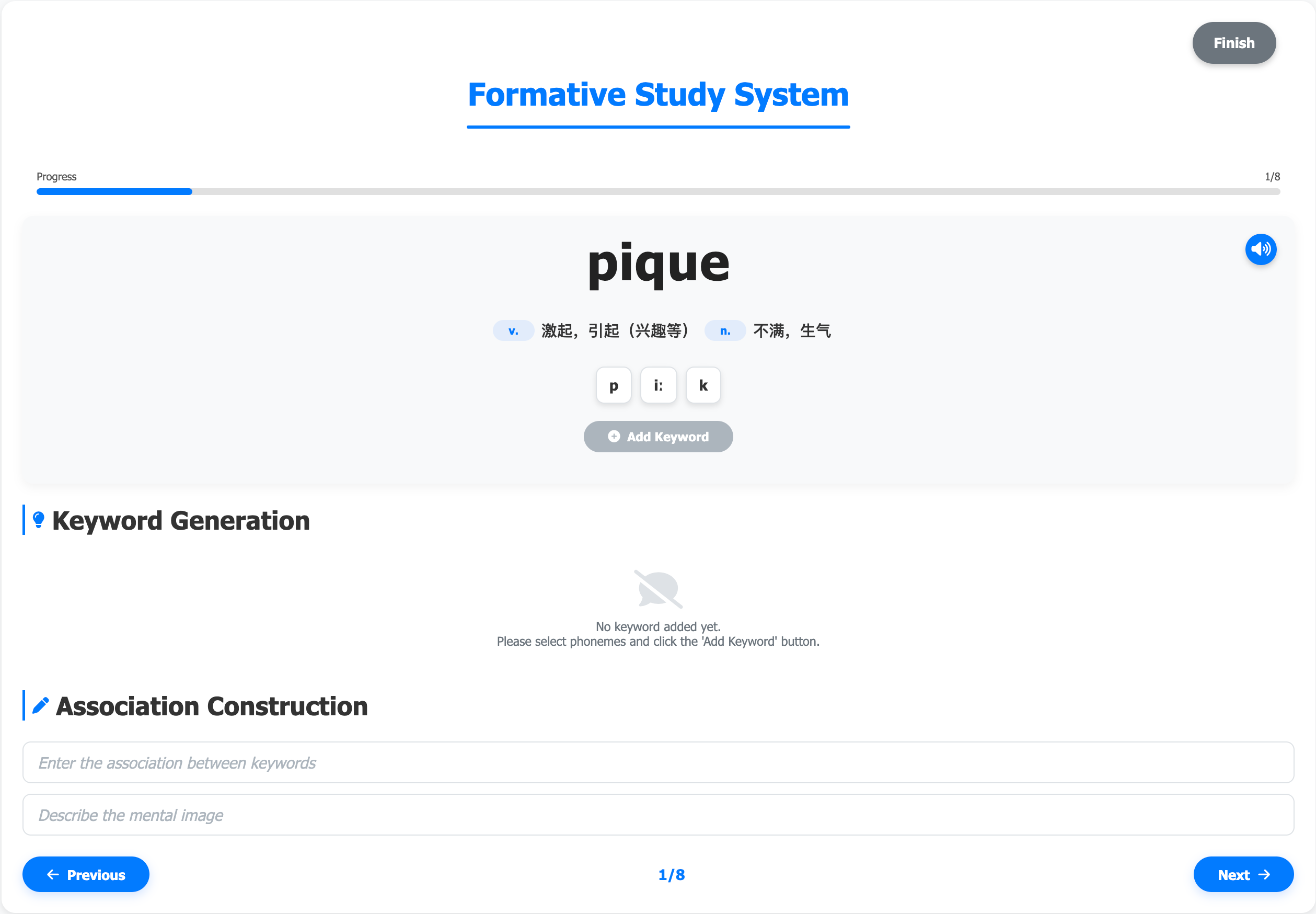}
    \caption{User interface of the experimental system. The interface displays the target vocabulary item with its definition and pronunciation, and provides input fields for participants to record their self-generated keywords and association.}
    \label{fig:formative_study_system}
\end{figure*}

\newpage
\section{CODE BOOK}
\label{code_book}


\begin{table*}[h]
\caption{The codebook in formative study.}
\label{tab:coding_scheme}
\centering
\begin{scriptsize}
\begin{tabular}{p{2.5cm} p{4.5cm} p{8cm}}
\toprule
\textbf{Theme} & \textbf{Category} & \textbf{Code} \\
\midrule

\multirow{15}{=}{Cognitive Workflow} 
    & \multirow{5}{=}{\textbf{Keyword Selection}} 
      & Phonological segmentation of target word \\*
    & & L1 phonological mapping attempts \\*
    & & Balancing phonetic similarity and concreteness \\*
    & & Meaning-guided keyword brainstorming \\*
    & & Iterative keyword replacement and refinement \\
\cmidrule{2-3}

    & \multirow{5}{=}{\textbf{Association Construction}} 
      & Integrating multiple keywords into one scenario \\*
    & & Using narrative or story structures \\*
    & & Experimenting with semantic relation types \\*
    & & Evaluating coherence and memorability of associations \\*
    & & Associative reconstruction after dead-ends \\
\cmidrule{2-3}

    & \multirow{5}{=}{\textbf{Image Formation}} 
      & Mapping keywords and meanings to visual elements \\*
    & & Spatial and structural arrangement of the scene \\*
    & & Visual simplification versus clutter management \\*
    & & Iterative refinement of mental imagery \\*
    & & Immersive elaboration of the imagined scene \\
\cmidrule{2-3}

    & \multirow{3}{=}{\textbf{Iterative and Non-linear Refinement}} 
      & Backtracking to earlier cognitive stages \\*
    & & Dynamic adjustment of strategies to optimize outcomes \\*
    & & Cyclical progression between keywords, associations, and images \\
\bottomrule

\end{tabular}
\end{scriptsize}
\end{table*}

\begin{table*}[h]
\ContinuedFloat
\caption{The codebook in formative study. (continued)}
\centering
\begin{scriptsize}
\begin{tabular}{p{2.5cm} p{4.5cm} p{8cm}}
\toprule
\textbf{Theme} & \textbf{Category} & \textbf{Code} \\
\midrule

\multirow{18}{=}{Challenge} 
    & \multirow{3}{=}{\textbf{Phonological Keyword Difficulty}} 
      & Lack of close phonological matches in L1 \\*
    & & Unfamiliar or non-word keyword candidates \\*
    & & Limited phonological search strategies \\
\cmidrule{2-3}

    & \multirow{3}{=}{\textbf{Semantic Understanding Difficulty}} 
      & Superficial dictionary-based understanding \\*
    & & Lack of concrete usage scenarios \\*
    & & Semantic vagueness or conflation \\
\cmidrule{2-3}

    & \multirow{3}{=}{\textbf{Cognitive Overload}} 
      & Overwhelmed by multiple candidate associations \\*
    & & Managing many elements simultaneously \\*
    & & Uncertainty about how to proceed \\
\cmidrule{2-3}

    & \multirow{3}{=}{\textbf{Quality Uncertainty}} 
      & Lack of evaluation criteria for associations \\*
    & & Post-hoc realization of weak associations \\*
    & & Reliance on “sounds okay” heuristics \\
\cmidrule{2-3}

    & \multirow{3}{=}{\textbf{Image Mapping Difficulty}} 
      & Unclear visual representation of abstract elements \\*
    & & Overcrowded or cluttered mental scenes \\*
    & & Fragmented or disjoint imagery \\
\cmidrule{2-3}

    & \multirow{3}{=}{\textbf{Recall Path Disrupted}} 
      & Remembering the image but not the word form \\*
    & & Losing the link between image and keyword \\*
    & & Forgetting the reasoning chain \\
\midrule

\multirow{16}{=}{MLLM Consideration} 
    & \multirow{4}{=}{\textbf{Over-Reliance on MLLMs}} 
      & Bypassing self-generation after initial success \\*
    & & Reduced time-on-task and cognitive engagement \\*
    & & Diminished personalization of generated content \\*
    & & Delegating step-by-step reasoning to the model \\
\cmidrule{2-3}

    & \multirow{4}{=}{\textbf{Low-Quality or Uncontrolled Generation}} 
      & Generating non-native or unfamiliar keywords \\*
    & & Misaligned balancing of phonetic and semantic constraints \\*
    & & Need for intensive post-filtering by learners \\*
    & & Erosion of learner confidence in model suggestions \\
\cmidrule{2-3}

    & \multirow{4}{=}{\textbf{Intent Articulation Difficulty}} 
      & Difficulty verbalizing multi-element associative logic \\*
    & & Under-specifying constraints in prompts \\*
    & & Divergent or incoherent associative outputs \\*
    & & Iterative reformulation of prompts to clarify intent \\
\cmidrule{2-3}

    & \multirow{4}{=}{\textbf{Image Alignment Difficulty}} 
      & Text–image mismatch despite detailed prompting \\*
    & & Iterative prompting with slow convergence \\*
    & & Extraneous elements introduced by the model \\*
    & & Conflict between internal imagery and external visuals \\
\bottomrule

\end{tabular}
\end{scriptsize}
\end{table*}

\clearpage
\section{USAGE SCENARIO}
\label{usage_scenario}
\begin{figure*}[h]
    \centering
    \includegraphics[width=\textwidth]{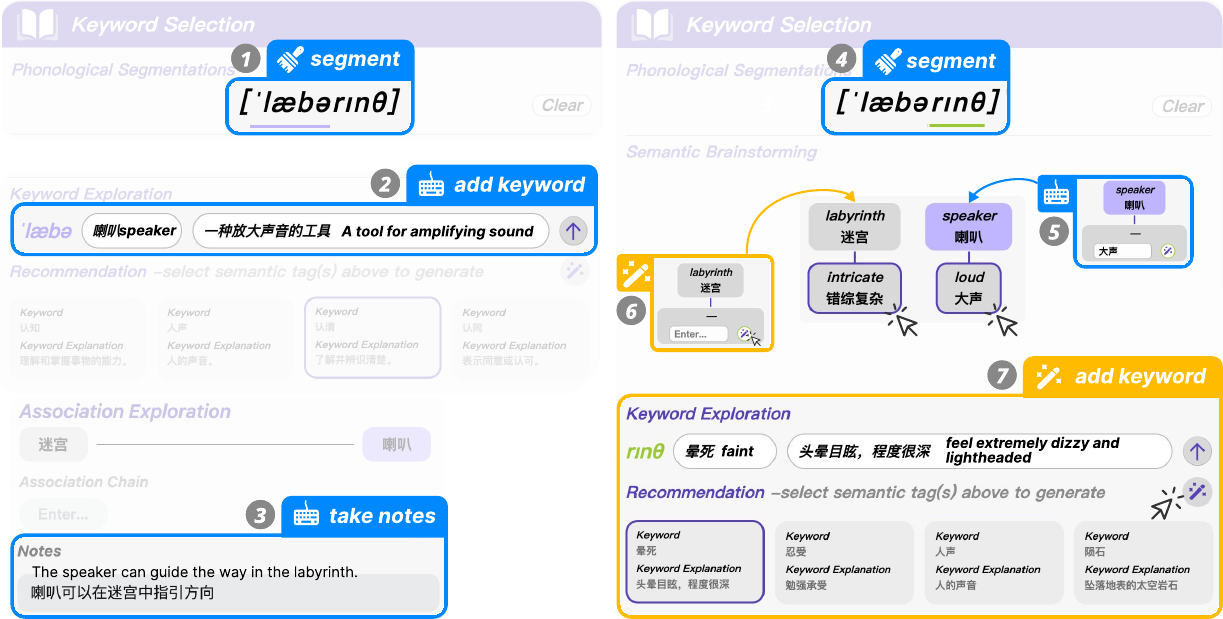}
    \caption{1. Select \includegraphics[page=1, height=0.7em]{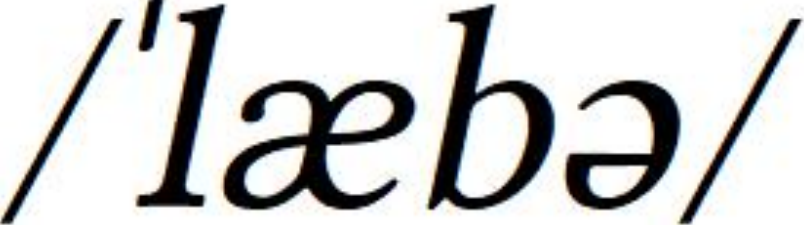} as initial syllable segment. 2. Add keyword ``\thinspace\raisebox{-0.15em}{\includegraphics[page=1, height=0.9em]{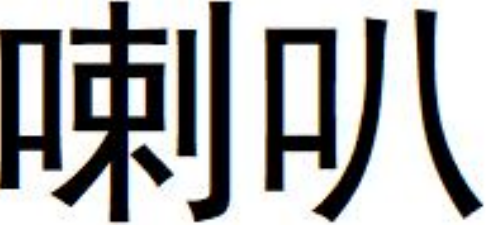}}'' (speaker) to this segment. 3. Take notes in the association map. 4. Select remaining syllable segment \includegraphics[page=1, height=0.7em]{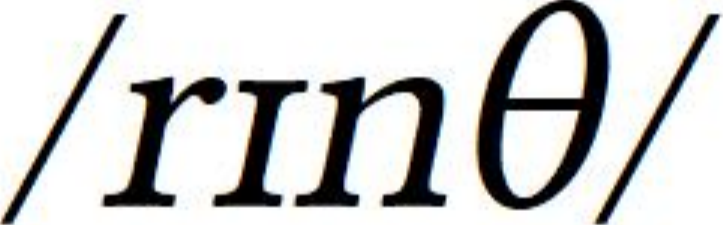}. 5-6. Add semantic nodes to two keywords: \textit{loud} to \textit{speaker} and \textit{intricate} to \textit{labyrinth}. 7. Request heuristic keyword suggestions for \includegraphics[page=1, height=0.7em]{Figure/rinth-yinbiao.pdf}.}
    \label{fig:scenario-1}
\end{figure*}

\begin{figure*}[h]
    \centering
    \includegraphics[width=\textwidth]{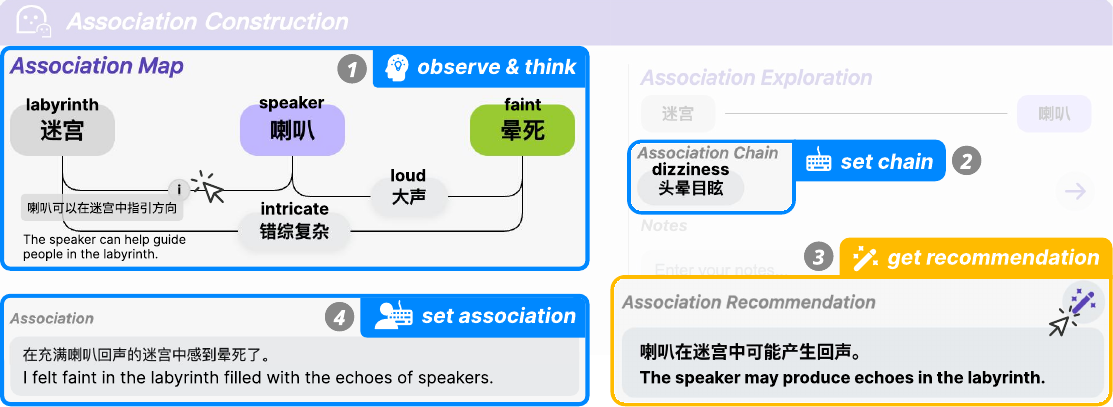}
    \caption{1. Observe and think keywords and notes in Association Map. 2. Set Association Chain to the target word. 3. Get heuristic prompt recommendation for the association. 4. Set final association based on the recommendation.}
    \label{fig:scenario-2}
\end{figure*}

\begin{figure*}[t]
    \centering

    \includegraphics[width=\linewidth]{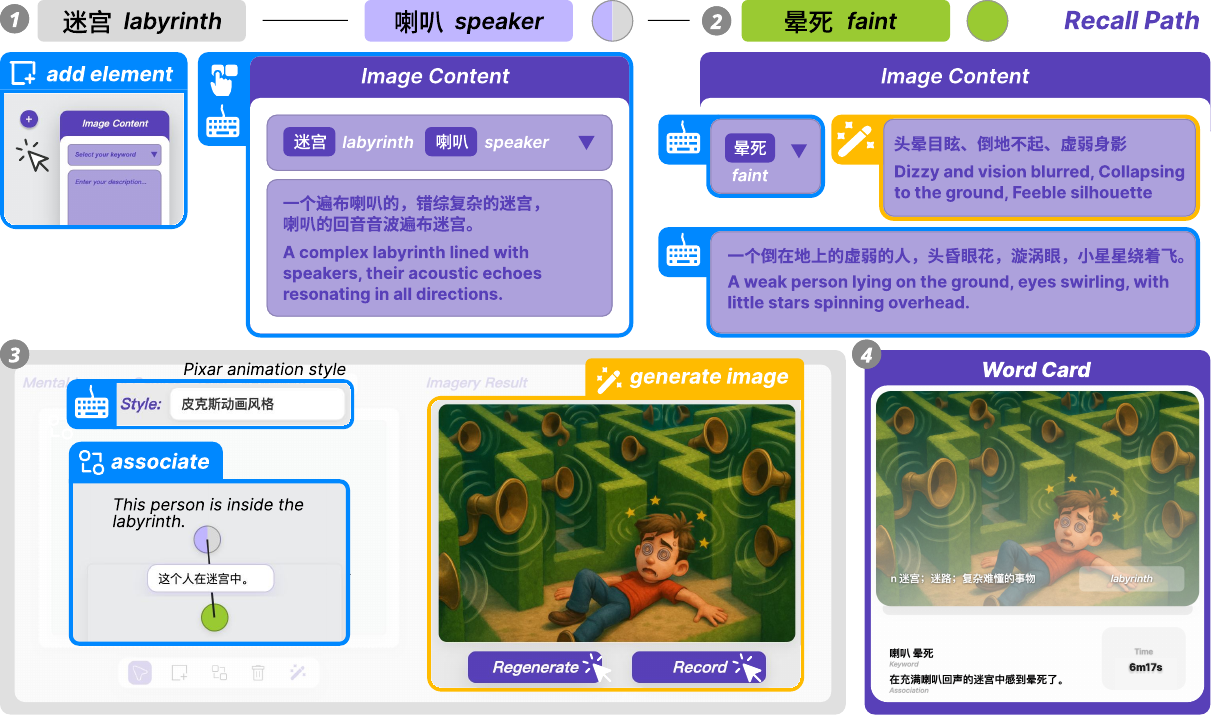}
    \caption{1. Add the first element representing ``labyrinth'' and ``speaker'' on the mental imagery canvas, and describe them in Image Content. 2. Place another faint element on the canvas and add its description. 3. Associate two elements and edit their relationship. Choose ``Pixar animation style'' click on ``Regenerate'' button to produce the image. 4. Finally, record the Word Card.}
    \label{fig:scenario-3}
\end{figure*}

\clearpage
\twocolumn
\par Yang is a second-year undergraduate student majoring in computer science. A native Chinese speaker, he is currently preparing for the TOEFL exam. He was first introduced to the keyword method for vocabulary learning in high school and found it fairly effective. However, due to limited imaginative capacity and the absence of systematic guidance, he often struggled to form coherent and concrete associations when applying the method in practice. To improve the efficiency of memorizing complex vocabulary, he turned to \textit{WordCraft} as a learning aid. The following scenario illustrates how he used the system to learn the word ``labyrinth''.

\par Upon accessing the system, Yang searched for the target word ``labyrinth'', reviewed its Chinese definition and illustrative English sentences, and listened to its standard pronunciation via the audio function. Among the multiple senses of the word, he chose the most common one ``maze'' as his memory target.

\par \textbf{\textit{Keyword Selcetion.}} In the keyword selection step, Yang began by brushing over the initial syllable segment \includegraphics[page=1, height=0.7em]{Figure/laba-yinbiao.pdf}, guided by his phonetic intuition (\cref{fig:scenario-1} \circlenum{1}). Noticing its close resemblance to the pronunciation of the Chinese word ``\thinspace\raisebox{-0.15em}{\includegraphics[page=1, height=0.9em]{Figure/Laba.pdf}}'' (speaker), he formed a vivid mental image and associated it with the idea of a loudspeaker being used in a labyrinth to transmit sound or provide navigational guidance. He thus selected ``\thinspace\raisebox{-0.15em}{\includegraphics[page=1, height=0.9em]{Figure/Laba.pdf}}'' as his first keyword (\cref{fig:scenario-1} \circlenum{2}) and manually linked it to ``\thinspace\raisebox{-0.15em}{\includegraphics[page=1, height=0.9em]{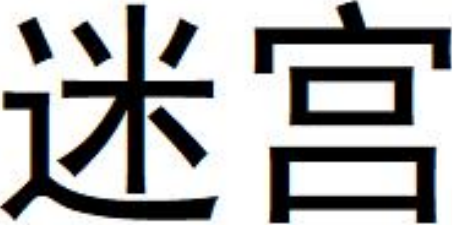}}\thinspace'' (maze) in the association map, adding the note: ``\textit{The speaker can guide the way in the labyrinth}'' (\cref{fig:scenario-1} \circlenum{3}).

\par He then turned to the remaining syllable segment \includegraphics[page=1, height=0.7em]{Figure/rinth-yinbiao.pdf} (\cref{fig:scenario-1} \circlenum{4}), but found it difficult to generate a meaningful keyword. To explore further, he used the semantic brainstorming module, focusing on the target sense ``labyrinth'' and the previously selected keyword ``\thinspace\raisebox{-0.15em}{\includegraphics[page=1, height=0.9em]{Figure/Laba.pdf}}''. He manually added the semantic node ``\thinspace\raisebox{-0.15em}{\includegraphics[page=1, height=0.9em]{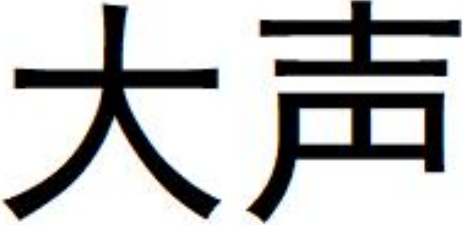}}\thinspace'' (loud) under ``\thinspace\raisebox{-0.15em}{\includegraphics[page=1, height=0.9em]{Figure/Laba.pdf}}'' (\cref{fig:scenario-1} \circlenum{5}), and, with system assistance, expanded ``labyrinth'' with the concept `` \raisebox{-0.15em}{\includegraphics[page=1, height=0.9em]{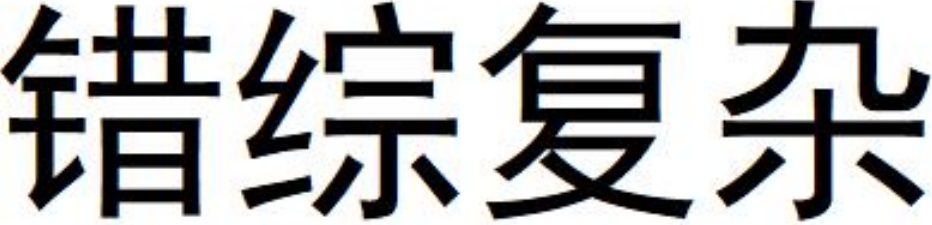}}\thinspace'' (intricate) (\cref{fig:scenario-1} \circlenum{6}). After selecting both nodes, he clicked \raisebox{-1ex}{\includegraphics[height=1.5em]{Figure/llm.pdf}} to request heuristic keyword suggestions for \includegraphics[page=1, height=0.7em]{Figure/rinth-yinbiao.pdf}. Based on the current semantic concepts and the selected phonetic fragment, the system proposed four candidate keywords: ``\thinspace\raisebox{-0.15em}{\includegraphics[page=1, height=0.9em]{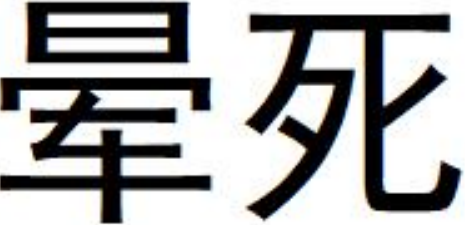}}\thinspace'' (faint), ``\thinspace\raisebox{-0.15em}{\includegraphics[page=1, height=0.9em]{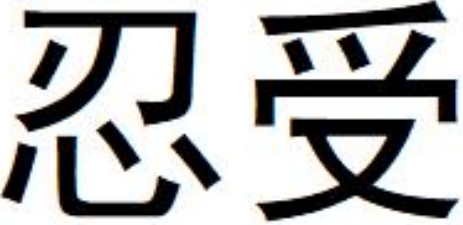}}\thinspace'' (endure), ``\thinspace\raisebox{-0.15em}{\includegraphics[page=1, height=0.9em]{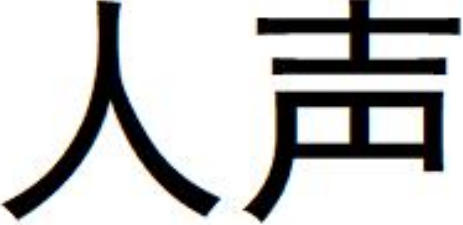}}\thinspace'' (human voice), and ``\thinspace\raisebox{-0.15em}{\includegraphics[page=1, height=0.9em]{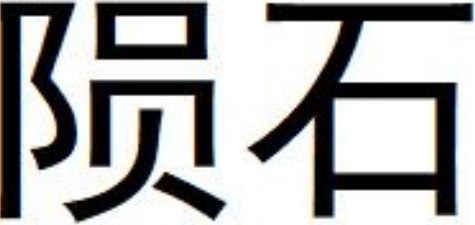}}'' (meteor) (\cref{fig:scenario-1} \circlenum{7}).  After reviewing the options, Yang chose ``\thinspace\raisebox{-0.15em}{\includegraphics[page=1, height=0.9em]{Figure/Yunsi.pdf}}\thinspace'', as it vividly conveyed the disorientation and pressure associated with a labyrinth, while also resonating with the noise and discomfort implied by the loudspeaker. He then clicked \raisebox{-1ex}{\includegraphics[height=1.5em]{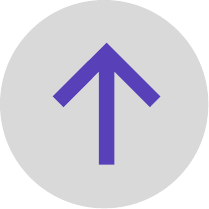}} to set ``\thinspace\raisebox{-0.15em}{\includegraphics[page=1, height=0.9em]{Figure/Yunsi.pdf}}\thinspace'' as the second keyword and began constructing a coherent associative path linking the target meaning with the selected keywords.

\par \textbf{\textit{Association Construction.} }Initially, Yang attempted to establish a connection based on his earlier note: ``\textit{The speaker can guide the way in the labyrinth.}'' However, he soon realized that this reasoning was overly positive and inconsistent with the negative connotation of ``faint'', resulting in a lack of coherence in the associative context. To reinforce the internal logic, he revisited the association map and his earlier descriptions, such as ``the intricacy of the labyrinth causes dizziness'' and ``the speaker noise is uncomfortable'' (\cref{fig:scenario-2} \circlenum{1}). Drawing on these elements, he decided to use ``dizziness'' as a new associative chain. He clicked on the connection between ``labyrinth'' and ``speaker'', entered the term ``dizziness'' in the relationship panel on the right (\cref{fig:scenario-2} \circlenum{2}) and clicked \raisebox{-1ex}{\includegraphics[height=1.5em]{Figure/llm.pdf}} (\cref{fig:scenario-2} \circlenum{3}). The system generated a heuristic prompt: ``\textit{The speaker may produce echoes in the labyrinth.}'' This suggestion sparked a new idea: in a labyrinth filled with echoing loudspeakers, the reverberating sound amplifies disorientation and discomfort, ultimately leading to the feeling of being ``faint''. Yang then composed the following associative sentence: ``\textit{I felt faint in the labyrinth filled with the echoes of speakers}'' (\cref{fig:scenario-2} \circlenum{4}). This sentence seamlessly linked the target meaning with the two selected keywords, forming a complete and logically coherent associative path.

\par \textbf{\textit{Image Formation.}} Yang then proceeded to the image formation component, where he used the canvas to further visualize the associative process. He first clicked the \raisebox{-1ex}{\includegraphics[height=1.5em]{Figure/add.pdf}}, drew a rectangular region on the canvas, and assigned it two tags: ``labyrinth'' and ``speaker.'' In the corresponding description field, he entered: ``\textit{A complex labyrinth lined with speakers, their acoustic echoes resonating in all directions}'' (\cref{fig:scenario-3} \circlenum{1}). Leveraging both the visual content and the semantic links between the tagged elements, the system then updated the Recall Path to reflect the evolving completeness of the conceptual mapping within the image.

\par Next, Yang drew a second rectangular region on the canvas and tagged it with ``faint''. To enrich the visual content, he clicked \raisebox{-1ex}{\includegraphics[height=1.5em]{Figure/llm.pdf}} to request system-generated visual cues. The system proposed several suggestions, such as ``feeble silhouette'' and ``collapsed on the ground.'' After briefly reviewing these prompts, Yang contributed his own imagery: ``\textit{A weak person lying on the ground, eyes swirling, with little stars spinning overhead}'', vividly capturing the feeling of faintness (\cref{fig:scenario-3} \circlenum{2}).

\par After defining both visual regions, Yang linked them using \raisebox{-1ex}{\includegraphics[height=1.5em]{Figure/associate.pdf}} on the canvas and added a brief note to specify their association: ``\textit{This person is inside the labyrinth}'' (\cref{fig:scenario-3} \circlenum{3}). This operation clarified the connection between the visual elements and reinforced the overall semantic coherence of the image. Upon noticing that the entire Recall Path had been fully activated, Yang selected his preferred visual style — \textit{Pixar-style animation} — and clicked \raisebox{-0.7ex}{\includegraphics[height=1.2em]{Figure/generate.pdf}} . The system then produced a complete memory image based on the canvas layout and corresponding textual descriptions. Satisfied with the generated result, he finalized the process by clicking \raisebox{-0.7ex}{\includegraphics[height=1.2em]{Figure/record.pdf}} to save the memory card (\cref{fig:scenario-3} \circlenum{4}), which captured the image, selected keywords, semantic associations, and time spent.

\section{Implementation Detail}
\label{detail}
\par During keyword selection phase, the system analyzes user-provided terms via a semantic tree and generates candidate keywords by combining semantic nodes with syllable fragments. The process uses a two-stage generation strategy: the first stage applies a high temperature setting of 1.0 to encourage diverse outputs, while the second stage uses a lower temperature of 0.3 to refine results for contextual coherence. In association construction phase, the model incorporates user-selected links, previous associations, and notes to heuristically suggest new semantic connections, supporting open-ended exploration. In image generation phase, the system integrates bounding-box layouts, region-specific descriptions, and inter-region relational text to guide the model in producing visually and semantically coherent outputs. All generative modules apply few-shot prompting with curated examples to ensure alignment with task goals and design intent. 

\onecolumn
\section{Prompts}

\label{app:prompt}

\begin{imageonly}
\setcounter{lstlisting}{0}
\begin{lstlisting}[caption={Keyword Exploration Recommendation Prompt}, label={keyword recommendation}]
"system": "You are an experienced English teacher, skilled at using the mnemonic method of homophony to help students memorize English words.

-- Task --
Given an IPA transcription and several 'related words,' please generate 20 candidate Chinese homophones/phrases.

Requirements for all candidates:
1) Pronunciation should be as close as possible to the English pronunciation (slight adjustments of initials/finals/tones are allowed, but it must sound natural and fluent).
2) Must be real, meaningful, and commonly used words or short phrases in modern Chinese (avoid obscure characters, dialects, or meaningless combinations).
3) Ensure semantic connection with the given 'related words' whenever possible (phonetic similarity takes priority, then semantic relevance).
4) Avoid vulgar/inappropriate content; avoid proper nouns, abbreviations, or names (unless highly generic).
5) Each candidate must include a concise and accurate Chinese explanation (one sentence), and highlight its relation to the associated word.

-- Examples --
{some_example}

-- User input --
{user_input}

-- Output format (strict) --
{output_format}

\end{lstlisting}
\end{imageonly}

\begin{imageonly}
\setcounter{lstlisting}{1}
\begin{lstlisting}[caption={Keyword Exploration Recommendation Reviewer Prompt}, label={keyword recommendation reviewer}]
"system": "You are a rigorous English teacher and reviewer.
Please select 4 'best homophones' from the 20 candidates, and rank them from highest to lowest quality.

-- Selection Rules (strict) --
1) Pronunciation similarity has priority:
- Preserve the initial sound of the English word.
- Ensure the main vowel is close to the original.
- The number of syllables and rhythm should be similar.
- If the original has a final consonant (-k/-t/-s/-n), try to reflect it.

2) Semantic association:
- Prefer words that relate closely to the 'related words.'
- If pronunciation similarity is equal, semantic closeness is decisive.

3) Frequency and naturalness:
- Must be common, natural expressions in modern Chinese.

4) Only output the final 4 items; do not include scoring or selection process.

-- User input --
{user_input}

-- Output format --
{output_format}


\end{lstlisting}
\end{imageonly}

\begin{imageonly}
\setcounter{lstlisting}{2}
\begin{lstlisting}[caption={Semantic Association Prompt}, label={canvus association suggestion}]
"system": "You are a semantic association expert. 
Given a target word, output a related Chinese word.

Requirements:
- Only return plain Chinese text, no explanation or extra characters.
- No more than 5 Chinese characters.
- Extend the semantic field of the original word.

-- Examples --
{some_examples}

-- User input --
{user_input}

-- Output format --
{output_format}

\end{lstlisting}
\end{imageonly}

\begin{imageonly}
\setcounter{lstlisting}{3}
\begin{lstlisting}[caption={Association Recommendation Prompt}, label={association recommendation}]
"system": "You are an expert at expressing relationships indirectly. 
Given two entities and associated clues, please provide 3-5 subtle but professional Chinese sentences that hint at their relationship.

Requirements:
- Each sentence should be implicit yet clear, avoiding direct conclusions.
- Keep expressions concise and natural, consistent with everyday reasoning.
- Provide clues from different perspectives to allow readers to infer the relationship.
- Avoid overly poetic or vague phrasing; maintain professionalism and readability.

-- User input --
{user_input}

-- Output format --
{output_format}

\end{lstlisting}
\end{imageonly}

\begin{imageonly}
\setcounter{lstlisting}{4}
\begin{lstlisting}[caption={Imagery Recommender Prompt}, label={Imagery Recommender}]
"system": "You are an 'Imagery Recommender.' 
Based on user-selected Chinese semantic nodes, generate 'visual elements' that are suitable to be drawn.

Strict requirements:
- Output only a JSON array (e.g., [\"skyscrapers\", \"neon lights\"]).
- Each item should be 2-6 Chinese characters, concrete and visual, preferably nouns.
- Metaphorical/associative mapping is allowed, but results must still be visually depictable.
- Avoid abstract words (e.g., 'modernization'), English, emojis, numbers, or sequences.
- Avoid repetition or synonyms.
- Content must be healthy and compliant.

-- Examples --
{some_examples}

-- User input --
{user_input}

-- Output format --
{output_format}


\end{lstlisting}
\end{imageonly}

\begin{imageonly}
\setcounter{lstlisting}{5}
\begin{lstlisting}[caption={Scene Relation Designer Prompt}, label={Scene Relation Designer}]
"system": "You are a 'Scene Relation Designer.' 
Based on imagery from the left and right, write one-sentence descriptions that connect the two.

Strict requirements:
- Output only a JSON array (e.g., [\"He wore a suit but stared at his empty wallet in silence\"]).
- Each sentence must be 12-26 Chinese characters, natural and colloquial, like a camera shot.
- Must include elements from both sides (directly or by implication).
- Can include transitions, causality, or contrast.
- Do not write explanations, numbering, or English.
- Content must be safe and appropriate.

-- Examples --
{some_examples}

-- User input --
{user_input}

-- Output format --
{output_format}

\end{lstlisting}
\end{imageonly}

\begin{imageonly}
\setcounter{lstlisting}{6}
\begin{lstlisting}[caption={Visual Design and Illustration Assistant Prompt}, label={visual design and illustration assistan}]
"system": "You are a professional visual design and illustration assistant.

-- Task --
I will provide you with a set of constraints for image generation, including:
1. A reference wireframe sketch image (showing only boxes and labels). Please follow this layout strictly, but visually integrate the elements with the background so that the final image looks natural.
2. A description of what should be drawn inside each box (objects, characters, items, scenes, etc.).
3. The relationships between these elements.

Please strictly follow these rules when generating the image:
- Each element must be placed accurately according to the bounding box provided.
- Maintain spatial and logical relationships between elements.
- All elements should share a consistent style (I will specify the overall style, e.g., realistic, cartoon, cyberpunk, sketch, etc.).
- Do not add any elements that I did not specify.
- The overall composition should be clear, coherent, and visually appealing.
- Do not retain the black bounding box lines; treat the wireframe only as layout guidance, blending it naturally into the image.

-- User Input --
{user_input}

-- Mandatory Constraints --
- Each element must fully fit within its assigned bounding box.
- Do not draw additional objects outside of the given boxes.
- Only draw the elements described and the minimal background necessary for coherence.

\end{lstlisting}
\end{imageonly}

\section{Baseline Systems}
\label{baseline_system}
\begin{figure*}[h]
    \centering
    \includegraphics[width=\textwidth]{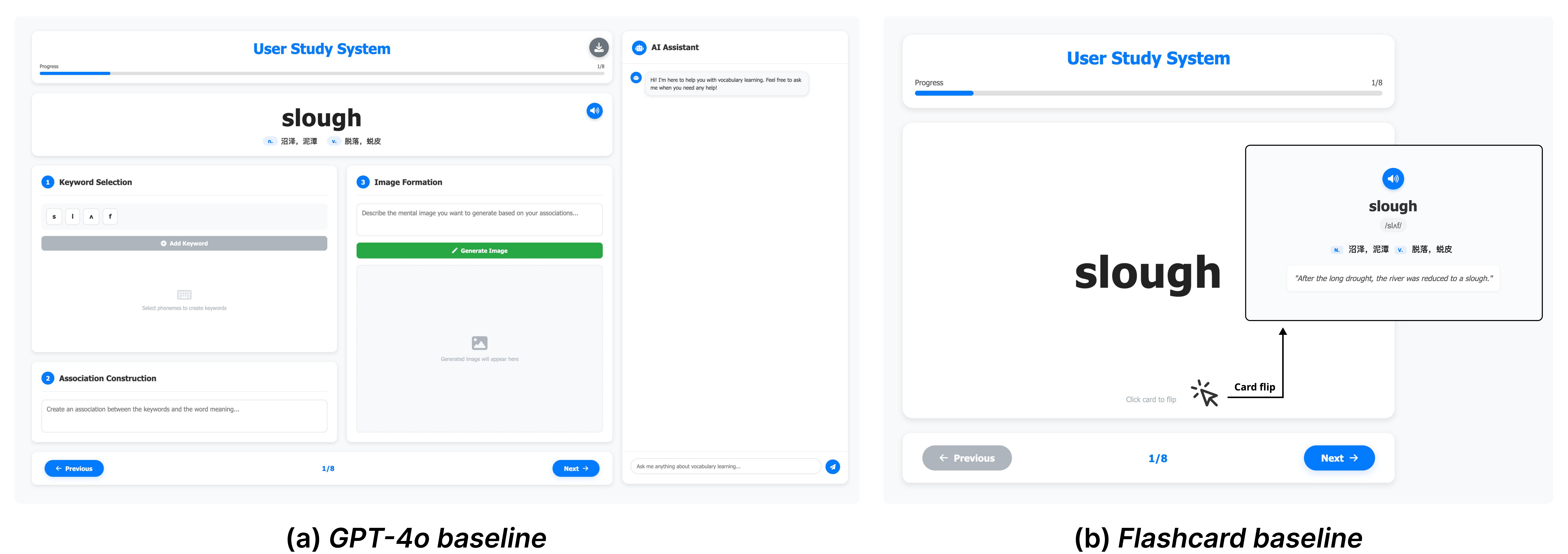}
    \caption{ (a) GPT-4o baseline system and (b) Flashcard baseline in user study~1.}
    \label{fig:baseline}
\end{figure*}
\clearpage

\section{Participants in Study~1}
\label{participants-study1}
\begin{table}[h]
  \centering
  \caption{Detailed information of the participants in study~1.}
  \label{tab:participants-study1}
  \setlength{\tabcolsep}{5pt}
  \renewcommand{\arraystretch}{1.1}
  \begin{tabular}{l l c l| l l c l | l l c l}
    \toprule
    \multicolumn{4}{c|}{\textbf{WordCraft}} &
    \multicolumn{4}{c|}{\textbf{GPT-4o}} &
    \multicolumn{4}{c}{\textbf{Flashcard}} \\
    \midrule
    ID & Gender & Age & Frequency &
    ID & Gender & Age & Frequency &
    ID & Gender & Age & Frequency \\
    \midrule
    P1  & Male   & 22 & Occasionally &
    P17 & Male   & 24 & Rarely &
    P33 & Male   & 19 & Occasionally \\
    P2  & Male   & 23 & Rarely &
    P18 & Female & 20 & Sometimes &
    P34 & Female & 20 & Frequently \\
    P3  & Female & 21 & Occasionally &
    P19 & Female & 24 & Occasionally &
    P35 & Male   & 21 & Rarely \\
    P4  & Male   & 21 & Sometimes &
    P20 & Female & 23 & Occasionally &
    P36 & Female & 22 & Occasionally \\
    P5  & Male   & 20 & Frequently &
    P21 & Male   & 24 & Rarely &
    P37 & Male   & 23 & Frequently \\
    P6  & Female & 21 & Frequently &
    P22 & Female & 22 & Frequently &
    P38 & Female & 21 & Sometimes \\
    P7  & Male   & 23 & Rarely &
    P23 & Male   & 22 & Frequently &
    P39 & Male   & 24 & Occasionally \\
    P8  & Male   & 22 & Sometimes &
    P24 & Male   & 20 & Occasionally &
    P40 & Female & 19 & Rarely \\
    P9  & Female & 24 & Occasionally &
    P25 & Female & 21 & Sometimes &
    P41 & Male   & 20 & Occasionally \\
    P10 & Female & 19 & Occasionally &
    P26 & Male   & 20 & Sometimes &
    P42 & Female & 22 & Sometimes \\
    P11 & Female & 22 & Frequently &
    P27 & Female & 22 & Rarely &
    P43 & Male   & 23 & Frequently \\
    P12 & Male   & 23 & Rarely &
    P28 & Female & 19 & Frequently &
    P44 & Female & 21 & Rarely \\
    P13 & Female & 23 & Sometimes &
    P29 & Female & 22 & Rarely &
    P45 & Male   & 22 & Occasionally \\
    P14 & Female & 20 & Frequently &
    P30 & Male   & 24 & Sometimes &
    P46 & Female & 24 & Sometimes \\
    P15 & Male   & 19 & Rarely &
    P31 & Female & 21 & Occasionally &
    P47 & Male   & 19 & Frequently \\
    P16 & Female & 20 & Occasionally &
    P32 & Male   & 22 & Occasionally &
    P48 & Female & 20 & Occasionally \\
    \bottomrule
  \end{tabular}
\end{table}

\section{Participants in Study~2}
\label{participants-study2}

\begin{table}[h]
  \centering
  \caption{Detailed information of the participants in study~2.}
  \label{tab:participants-study2}
  \setlength{\tabcolsep}{6pt}
  \renewcommand{\arraystretch}{1.1}
  \begin{tabular}{l c c l | l c c l}
    \toprule
    \multicolumn{4}{c|}{\textbf{Group A}} &
    \multicolumn{4}{c}{\textbf{Group B}} \\
    \midrule
    \textbf{ID} & \textbf{Gender} & \textbf{Age} & \textbf{Frequency} & \textbf{ID} & \textbf{Gender} & \textbf{Age} & \textbf{Frequency} \\
    \midrule
    A1 & Male   & 24 & Frequently & B1 & Female & 24 & Sometimes \\
    A2 & Female & 21 & Rarely & B2 & Male   & 19 & Frequently \\
    A3 & Female & 22 & Sometimes & B3 & Female & 22 & Occasionally \\
    A4 & Male   & 21 & Sometimes & B4 & Male   & 21 & Rarely \\
    A5 & Male   & 20 & Occasionally & B5 & Female & 23 & Rarely \\
    A6 & Female & 22 & Rarely & B6 & Female & 23 & Occasionally \\
    A7 & Male   & 22 & Occasionally & B7 & Male   & 21 & Frequently \\
    A8 & Female & 24 & Frequently & B8 & Female & 24 & Sometimes \\
    A9 & Male   & 24 & Frequently & B9 & Female & 20 & Occasionally \\
    A10 & Female & 24 & Occasionally & B10 & Male   & 19 & Occasionally \\
    \bottomrule
  \end{tabular}
\end{table}

\end{document}